\documentclass{article}

\usepackage{arxiv}

\usepackage[utf8]{inputenc}     % allow utf-8 input
\usepackage[T1]{fontenc}        % use 8-bit T1 fonts
\usepackage{hyperref}           % hyperlinks
\usepackage{url}                % URL typesetting
\usepackage{booktabs}           % professional-quality tables
\usepackage{amsfonts}           % math fonts (e.g., \mathbb)
\usepackage{amssymb}            % common math symbols
\usepackage{mathtools}          % additional math tools
\usepackage{mathrsfs}           % script fonts for math (\mathscr)
\usepackage{nicefrac}           % compact fractions, e.g., 1/2
\usepackage{microtype}          % improved typography
\usepackage{lipsum}             % filler text (remove when done)
\usepackage{graphicx}           % for including graphics
\usepackage{subcaption}         % subfigures and subcaptions
\usepackage[space]{grffile}     % allow spaces in file names for graphics
\usepackage{natbib}             % bibliography management
\usepackage{doi}                % DOI formatting
\usepackage{enumitem}           % customized list environments
\usepackage{kantlipsum}         % additional filler text
\usepackage{soul}               % letter spacing/highlighting
\usepackage{verbatim}           % for verbatim text
\usepackage{color}              % color support
\usepackage{algorithm}          % algorithms
\usepackage{algorithmic}        % algorithmic environment
\usepackage{comment}            % for commenting out blocks

\newcommand{\indep}{\perp \!\!\! \perp}
\newcommand{\mA}{\mathcal{A}}  % action space A
\newcommand{\mS}{\mathcal{S}}

\newtheorem{theorem}{Theorem}
\newtheorem{asmp}{Assumption}

\newtheorem{Coro}{Corollary}
\newtheorem{lemma}{Lemma}

\newtheorem{Def}{Definition}

\renewcommand{\hat}{\widehat}

\title{Off-policy Evaluation with Deeply-abstracted States}
\author{
\hspace{1mm}Meiling Hao\thanks{Co-first author} \\
	School of Statistics\\
	University of International Business and Economics\\
	\texttt{meilinghao@uibe.edu.cn} \\
    \And
 \hspace{1mm} Pingfan Su$^*$ \\
	Department of Statistics\\
	London School of Economics and Political Science\\
	\texttt{p.su1@lse.ac.uk} \\
	\And
	\hspace{1mm} Liyuan Hu \\
	Department of Statistics\\
	London School of Economics and Political Science\\
	\texttt{l.hu11@lse.ac.uk} \\
	\And
    \hspace{1mm} Zolt$\acute{a}$n Szab$\acute{o}$ \\
	Department of Statistics\\
	London School of Economics and Political Science\\
	\texttt{Z.Szabo@lse.ac.uk} \\
	\And
    \hspace{8mm} Qingyuan Zhao \\
   \hspace{8mm}Statistical Laboratory, DPMMS\\
  \hspace{8mm} University of Cambridge \\
	\hspace{8mm}\texttt{qyzhao@statslab.cam.ac.uk} \\
    \And
    \hspace{15mm} Chengchun Shi \thanks{Corresponding author} \\
	\hspace{15mm}Department of Statistics\\
	\hspace{15mm}London School of Economics and Political Science\\
	\hspace{15mm}\texttt{Shi7@lse.ac.uk} \\
}
\date{}
\begin{document}

\maketitle  % Make the title section

\begin{abstract}
Off-policy evaluation (OPE) is crucial for assessing a target policy's impact offline before its deployment. However, achieving accurate OPE in large state spaces remains challenging. This paper studies state abstractions -- originally designed for policy learning -- in the context of OPE. Our contributions are three-fold: (i) We derive a bias-variance decomposition for the mean squared error (MSE) of various post-abstraction OPE estimators, e.g., estimators when applied to the abstract state space. Based on this decomposition, we demonstrate that the iterative application of Markov state abstractions (MSAs) can sequentially reduce the MSE of the resulting OPE estimator, ultimately leading to a substantial error reduction. In general, the finer the abstract state space, the lower the MSE of these estimators. (ii) Building on these insights, we propose a novel iterative procedure that sequentially projects the original state space into a smaller space while preserving the Markov assumption. This results in a deeply-abstracted state space, which is substantially smaller than the original and considerably reduces the sample complexity of OPE. (iii) We conduct numerical experiments to demonstrate the advantages of the proposed state abstraction for OPE in finite samples and to validate our theoretical findings.
\end{abstract}

%%%%%%%%%%%%%%%%%%%%%%%%%%%%%%%%%%%%%%%%%%%%%%%%%%%%%%%%%%%%%%%%
%% Section: Submission of papers to RLJ/RLC
%%%%%%%%%%%%%%%%%%%%%%%%%%%%%%%%%%%%%%%%%%%%%%%%%%%%%%%%%%%%%%%%
\section{Introduction}
\textbf{Motivation.}
Off-policy evaluation (OPE) serves as a crucial tool for assessing the impact of a newly developed policy using a pre-collected dataset before its deployment in 
high-stake applications, such as healthcare \citep{murphy2001marginal}, recommendation systems \citep{chapelle2011empirical}, education \citep{mandel2014offline}, dialog systems \citep{jiang2021towards} and robotics \citep{levine2020offline}. A fundamental challenge in OPE is its %inherent 
``off-policy'' nature, wherein the target policy to be evaluated differs from the behavior policy that generates the offline data. This distributional shift is particularly pronounced in environments with large state spaces of high cardinality. Theoretically, %it has been demonstrated that 
the error bounds for estimating the target policy's Q-function and value decrease rapidly as the state space dimension increases \citep{hao2021sparse,chen2022well}. Empirically, %high-dimensionality 
large state space significantly challenges the performance of state-of-the-art OPE algorithms \citep{fu2020benchmarks,voloshin2021empirical}. 

Although different policies induce different trajectories in %high-dimensional 
the %original 
large ground state space, they can produce similar paths when restricted to relevant, lower-dimensional  state spaces \citep{pavse2023scaling}. Consequently, applying OPE to these abstract spaces can substantially mitigate the distributional shift between target and behavior policies, enhancing the accuracy in predicting the target policy's value. This makes state abstraction, designed to reduce state space %dimensionality, 
cardinality, particularly appealing for OPE. However, despite the extensive literature on studying state abstractions for policy learning (see Section \ref{sec:relatedworks} for details), it has been hardly explored in the context of OPE.

\textbf{Contributions.}
 This paper aims to systematically investigate state abstractions for OPE to address the aforementioned gap. Our main contributions include:

 \begin{enumerate}[leftmargin=*]
     \item Derivation of  a bias-variance decomposition for  the mean squared error (MSE) of various OPE estimators. Based on this decomposition, we demonstrate that the iterative application of Markov state abstractions (MSAs) can sequentially reduce the MSE of the resulting OPE estimator and ultimately lead to a substantial reduction in MSE. In general, the finer the abstract state space, the lower the MSE of these estimators.

    \item  
    Development of a novel iterative state abstraction procedure  sequentially compresses the state space. Specifically, within each iteration, our algorithm projects the original state space into a smaller space while preserving the Markov assumption. This results in a deeply-abstracted state, which aims to maximally compress the state space and reduce the sample complexity of OPE in large state spaces.
    \item 
    Empirical validations of OPE methods when applied to the proposed abstract state spaces. Our results  highlight (i) the effectiveness of state abstraction in improving OPE, (ii) the benefits of multi-iteration abstractions over single-iteration abstractions and (iii) the superiority of the proposed state abstraction over existing approaches.
\end{enumerate}
\textbf{Organization}. The rest of the paper is structured as follows. Section~\ref{sec:relatedworks} is dedicated to the literature review of related works. MDP-related notions and OPE methodologies relevant to our proposal are recalled in Section~\ref{sec:preliminaries}. Our unified theory for the MSE decomposition and proposed state abstractions for OPE are presented in Section~\ref{sec:proposed-state-abstractions}. Section~\ref{sec:experiments} conducts numerical experiments to demonstrate the efficiency of our approach and to validate our theories. Finally, Section~\ref{sec:conclusion} concludes the paper.
\vspace{-1em}
\section{Related work}\label{sec:relatedworks}
\vspace{-1em}
Our proposal is closely related to  OPE and state abstraction. Additional related work on confounder selection in causal inference is relegated to Appendix~\ref{sec:confounder}. 

\textbf{Off-policy evaluation}. 
OPE aims to estimate the expected return 
of a given target
policy,  utilizing historical data generated by a possibly different behavior policy \citep{dudik2014doubly,uehara2022review}.
The majority of methods in the literature can be classified into the following three categories: 
\begin{enumerate}[leftmargin=*]
    \item \textbf{Value-based methods} that estimate the target policy's return by learning either a value function \citep{sutton2008convergent,luckett2019estimating,li2024high} or a Q-function \citep{le2019batch, feng2020accountable,hao2021bootstrapping,liao2021off,chen2022well,shi2020statistical} from the data.
    \item \textbf{Importance sampling (IS) methods} that adjust the observed rewards using the IS
    ratio, i.e., the ratio 
    of the target policy over the behavior policy, to address their distributional shift. There are two major types: sequential IS \citep[SIS,][]{precup2000eligibility,thomas2015high,hanna2019importance,hu2023off} which employs the product of IS ratios, and marginalized IS \citep[MIS,][]{liu2018breaking,nachum2019dualdice,xie2019towards,dai2020coindice,yin2020asymptotically,wang2023projected} which uses the MIS ratio to mitigate the high variance of the SIS estimator.
    \item \textbf{Doubly robust methods} or their variants that employ both the IS ratio and the value/reward function to enhance the robustness of OPE  \citep{zhang2013robust,jiang2016doubly,thomas2016data,farajtabar2018more,kallus2020double,tang2020doubly,uehara2020minimax,shi2021deeply,kallus2022efficiently,liao2022batch,xie2023semiparametrically}. 
\end{enumerate}
However, none of the aforementioned works studied state abstraction, 
which is our primary focus.

\textbf{State abstraction.}
State abstraction aims to obtain a parsimonious state representation to simplify the sample complexity of reinforcement learning (RL), while ensuring that the optimal policy restricted to the abstract state space attains comparable values as in the original, ground state space. There is an extensive literature on the theoretical and methodological development of state abstraction, particularly bisimulation — a type of abstractions that preserve the Markov property   \citep{singh1994reinforcement,dean1997model,givan2003equivalence,ferns2004metrics, ravindran2004algebraic,jong2005state,li2006towards,ferns2011bisimulation,abel2016near,wang2017sufficient,castro2020scalable,allen2021learning,abel2022theory}. In particular, \citet{li2006towards} analyzed five irrelevance conditions for optimal policy learning.
Unlike the aforementioned works that focus on policy learning, we %introduce irrelevance conditions for 
focus on OPE. %and propose abstractions that satisfy these irrelevant properties. 
Meanwhile, the proposed abstraction %for achieving irrelevance for the MIS ratio  resembles 
is closely related the Markov state abstraction developed by \citet{allen2021learning} in the context of policy learning, while relaxing their requirement for the behavior policy to be Markovian. 

More recently, \cite{pavse2023scaling} made a pioneering attempt to study state abstraction for OPE, proving its benefits in improving the accuracy of MIS estimators. %However, they primarily focused on MIS estimators.  \hao{
\cite{pavse2024state} introduced a new diffuse metric to measure the similarity between state-action pairs 
and employed it to learn state-action representations for subsequent use in fitted Q-evaluation \citep[FQE,][]{le2019batch} -- a popular value-based OPE method. Different from these methods that are tailored to either MIS or FQE, our analysis applies to a broader range of OPE estimators, covering all three aforementioned categories. \cite{chaudhari2024abstract} leveraged state abstraction to distill complex, continuous systems into compact, discrete models for OPE. They theoretically showed that certain abstractions yield consistent OPE estimators applied to the abstract space. However, they did not theoretically demonstrate that state abstraction can reduce the MSE of the resulting OPE estimator, nor did they provide a comprehensive discussion on the approaches for computing these abstractions. %Moreover, 

Lastly, state abstraction is also related to variable selection 
\citep{kolter2009regularization,geist2011,geist2012dantzig,nguyen2013online,fan2016sequential,guo2017sample,shi2018high,zhang2018variable,qi2020multi,hao2021sparse,ma2023sequential} as well as representation learning for both policy learning \citep[see e.g.,][]{gelada2019deepmdp,zhang2020learning,uehara2021representation} and OPE \citep[see e.g.,][]{wang2021instabilities,chang2022learning,ni2023learning}.

We remark that none of the aforementioned works considered the iterative use of state abstraction for accurate OPE, which is the cornerstone of our methodology. 

\section{Preliminaries} \label{sec:preliminaries}
In this section, we first introduce some key concepts relevant to OPE in RL, such as MDP, target and behavior 
policies, value functions, 
IS ratios (Section~\ref{sec:data}). Subsequently, we review four prominent OPE methodologies in Section~\ref{sec:OPEmethod}.

\subsection{Data generating process, policy, value and IS ratio}\label{sec:data} 
\textbf{Data}. Assume the offline dataset $\mathcal{D}$ comprises $N$ trajectories, each containing a sequence of state-action-reward-next-state tuples $(S_t,A_t,R_t,S_{t+1})_{1\le t\le T}$ following a finite MDP, %model, 
denoted by $\mathcal{M}=\langle \mathcal{S},\mathcal{A},\mathcal{T},\mathcal{R},\rho_0,\gamma\rangle$.  Here, $\mathcal{S}$ and $\mathcal{A}$ are the discrete state and action spaces, both with finite cardinalities%\z{Isn't this what people call finite MDP ($|\mathcal{A}|<\infty$ and  $|\mathcal{S}|<\infty$)? -- We might wish to use that name.}
, $\mathcal{T}$ and $\mathcal{R}$ are the state transition and reward functions, $\rho_0$ denotes the initial state distribution, and $\gamma\in (0,1)$ is the discount factor. 
%Specifically, 
The data is generated as follows: 
\begin{enumerate}[leftmargin=*]
    \item At the initial time, the state $S_1$ is generated according to $\rho_0$;
    \item Subsequently, at each time $t$, the agent finds the environment in a specific state $S_t\in \mathcal{S}$ and selects an action $A_t\in \mathcal{A}$ according to a behavior policy $b$  
such that $\mathbb{P}(A_t=a|S_t)=b(a|S_t)$;
    \item The environment delivers an immediate reward $R$ with an expected value of $\mathcal{R}(A_t,S_t)$,
    and transits into the next state $S_{t+1}\stackrel{d}{\sim} \mathcal{T}(\bullet \mid A_t,S_t)$ according to the transition function $\mathcal{T}$. 
\end{enumerate}
Notice that both the reward and transition functions rely only on the current state-action pair $(S_t,A_t)$, independent of the past data history. This ensures that the data satisfies the Markov assumption. 

\textbf{Policy and value}. Let $\pi$ denote a given target policy we wish to evaluate. %We assume that when $\pi(a|s)>0$, $b(a|s)>0$. \hao{
We use $\mathbb{E}^{\pi}$ and $\mathbb{P}^{\pi}$ %($\mathbb{E}$ and $\mathbb{P}$) 
to denote the expectation and probability assuming the actions are chosen according to $\pi$ %($b$) 
at each time. %}
The regular $\mathbb{E}$ and $\mathbb{P}$ without superscript are taken with respect to the behavior policy $b$. 
Our objective lies in estimating the expected cumulative reward under $\pi$, denoted by $J(\pi)=\mathbb{E}^{\pi}\Big[\sum_{t=1}^{+\infty}\gamma^{t-1} R_t\Big]$ using the offline dataset generated under a different policy $b$.  
%and $\pi^*$ denote the optimal policy. % $\pi^b$ be the behaviour policy, and $\pi^e$ be the evaluation policy. 
%Let $V^{\pi}(s)$ denote the expected value of the cumulative reward under policy $\pi$ starting from a given state $s$. Similarly, let $Q^{\pi}( a,s)$ denote the expected value of the cumulative reward when first taking an action $a$ from the state $s$ and then following policy $\pi$ afterward.
Additionally, denote $V^{\pi}$ and $Q^{\pi}$ as the state value function and state-action value function (better known as the Q-function), namely,
\begin{align}\label{eqn:Qandvalue}
V^{\pi}(s)=\mathbb{E}^{\pi}\Big[\sum_{t=1}^{+\infty}\gamma^{t-1}R_t|S_1=s\Big]\,\,\hbox{and}\,\,Q^{\pi}(a,s)=\mathbb{E}^{\pi}\Big[\sum_{t=1}^{+\infty}\gamma^{t-1}R_t|S_1=s,A_1=a\Big].
\end{align}
%where $\gamma\in [0,1)$ is the discount factor.
%When $\pi=\pi^*$,
These functions are pivotal in developing value-based estimators, as described in Method 1 of Section \ref{sec:OPEmethod}. %Moreover, %\hao{In general???}, 
%we use $\pi^*$ to denote the optimal policy that maximizes $J(\pi)$, %\hao{when it exists}, 
%i.e., $\pi^*\in \arg\max_{\pi} J(\pi)$. %and write the optimal Q- and value functions $Q^{\pi^*}$, $V^{\pi^*}$ as $Q^{*}$, $V^*$ for brevity. 

\textbf{IS ratio}. We also introduce the IS ratio $\rho^{\pi}(a,s)=\pi(a|s)/b(a|s)$, which quantifies the discrepancy between the target policy $\pi$ and the behavior policy $b$. Furthermore, define the MIS ratio 
\begin{eqnarray}\label{eqn:MISratio}
w^{\pi}(a,s)=(1-\gamma)\sum_{t\ge 1} \frac{\gamma^{t-1}\mathbb{P}^{\pi}(S_t=s, A_t=a)}{\displaystyle\lim_{T \rightarrow \infty} \mathbb{P}(S_T=s, A_T=a)}.
\end{eqnarray}
Here, the numerator represents %the discounted visitation probability \z{
the discounted probability of visiting a given state-action pair under the target policy $\pi$ -- 
%(indicated by the superscript $\pi$ in $\mathbb{P}^{\pi}$), 
a crucial component in policy-based learning %approaches 
for estimating the optimal policy (denoted by $\pi^*$) %, such as policy gradient \citep{sutton1999policy} and trust region policy optimization 
\citep{sutton1999policy,schulman2015trust}. The denominator corresponds to the limiting state-action distribution under the behavior policy. These ratios are fundamental in constructing IS estimators, as detailed in Methods 2 and 3 of Section \ref{sec:OPEmethod}.

\subsection{OPE methodologies}\label{sec:OPEmethod}
We focus on four OPE methods, covering the three families of estimators introduced in Section \ref{sec:relatedworks}. 
Each method employs a specific formula to identify $J(\pi)$, which we detail below. The first method is a popular value-based approach -- the Q-function-based method. The second and third methods are the two major IS estimators: SIS and MIS. The fourth method is a semi-parametrically efficient doubly robust method, double RL (DRL), known for achieving the smallest possible MSE among a broad class of OPE estimators \citep{kallus2020double,kallus2022efficiently,liao2022batch}. 

\textbf{Method 1 (Q-function-based method)}. For a given Q-function $Q$, define $f_1(Q)$ as the estimating function   $\sum_{a \in \mathcal{A}}\pi(a|S_1)Q(a,S_1)$ with $S_1$ being the initial state. By \eqref{eqn:Qandvalue} and the definition of $J(\pi)$, it is immediate to see that $J(\pi)=\mathbb{E} [f_1(Q^{\pi})]$. %\z{What is the $\E$ taken with respect to, what is random here?} {\color{red} $\E$ is short for $\E^{b}$}. 
This motivates the Q-function-based method which uses a plug-in estimator to approximate  $\mathbb{E} [f_1(Q^{\pi})]$ and estimate $J(\pi)$. In particular, $Q^{\pi}$ can be estimated by Q-learning type algorithms such as FQE, and the expectation can be approximated via the empirical initial state distribution.  

\textbf{Method 2 (Sequential importance sampling)}. For a given IS ratio $\rho^{\pi}$, 
%SIS is motivated by the change of measure theorem. Specifically, 
let $\rho_{1:t}^{\pi}$ denote the sequential IS ratio $\prod_{j=1}^t \rho^{\pi}(A_j,S_j)$. It follows from the change of measure theorem that the counterfactual reward $\mathbb{E}^{\pi} (R_t)$ is equivalent to $\mathbb{E} (\rho^{\pi}_{1:t} R_t)$ whose expectation is taken with respect to the offline data distribution. Assuming all trajectories in $
\mathcal{D}$ terminate after a finite time $T$, this allows us to represent $J(\pi)$ by $\mathbb{E} [f_2 (\rho^{\pi})]$ where $f_2(\rho^{\pi})=\sum_{t=1}^{T} \gamma^{t-1} \rho_{1:t}^{\pi} R_t$. %The approximation error is bounded by $O(\gamma^T)$, which decays exponentially fast with respect to $T$. 
SIS utilizes a plug-in estimator to first estimate $\rho^{\pi}$ (when the behavior policy is unknown), and then employs this estimator, along with the empirical data distribution, to approximate $\mathbb{E} [f_2 (\rho^{\pi})]$. However, a notable limitation of this estimator is its rapidly increasing variance due to the use of the SIS ratio $\rho^{\pi}_{1:t}$. Specifically, this variance tends to grow exponentially with respect to $t$, a phenomenon referred to as \textit{the curse of horizon} \citep{liu2018breaking}.

\textbf{Method 3 (Marginalized importance sampling)}. The MIS estimator is designed to overcome the limitations of the SIS estimator. It breaks the curse of horizon by taking the structure of the MDP model into account. As noted previously, under the Markov assumption, the reward depends only on the current state-action pair, rather than the entire history. This insight allows us to replace the SIS ratio with the MIS ratio, which depends solely on the current state-action pair. This modification considerably reduces variance because $w^{\pi}$ is no longer history-dependent. %Specifically, 
Assuming the data trajectory is stationary over time -- that is, all state-action-reward $(S,A,R)$ triplets have the same distribution -- it can be shown that $J(\pi)=\mathbb{E} [f_3(w^{\pi})]$ where $f_3(w^{\pi})=(1-\gamma)^{-1}w^{\pi}(A,S)R$ for any triplet $(S,A,R)$. Both $w^{\pi}$ and the expectation can be approximated using offline data. 

\textbf{Method 4 (Double reinforcement learning)}. DRL combines the Q-function-based method with MIS. Let $f_4(Q,w)=f_1(Q)+(1-\gamma)^{-1}w(A,S) [R+\gamma \sum_a \pi(a|S') Q(a,S')-Q(A,S)]$, where $f_1$ is defined in Method 1 and $(S,A,R,S')$ denotes a state-action-reward-next-state tuple. Under the stationarity assumption, it can be shown that $J(\pi)=\mathbb{E} [f_4(Q,w)]$ when either $Q=Q^{\pi}$ or $w=w^{\pi}$ \citep{kallus2022efficiently}. DRL proposes to learn both $Q^{\pi}$ and $w^{\pi}$ from the data, employ these estimators to calculate $\mathbb{E} [f_4(Q,w)]$ and approximate the expectation with the empirical data distribution. The resulting estimator benefits from double robustness: it is consistent when either $Q^{\pi}$ or $w^{\pi}$ is correctly specified. 

\section{Proposed state abstractions for policy evaluation} \label{sec:proposed-state-abstractions} 
This section presents our proposed iterative procedure for state abstraction. We begin by offering a bias-variance decomposition for the MSE of various post-abstraction OPE estimators, i.e., estimators applied to a given abstract space, and demonstrate that when the abstraction is an MSA, it generally reduces the MSE of the OPE estimator. Meanwhile, the finer the abstraction, the smaller the MSE (see Section~\ref{sec:postMSE}). Inspired by this analysis, we put forward our iterative procedure (in Section~\ref{sec:iterative}).

\subsection{A unified theory for post-abstraction OPE estimators}\label{sec:postMSE}
Recall that $\mathcal{M}=\langle \mathcal{S},\mathcal{A},\mathcal{T},\mathcal{R},\rho_0,\gamma\rangle$ denotes the ground MDP. An abstraction $\phi$ is a mapping from the state space $\mathcal{S}$ to an abstract state space $\mathcal{X}=\{\phi(s): s\in \mathcal{S}\}$. It transforms the raw MDP data into a new sequence of abstract state-action-reward-next-state tuples $(\phi(S), A, R, \phi(S'))$. Recall that $\pi$ represents the target policy we aim to evaluate. Our analysis focuses on a specific subclass of abstractions that are $\pi$-irrelevant and preserve the Markov property, known as Markov state abstractions \citep{allen2021learning}, and their formal definitions are provided below. 
\begin{Def}[$\pi$-irrelevance]\label{def:pi}
    $\phi$ is $\pi$-irrelevant if for any $s^{(1)}, s^{(2)} \in \mathcal{S}$ whenever $\phi(s^{(1)})=\phi(s^{(2)})$, we have 
$\pi(a|s^{(1)})=\pi(a|s^{(2)})$ for any $a\in \mathcal{A}$.
\end{Def}
\begin{Def}[Markov State Abstraction]\label{Def:MSA}
$\phi$ is an MSA if and only if both the transition and reward function on the transformed data satisfy the Markov property, i.e., for any $t$ and $x$, we have
\begin{eqnarray*}
    \mathbb{E}[R_t|A_t,\phi(S_t),A_{t-1},\phi(S_{t-1}),\cdots,A_1,\phi(S_1)]&=&\mathbb{E}[R_t|A_t,\phi(S_t)], \\
    \mathbb{P}(\phi(S_{t+1})=x|A_t,\phi(S_t),A_{t-1},\phi(S_{t-1}),\cdots,A_1,\phi(S_1))&=&\mathbb{P}(\phi(S_{t+1})=x|A_t,\phi(S_t)).
\end{eqnarray*} 
\end{Def}
The $\pi$-irrelevance condition in Definition \ref{def:pi} ensures that the Q-function $Q_{\phi}^{\pi}$, value function $V_{\phi}^{\pi}$ and IS ratios $\rho_{\phi}^{\pi}$, $w_{\phi}^{\pi}$ are well-defined on the abstract space. According to Definition \ref{Def:MSA}, it is immediate to see that if $\phi$ is an MSA, then the transformed data  tuples $(\phi(S), A, R,\phi(S'))$ retain the MDP structure (we will discuss how to construct such MSAs in Section~\ref{sec:iterative}). This together with the $\pi$-irrelevance condition guarantees the identifiability of the nuisance functions $Q_{\phi}^{\pi}$, $V_{\phi}^{\pi}$ and $w_{\phi}^{\pi}$, meaning that they can be consistently estimated from the offline dataset. 
Below, we introduce a set of conditions to analyze the MSE of post-abstraction OPE estimators derived from these methods. 

\begin{asmp}[Boundedness]\label{asmp:bounded}
    The absolute value of all immediate rewards are uniformly bounded by a positive constant $R_{\mathrm{max}}<\infty$. \vspace{-0.5em}
\end{asmp}
\begin{asmp}[Coverage]\label{asmp:coverage}
    Both $\rho^{\pi}$ and $w^{\pi}$ are finite, i.e., $\rho^{\pi}(a,s),w^{\pi}(a,s)<\infty$ for any $a$ and $s$.\vspace{-0.5em}
\end{asmp}
\begin{asmp}[Stationary]\label{asmp:stationary}
    The state-action-reward triplets $(S_t,A_t,R_t)_{t\ge 1}$ from the ground MDP are stationary over time.\vspace{-0.5em}
\end{asmp}
\begin{asmp}[Nuisance function estimators]\label{VC}
(i) Both the estimated $Q$-function $\widehat{Q}_{\phi}^{\pi}$ and MIS ratio $\widehat{w}_{\phi}^{\pi}$ belong to VC-type classes (see e.g., Definition 2.1, \cite{chernozhukov2014gaussian}) with a finite VC index. (ii) These function classes are upper bounded by $R_{\mathrm{max}} /(1-\gamma)$ and $O(1)$, respectively. (iii) The estimation errors $\|\widehat{Q}_{\phi}^{\pi}-Q_{\phi}^{\pi}\|$ and $\|\widehat{w}_{\phi}^{\pi}-w_{\phi}^{\pi}\|$ (refer to Appendix~\ref{sec:proofs} for the detailed definitions of the norm $\|\bullet\|$) are upper bounded by $\kappa_q(\phi) (1-\gamma)^{-1} R_{\max}$ and $\kappa_w(\phi)$, respectively, for some $\kappa_q(\phi),\kappa_w(\phi)=O(1)$.\vspace{-0.5em}
\end{asmp}
We make a few remarks: (i) Assumptions \ref{asmp:bounded} and \ref{asmp:coverage} are widely adopted in the RL and OPE literature \citep[see, e.g.,][]{chen2019information,fan2020theoretical,uehara2020minimax,yin2020asymptotically}.
(ii) Assumption \ref{asmp:stationary} is introduced for convenience to ensure that the denominator of the MIS ratio (see \eqref{eqn:MISratio}) is well-defined. Notably, we do not require the state-action-reward-next-state tuples to be i.i.d., as in \citet{munos2008finite,sutton2008convergent,chen2019information,fan2020theoretical}, since the independence assumption is clearly violated in MDPs. Similarly, we do not impose any restrictive ergodicity or mixing conditions on the ground MDP, as in \citet{bhandari2018finite,zou2019finite,luckett2019estimating,shi2021deeply,chen2022well}. (iii) The VC-class type condition in Assumption \ref{VC} is commonly used in machine learning and statistics \citep[see, e.g.,][]{van1996weak,shalev2014understanding} to measure the complexity of a model class. The boundedness conditions on the Q-function and ratio function classes are also mild. Given Assumptions \eqref{asmp:bounded} and \eqref{asmp:coverage}, it follows immediately that the oracle Q-function and MIS ratio are bounded by $R_{\mathrm{max}} /(1-\gamma)$ and $O(1)$, respectively. It is thus reasonable to assume these function classes are bounded by these constants as well.

\begin{theorem}
[Bias-variance decomposition]\label{postmse}
    Assume Assumptions \ref{asmp:bounded} -- \ref{VC} hold. Then the followings hold for any $\pi$-irrelevant MSA $\phi$:
    \begin{itemize}[leftmargin=*]
        \item The MSE of \textbf{DRL} (Method 4) applied to the abstract MDP is given by
\begin{eqnarray}\label{eqn:DRMSE}
\begin{split}
&\frac{1}{N}\mathbb{V}\left[V_{\phi}^{\pi}\left(\phi(S)\right)\right]+\frac{1}{NT(1-\gamma)^2}\mathbb{V}\Big[w_{\phi}^{\pi}(A,\phi(S)) \big[R+\gamma V_{\phi}^{\pi}\left(\phi(S')\right)-Q_{\phi}^{\pi}(A,\phi(S))\big]\Big]\\
&
+O\Big(\frac{R_{\max}^2\max(\kappa_{q}(\phi),\kappa_w(\phi),N^{-1/2})\log (N)}{(1-\gamma)^2N}+\frac{R_{\max}^2\max(\kappa_{q}(\phi)\kappa_{w}(\phi), \sqrt{N}\kappa_{q}^2(\phi)\kappa_{w}^2(\phi))}{\sqrt{N}(1-\gamma)^{2}}\Big).
\end{split}
 \end{eqnarray}
Here, $\mathbb{V}[\bullet]$ denotes the variance of a random variable, and we recall that $N$ denotes the number of trajectories, whereas $T$ denotes the number of state-action-reward-next-state tuples per trajectory. 
    \item When both the Q-function and MIS ratio are estimated via linear function approximation (refer to Appendix~\ref{proofofmse} for the detailed estimation procedure), the MSEs of \textbf{Q-function-based estimator} (Method 1) and \textbf{MIS} (Method 3) applied to the abstract MDP equal to \eqref{eqn:DRMSE} as well. 
        \item When $\rho_{\phi}^{\pi}$ is known, the MSE of \textbf{SIS} (Method 2) applied to the abstract MDP is upper bounded by   
        \begin{align}\label{eqn:ISMSE}
            \frac{4R_{\max}^2}{NT^2(1-\gamma)^2}\sum_{t=1}^T \gamma^{2t-2}\mathbb{E}(\rho_{\phi,1:t}^{\pi})^2
+\frac{\gamma^{2T}R_{\max}^2}{(1-\gamma)^2}.
\end{align}
%where $\rho_{\phi,1:t}^{\pi}=\prod_{j=1}^t \rho_{\phi}^{\pi}(A_j,\phi(S_j))$. 
    \end{itemize}
\end{theorem}
At first glance, Theorem \ref{postmse} may appear complex, but both the theorem and its proof have several important implications, as we elaborate below:
\begin{enumerate}
    \item Equation \eqref{eqn:DRMSE} provides a bias-variance decomposition for the MSE of DRL. Specifically, in Equation \eqref{eqn:DRMSE}, \ul{\textit{the first line represents the asymptotic variance}} of DRL, which is of order $O(N^{-1})$, while \ul{\textit{the second line upper bounds their squared finite-sample bias}} arising from the estimation of the Q-function and MIS ratio. Notably, when the estimation errors satisfy $\kappa_q(\phi) = O(N^{-\ell_1})$ and $\kappa_w(\phi) = O(N^{-\ell_2})$ for $\ell_1, \ell_2 > 0$ and $\ell_1 + \ell_2 > 1/2$ -- conditions commonly imposed on doubly robust estimators for valid uncertainty quantification \citep[see, e.g.,][]{chernozhukov2018double,farrell2018deep,kallus2022efficiently} -- \ul{\textit{the bias term decays to zero at a much faster rate than the asymptotic variance term}} as $N$ increases to infinity. 
    \item When employing linear function approximation to model the nuisance functions, \ul{\textit{the resulting Q-function-based and MIS estimators become essentially equivalent to a DRL estimator}} (see Appendix~\ref{proofofmse} for a formal proof). As a result, the same bias-variance decomposition in Equation \eqref{eqn:DRMSE} applies to the Q-function-based and MIS estimators as well.
    \item Equation \eqref{eqn:ISMSE} provides a bias-variance decomposition for the MSE of SIS. Specifically, \ul{\textit{the two terms upper bound the variance and squared bias of SIS}}, respectively. Notably, the variance upper bound is proportional to the second moment of the product of the IS ratios, which reflects the curse of horizon. The bias term arises from truncating the data at time $T$, while the target policy to be evaluated involves an infinite sum. Similar to the other three estimators, \ul{\textit{the squared bias is typically much smaller than the variance}}. 
\end{enumerate}
Building on Theorem \ref{postmse}, we next investigate the impact of different abstractions on the bias and variance of the resulting OPE estimators. As shown in Equation \eqref{eqn:DRMSE}, the bias term depends on the estimation errors of the Q-function and MIS ratio. To simplify the analysis, we use lookup tables to parameterize the Q-function and MIS ratio. For any abstraction $\phi$, let $\textrm{Var}^{(1)}(\phi)$ and $\textrm{Var}^{(2)}(\phi)$ denote the variance terms in Equations \eqref{eqn:DRMSE} and \eqref{eqn:ISMSE}, respectively. The following theorem summarizes our findings.
\begin{theorem}[Bias, variance and abstraction: The role of $\phi$ in OPE]\label{Tpostmse}Let $\phi$ denote an MSA from $\mathcal{S}$ to $\mathcal{X}$ and $\phi^*$ denote another MSA from $\mathcal{X}$ to certain $\mathcal{X}^*=\{\phi^*(x):x\in \mathcal{X}\}$. Then we have:
 \begin{itemize}
     \item $\displaystyle\lim_{T\to \infty} \textrm{Var}^{(1)}(\phi)\ge \lim_{T\to \infty} \textrm{Var}^{(1)}(\phi^*\circ \phi)$ and the equality holds if and only if $V^{\pi}_{\phi}$ is $\phi^*$-irrelevant, i.e., for any $x_1,x_2\in \mathcal{X}$, $V_{\phi}^{\pi}(x_1)=V_{\phi}^{\pi}(x_2)$ whenever $\phi^*(x_1)=\phi^*(x_2)$. 
     \item When lookup tables are employed to parameterize the Q-function and MIS ratio, under certain matrix invertibility condition detailed in Appendix~\ref{sec:proofs}, we have
     \begin{eqnarray*}
         \kappa_q(\phi)=O\Big(\frac{{|\mathcal{X}||\mathcal{A}|}}{\sqrt{NT}}\Big)\,\,\hbox{and}\,\,\kappa_w(\phi)=O\Big(\frac{{|\mathcal{X}||\mathcal{A}|}}{\sqrt{N}}\Big).
     \end{eqnarray*}
     \item $\textrm{Var}^{(2)}(\phi)\ge \textrm{Var}^{(2)}(\phi^*\circ \phi)$ and the equality holds if and only if $\rho_{\phi}^{\pi}$ is $\phi^*$-irrelevant. 
     \item The bias term in \eqref{eqn:ISMSE} is a constant function of $\phi$. 
 \end{itemize}
\end{theorem}
Given that $\phi$ and $\phi^*\circ\phi$ are two nested abstractions, we summarize the implications of Theorem \ref{Tpostmse} as follows:
\begin{enumerate}
\item Results in the first bullet point suggest that \ul{\textit{the asymptotic variances of the Q-function-based estimator, MIS, and DRL decay with more refined abstractions for sufficiently large $T$}}.
\item Results in the second bullet point imply that the Q-function and MIS estimation errors generally decrease with the size of the state space. Combined with Equation \eqref{eqn:DRMSE}, this suggests that \ul{\textit{the finite-sample biases of the Q-function-based estimator, MIS, and DRL also decay with more refined abstractions}}.
\item Results in the third bullet point suggest that \ul{\textit{the asymptotic variance of SIS decays with more refined abstractions}}.
\item Results in the last bullet point indicate that \ul{\textit{the bias of SIS is independent of the abstraction}}.
\end{enumerate}
In summary, for all the four OPE estimators, \ul{\textit{their MSEs generally decrease with more refined abstractions}}. However, it is important to note that all the aforementioned results apply only to MSAs. When the abstraction fails to preserve the Markov property, the resulting OPE estimator will suffer from a large bias. The aforementioned discussions motivate us to identify the finest MSA for minimizing the resulting OPE estimator's MSE. However, direct identification of such an MSA can be challenging. Instead, we propose an iterative state abstraction procedure, which identifies a sequence of MSAs $\phi_1$, $\phi_2$, $\cdots$, $\phi_K$ to progressive reduce the state space. Based on our theories, we expect the MSE of the resulting OPE estimator decreases with the number of iterations, i.e.,
\begin{eqnarray*}
\textrm{MSE}(\phi_1)\ge \textrm{MSE}(\phi_2\circ \phi_1) \ge \cdots \ge \textrm{MSE}(\phi_K\circ \cdots \circ \phi_1),
\end{eqnarray*}
where the inequalities are likely to hold when these abstractions are not identical mappings. Thus, given a sufficiently large $K$, the accuracy of the OPE estimator can be substantially improved. We detail our methodology in the next section.
%Corollary~\ref{Tpostmse} indicates that by reducing the dimension of \(|\mathcal{\phi(S)}|\), we can effectively reduce the bias term of the MSE. This motivates us to perform an OPE on the abstract state space to improve accuracy. The following theorem will provide the theoretical foundation.

\subsection{Iterative state abstraction}%forward-type-II MSA}
\label{sec:iterative}
\textbf{Summary of our approach}. We begin with an overview of the proposed iterative procedure. As outlined earlier, our goal is to construct a sequence of abstractions $\phi_1, \phi_2, \dots, \phi_K$ to progressively compress the state space. Starting from $\mathcal{X}_0 = \mathcal{S}$,  the original state space, each $\phi_k$ maps from $\mathcal{X}_{k-1}$ to $\mathcal{X}_k$. The final abstraction is obtained by composing the sequence as $\phi_K \circ \cdots \circ \phi_2 \circ \phi_1$, yielding a deeply-abstracted state space.

Based on our theoretical framework, these abstractions must satisfy the following requirements:
\begin{enumerate}
    \item Each $\phi_k$ must be a $\pi$-irrelevant \textbf{MSA} to guarantee the bias-variance decompositions in Equations \eqref{eqn:DRMSE} and \eqref{eqn:ISMSE} hold.
    \item Any two adjacent abstractions must employ \textbf{different} abstraction methods; otherwise, their composition may result in the same mapping that fails to reduce the state space. 
\end{enumerate}
To simultaneously satisfy both requirements, we observe that there are two types of MSAs available:
\begin{itemize}
\item[(i)] The first is based on the notion of model-irrelevance \citep[see e.g.,][]{li2006towards}; see also, Definition~ \ref{def:model} below.
\item[(ii)] The second is the abstraction developed by \citet{allen2021learning}.
\end{itemize}

\begin{figure}[t]
    \centering    \includegraphics[width=9cm,height=3cm]{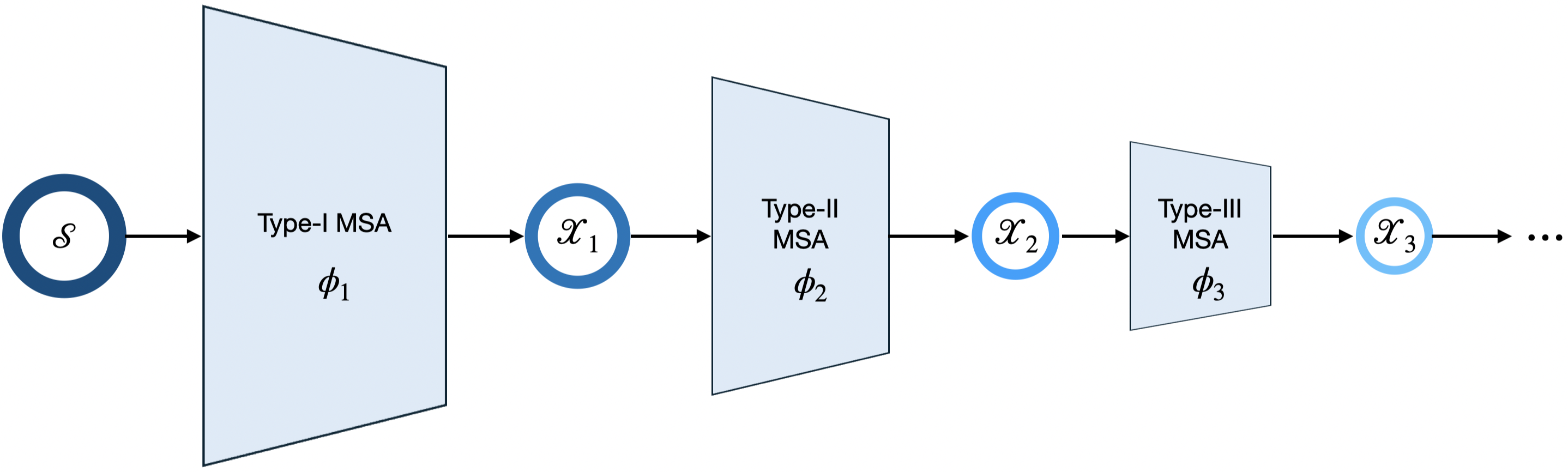}
    \caption{Illustrations of the iterative procedure. %and (b) an MDP with three groups of state variables, denoted by $\{S_{t,1}\}_t$, $\{S_{t,2}\}_t$ and $\{S_{t,3}\}_t$. 
    }\label{fig:iterative}
\end{figure}

This motivates us to alternate between these two abstractions at each iteration of the procedure (see Figure \ref{fig:iterative} for an illustration). For both types of MSAs, we make necessary adjustments to adapt them to our setup. For instance, we tailor them to be $\pi$-irrelevant by concatenating them with $\{\pi(a|\bullet):a\in \mathcal{A}\}$ and extend \citet{allen2021learning}'s abstraction by relaxing their requirement for the behavior policy to be Markovian —- an assumption likely violated due to the iterative nature of our procedure. We emphasize that the abstractions employed in each iteration of our procedure are not entirely new. Our primary methodological contribution lies in the \textbf{iterative use} of these abstraction, which, guided by our theoretical framework, can significantly enhance the performance of OPE. We next detail the two different types of MSAs used in our iterative procedure. 

\textbf{Type-I MSA}. We present model-irrelevance, a widely studied condition in the state abstraction literature, designed for learning the optimal policy  $\pi^*$. To better understand this condition, we begin by reviewing some basic definitions of state abstraction for policy learning; see also 
\cite{jiang2018notes}.

\begin{Def}
[$\pi^*$-irrelevance]\label{def:pistar}$\phi$ is $\pi^*$-irrelevant if there exists an optimal policy $\pi^*$, such that for any $s^{(1)}$, $s^{(2)}\in \mS$ whenever $\phi(s^{(1)})=\phi(s^{(2)})$, we have 
$\pi^*(a|s^{(1)})=\pi^*(a|s^{(2)})$ for any $a\in \mA$.
\end{Def}
\begin{Def}[$Q^{\pi^*}$-irrelevance]\label{def:Qstar} $\phi$ is $Q^{\pi^*}$-irrelevant if for any $s^{(1)}$, $s^{(2)}\in \mS$ whenever $\phi(s^{(1)})=\phi(s^{(2)})$, the optimal Q-function satisfies $Q^{\pi^*}(a,s^{(1)})=Q^{\pi^*}(a,s^{(2)})$ for any $a\in \mathcal{A}$. 
\end{Def}

Definitions \ref{def:pistar} and \ref{def:Qstar} are easy to understand, requiring the optimal policy/Q-function to depend on a state $s$ only through its abstraction $\phi(s)$. However, the resulting $\phi$ may not necessarily preserve the Markov property. This leads  to the following model-irrelevance, which preserves the MDP structure while ensuring $\pi^*$- and $Q^*$-irrelevance \citep{li2006towards}. 

\begin{Def}[Model-irrelevance]\label{def:model}
    $\phi$ is model-irrelevant if for any $s^{(1)}$, $s^{(2)}\in \mS$ whenever $\phi(s^{(1)})=\phi(s^{(2)})$, the following holds for any $a\in \mA$, $s'\in \mathcal{S}$ and $x'\in\mathcal{X}$:
\begin{eqnarray}\label{eqn:model-irrelevant}
        \mathcal{R}(a,s^{(1)})=\mathcal{R}(a,s^{(2)})\,\,\,\hbox{and}\,\,\sum_{s'\in \phi^{-1}(x')}\mathcal{T}(s'|a,s^{(1)})=\sum_{s'\in \phi^{-1}(x')}\mathcal{T}(s'|a,s^{(2)}).
    \end{eqnarray}    
\end{Def}
The first condition in \eqref{eqn:model-irrelevant} corresponds to ``reward-irrelevance'' whereas the second condition represents ``transition-irrelevance''. Notice that the term $\sum_{s'\in \phi^{-1}(x')}\mathcal{T}(s'|a,s)$ -- appearing in the second equation of \eqref{eqn:model-irrelevant} -- represents the probability of transitioning to $\phi(S')=x'$ in the abstract state space. 
Thus, the second condition essentially requires %the transition to 
the abstract next state $\phi(S')$ to be conditionally independent of $S$ given $A$ and $\phi(S)$. Assuming $S$ can be decomposed into the union of relevant features and irrelevant features. The condition implies that 
the evolution of those relevant features depends solely on themselves, independent of those irrelevant features. This ensures that $\phi$ is an MSA, as shown in the following lemma.
\begin{lemma}
    The abstraction derived from model-irrelevance \eqref{def:model} is an MSA. 
\end{lemma}
To compute such a type-I MSA, we first adapt existing algorithms \citep{ha2018recurrent,franccois2019combined,gelada2019deepmdp} to train a model-irrelevant abstraction $\phi$, parameterized via deep neural networks. We next combine $\phi(s)$ with $\{\pi(a|s): a\in \mathcal{A}\}$ to ensure $\pi$-irrelevance. Refer to Appendix \ref{sec:implementforward} for the detailed procedures. 

\begin{figure}[t]
    \centering
\includegraphics[width=5.5cm]{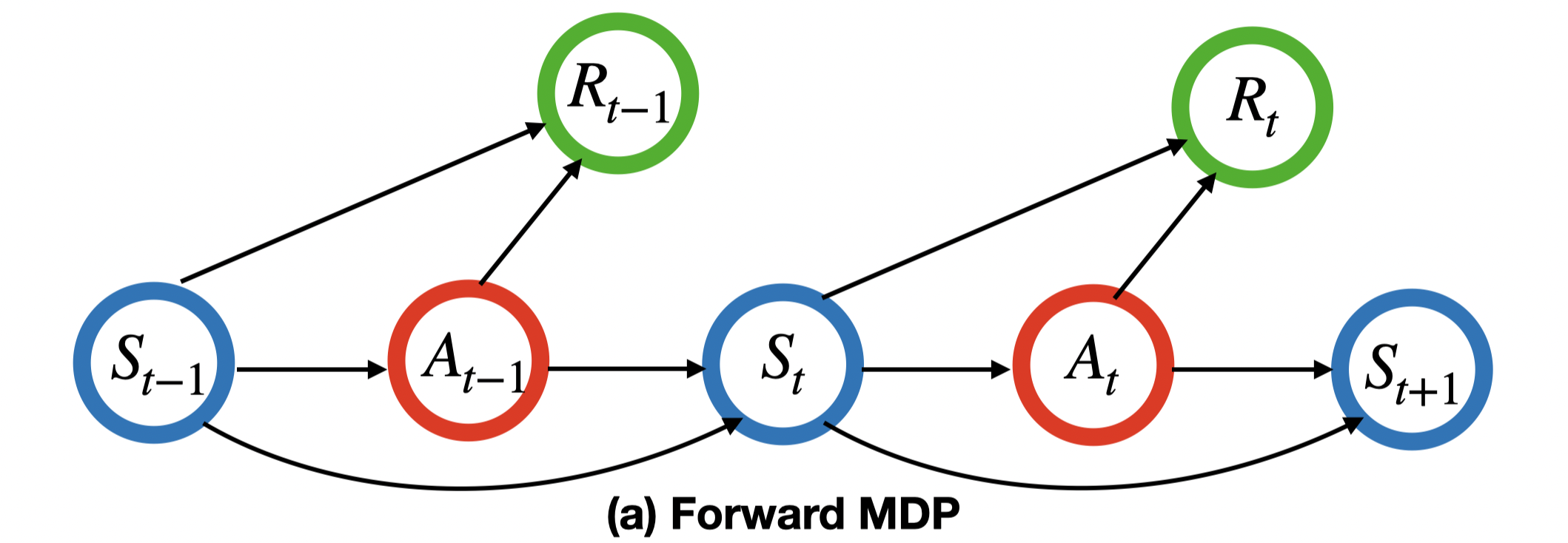}
\hspace{1cm}
 \includegraphics[width=4.5cm]{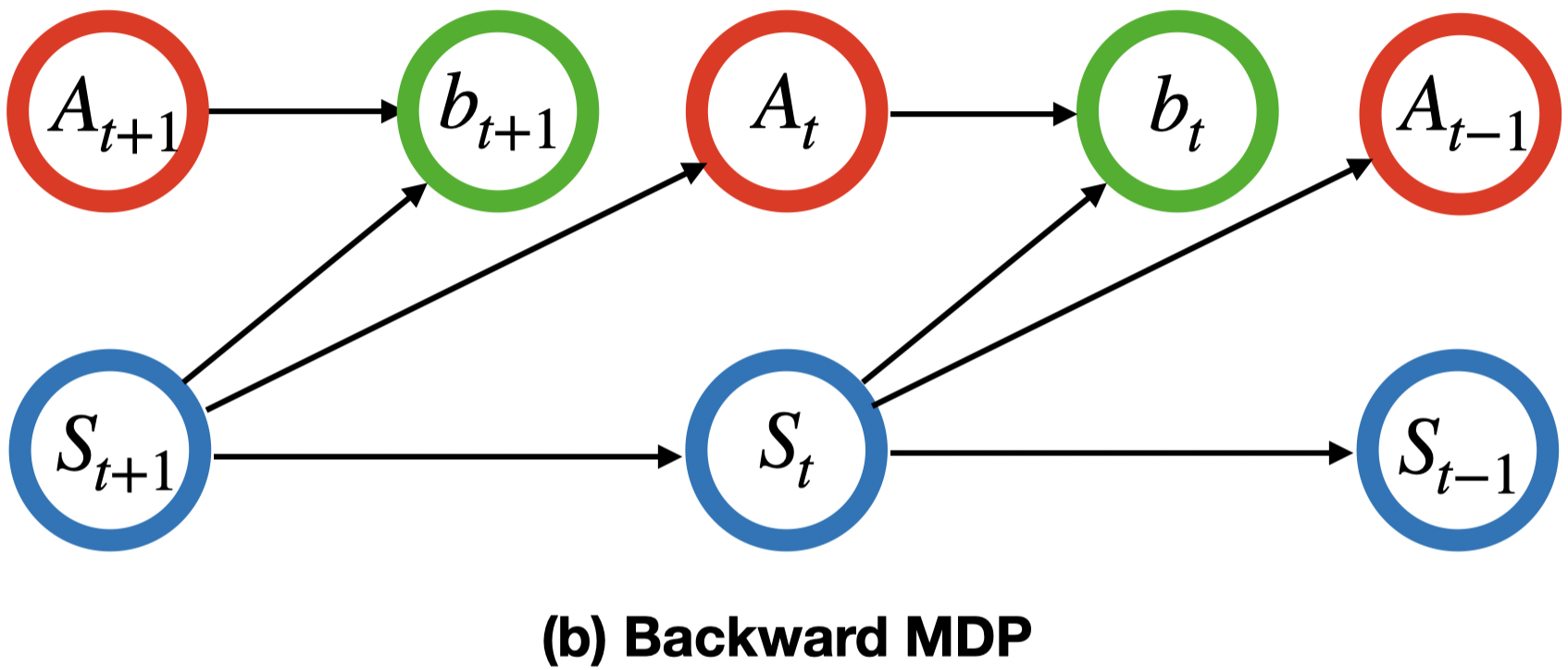}
    \caption{Illustrations of (a) the forward MDP model  and (b) the backward MDP model. $b_t$ is a shorthand for $b(A_t|S_t)$ for any $t\ge 1$.}\label{fig:twoviews} 
\end{figure}

\textbf{Type-II MSA}. To illustrate the second type of MSA, we begin by discussing two perspectives of the data generated the MDP framework (see Figure \ref{fig:twoviews} for graphical illustrations of both perspectives):

\begin{enumerate}[leftmargin=*]
\item The first perspective is the traditional (forward) MDP model with all state-action-reward triplets sequenced by time index. This yields the model-based irrelevance condition defined in Definition \ref{def:model}. 
\item The second perspective offers a backward view by reversing the time order. Specifically, 
due to the symmetric nature of the Markov assumption — implying that if the future is independent of the past given the present, the past must also be independent of the future given the present — the reversed state-action pairs also maintain the Markov property. Leveraging this property, we define another backward MDP model below, which motivates the second type of MSA. 
\end{enumerate}
To elaborate, we assume for now that the behavior policy $b$ is not history-dependent. We will relax this condition later. Under the MDP model assumption, actions and states following $S_t$ are independent of those occurred prior to the realization of $S_t$. Accordingly, $(S_{t-1},A_{t-1})$ is conditionally independent of $\{(S_k,A_k)\}_{k>t}$ given $S_t$. Recall that $T$ corresponds to the termination time of trajectories in the offline dataset. We define a time-reversed MDP consisting of state-action-reward triplets $\{(S_t,A_t,b(A_t|S_t)): t=T,\dots,1\}$. Its dynamics is described as follows (see also Figure \ref{fig:twoviews}(b) for the configuration):
\begin{itemize}[leftmargin=*]
    \item \textbf{State-action transition}: Due to the aforementioned Markov property, the transition of the past state $S_{t+1}$ in the reversed process (future state in the original process) into the current state $S_{t}$ is independent of the past action $A_{t+1}$ in the reversed process (future action in the original process) while the behavior policy that generates $A_t$ depends on both the current state $S_t$ and the past state $S_{t+1}$ in the reversed process. This yields the time-reversed state-action transition function $\mathbb{P}(A_t=a,S_t=s|S_{t+1})$.
    \item \textbf{Reward generation}: For each state-action pair $(S_{t}, A_{t})$, we manually set the reward to the behavior policy $b(A_{t}|S_{t})$, which plays a crucial role in constructing IS estimators.
\end{itemize}   
Given this MDP, the second type of MSA is readily available by applying the model-irrelevance condition to the reversed process. Analogous to Definition \ref{def:model}, our objective is to identify a state abstraction that is crucial for predicting the reward (e.g., the behavior policy) and the reversed transition function. We refer to the resulting condition as backward-model-irrelevance (see the formal definition below), since it is derived from the backward MDP model. 
\begin{Def}[Backward-model-irrelevance]\label{def:backwardmodel}
    $\phi$ is backward-model-irrelevant if %the followings hold
    for any $s^{(1)},s^{(2)} \in \mathcal{S}$ whenever $\phi(s^{(1)})=\phi(s^{(2)})$, the followings hold for any $a \in\mathcal{A}$ , $x \in \mathcal{X}$ and $t \in \mathbb{N}^+$: 
    %$(i) \rho^{\pi}(a,s^{(1)})=\rho^{\pi}(a,s^{(2)})$;
    \begin{align}
        &(i)~b(a|s^{(1)})=b(a|s^{(2)});\label{eqn:backirrelevance}\\ 
        &(ii) \sum_{s\in \phi^{-1}(x)}\mathbb{P}(A_t=a,S_t=s|S_{t+1}=s^{(1)})=\sum_{s\in \phi^{-1}(x)}\mathbb{P}(A_t=a,S_t=s|S_{t+1}=s^{(2)}).\label{eqn:backirrelevance-state}
    \end{align}
\end{Def}

The conditions of backward-model-irrelevance are similar to those specified for model-irrelevance outlined in Definition \ref{def:model}. Equation \eqref{eqn:backirrelevance} essentially requires behavior-policy-irrelevance, or reward-irrelevance in the backward MDP, whereas Equation \eqref{eqn:backirrelevance-state} corresponds to the ``backward-transition-irrelevance'', and is equivalent to %$(A_t,\phi(S_t))\indep S_{t+1}|\phi(S_{t+1})$, 
the conditional independence assumption between the pair $(A_t, \phi(S_t))$ and $S_{t+1}$ given $\phi(S_{t+1})$. 

Next, we draw a connection between the abstraction derived from the backward-model-irrelevant condition and that developed by \citet{allen2021learning} for policy learning. The latter imposes two conditions: (i) inverse-model-irrelevance, which requires $A_t$ to be conditionally independent of $S_t$ and $S_{t+1}$ given $\phi(S_t)$ and $\phi(S_{t+1})$; (ii) density-ratio-irrelevance, which requires $\phi(S_t)$ to be conditionally independent of $S_{t+1}$ given $\phi(S_{t+1})$. For effective policy learning, \citet{allen2021learning} requires both conditions to hold in data generating processes following a diverse range of behavior policies. When restricting them to one behavior policy, (i) and (ii) imply the backward-transition-irrelevance condition in \eqref{eqn:backirrelevance-state} whereas backward-transition-irrelevance in turn yields (ii). This allows us to adapt their algorithm to train our type-II MSA that satisfies backward-model-irrelevance; see Appendix \ref{sec:implementbackward} for details. 

As mentioned earlier, due to the iterative nature of our procedure, after type-I MSA, the behavior policy when restricted to the abstract space can be history-dependent. To address this challenge, we modify the proposed backward-model-irrelevance by employing a history-dependent behavior policy 
$
    b_t(a_t|s_t,a_{t-1},s_{t-1},\cdots,s_1)=\mathbb{P}(A_t=a_t|S_t=s_t,A_{t-1}=a_{t-1},S_{t-1}=s_{t-1},\cdots,S_1=s_1)
$
in \eqref{eqn:backirrelevance}. Specifically, we require that for any two state sequences $\{s_\ell^{(1)}\}_\ell$ ,$\{s_\ell^{(2)}\}_\ell$ such that $\phi(s_\ell^{(1)})=\phi(s_\ell^{(2)})$, any action sequence $\{a_\ell\}_\ell$ and any time index $t\ge 1$, 
\begin{eqnarray}\label{eqn:historydependentbehaviorpolicy}
    b_t(a_t|s_t^{(1)},a_{t-1},s_{t-1}^{(1)},\cdots,s_1^{(1)})=b_t(a_t|s_t^{(2)},a_{t-1},s_{t-1}^{(2)},\cdots,s_1^{(2)}).
\end{eqnarray}
The following lemma guarantees that the resulting abstraction \(\phi\), obtained by applying the refined backward-model-irrelevance condition to the original MDP \((S_t, A_t, R_t)_{t \geq 1}\) with a history-dependent behavior policy, remains an MSA. 
\begin{lemma}\label{backMSA}
The abstraction \(\phi\) derived from \eqref{eqn:backirrelevance-state} and \eqref{eqn:historydependentbehaviorpolicy} is an MSA, regardless of whether the behavior policy is history-dependent or not.
\end{lemma} 
Finally, given an abstraction derived from \eqref{eqn:backirrelevance-state} and \eqref{eqn:historydependentbehaviorpolicy}, we similarly augment it with $\{\pi(a|\bullet): a\in \mathcal{A}\}$ to ensure $\pi$-irrelevance.

\textbf{Summary}. To summarize, we have presented the two types of MSAs in this section. By iteratively alternating between these abstractions, the state space becomes progressively finer. As a result of our theory, the MSE of the post-abstraction OPE estimators is expected to decrease significantly. To conclude this section, we note that the initialization of the iterative procedure doesn't necessarily have to begin with type-I MSA; type-II MSA can serve as the starting point as well. In our experiments, both starting points have their merits, with their effectiveness varying depending on the environment. However, the overall differences in results are small.

\section{Numerical experiments}\label{sec:experiments} 
We investigate the finite sample performance of our proposal in this section and detail its implementation in Appendix \ref{sec:implement}. 

\textbf{Comparisons}. We compare the proposed iterative procedure, which yields a deeply abstracted state space (denoted by `DSA', short for deep state abstraction), against the following baselines: single-iteration type-I MSA (`Type-I'), type-II MSA (`Type-II'), the abstract reward process \citep[`STAR']{chaudhari2024abstract}, the MSA proposed by \citet[`MSA']{allen2021learning},  and a reconstruction-based abstraction \citep[`Auto-Encoder']{lange2010deep}. For a fair comparison, except for STAR, which uses a built-in IS estimator for OPE, the performance of other abstractions is evaluated by applying a base FQE algorithm \citep{le2019batch} to the abstract state space. We also report the performance of FQE applied to the original ground state space (`FQE'). This yields a total of six baselines, where:
\begin{itemize}
    \item \textbf{Type-I} and \textbf{Type-II} are variants of the proposed procedure;
    \item \textbf{FQE} is a popular OPE method without any abstraction;
    \item \textbf{STAR} is the state-of-the-art abstraction designed for OPE;
    \item \textbf{MSA} and \textbf{Auto-Encoder} are state-of-the-art abstractions for policy learning. 
\end{itemize}
The behavior policy is $\epsilon$-greedy with respect to the target policy, with $\epsilon=0.1, 0.3, 0.5$. The simulation results presented are averaged over 30 trials. 

\textbf{Environments}. We consider four environments, including two complex games — ``Atlantis-ram-v5" and ``SpaceInvaders-ram-v5" — a medium-to-high difficulty environment, ``LunarLander-v2", and a simple control problem, ``Cartpole-v1". For ``LunarLander-v2" and ``Cartpole-v1", we manually introduce 292 and 296 irrelevant variables into the state, respectively, resulting in a challenging 300-dimensional system. Additional details about these environments can be found in Appendix~\ref{sec:experiments-details}.

\begin{figure}[h]
    \centering
    \includegraphics[width=0.85\columnwidth,height=4.5cm]{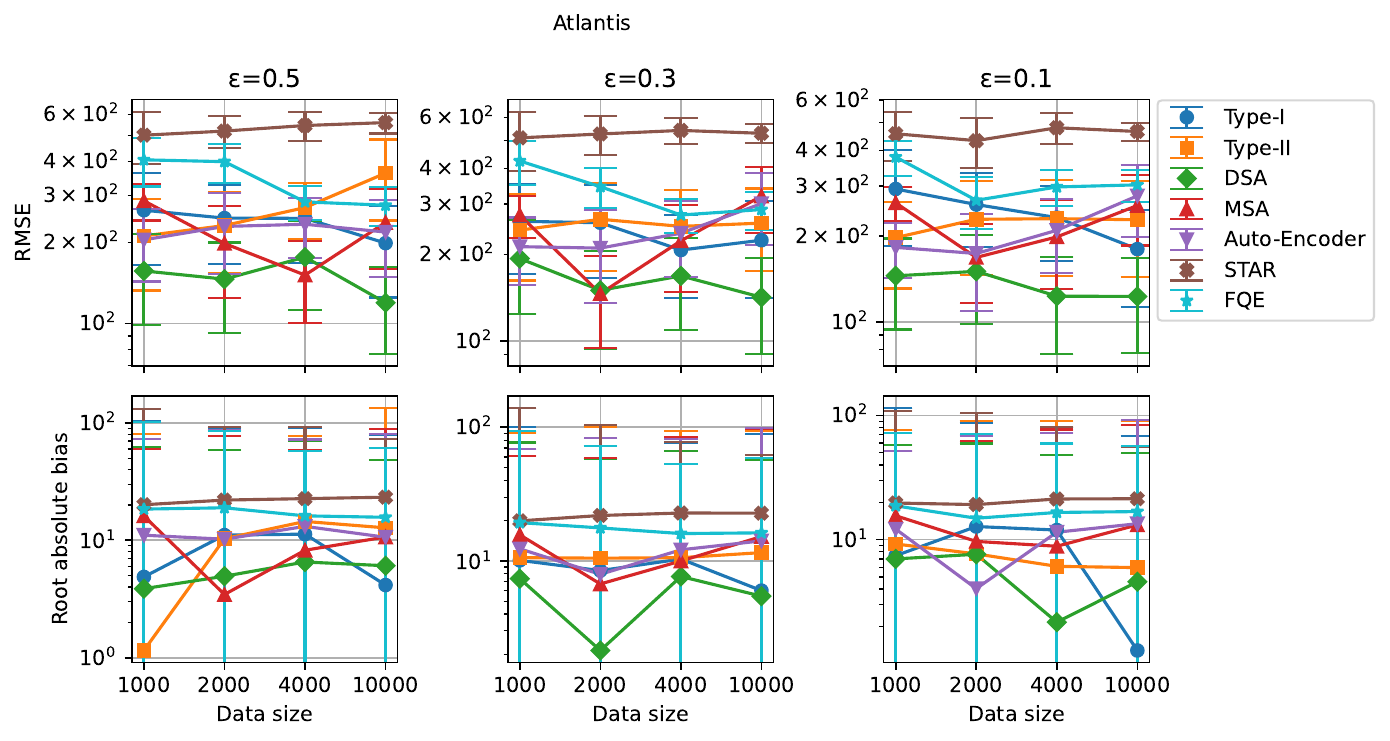}
 \vspace{-0.1cm}
 \includegraphics[width=0.85\columnwidth,height=4.5cm]{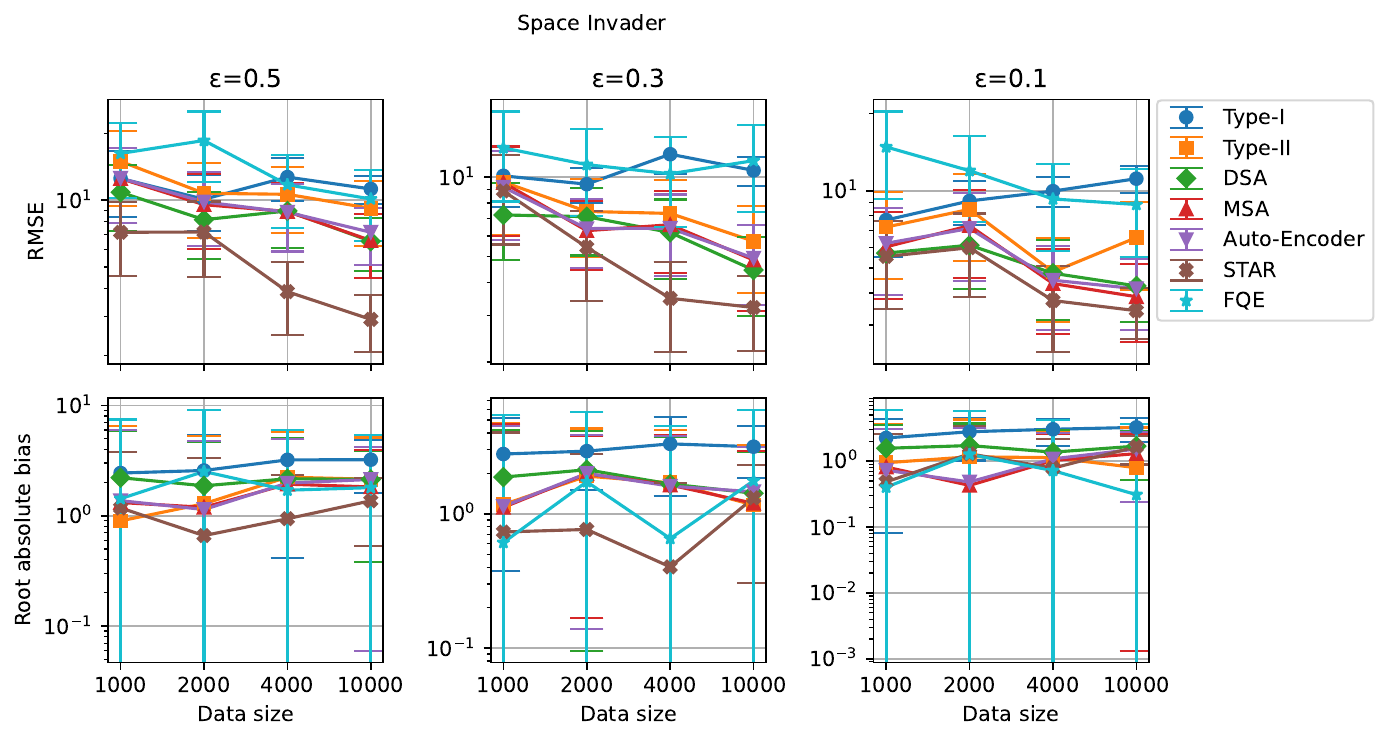}
 \includegraphics[width=0.85\columnwidth,height=4.5cm]{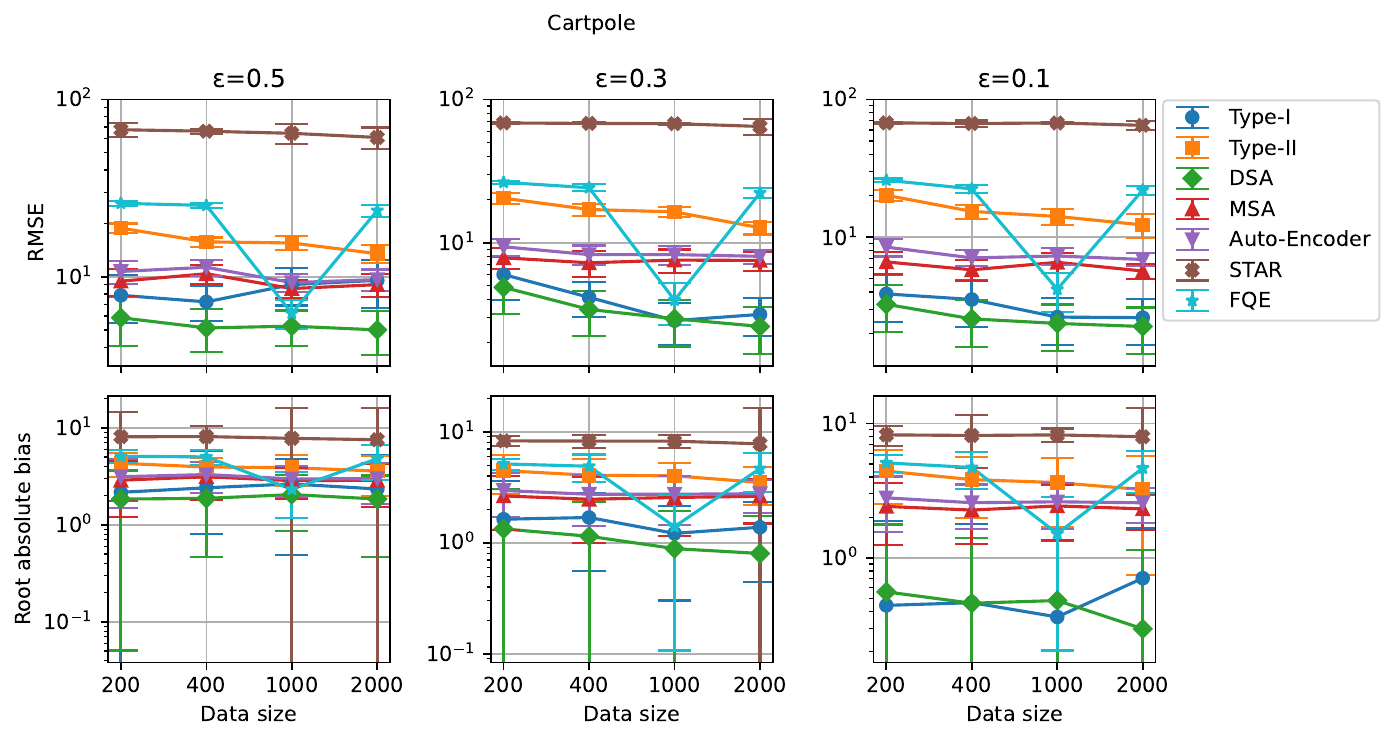}
 \includegraphics[width=0.85\columnwidth,height=4.5cm]{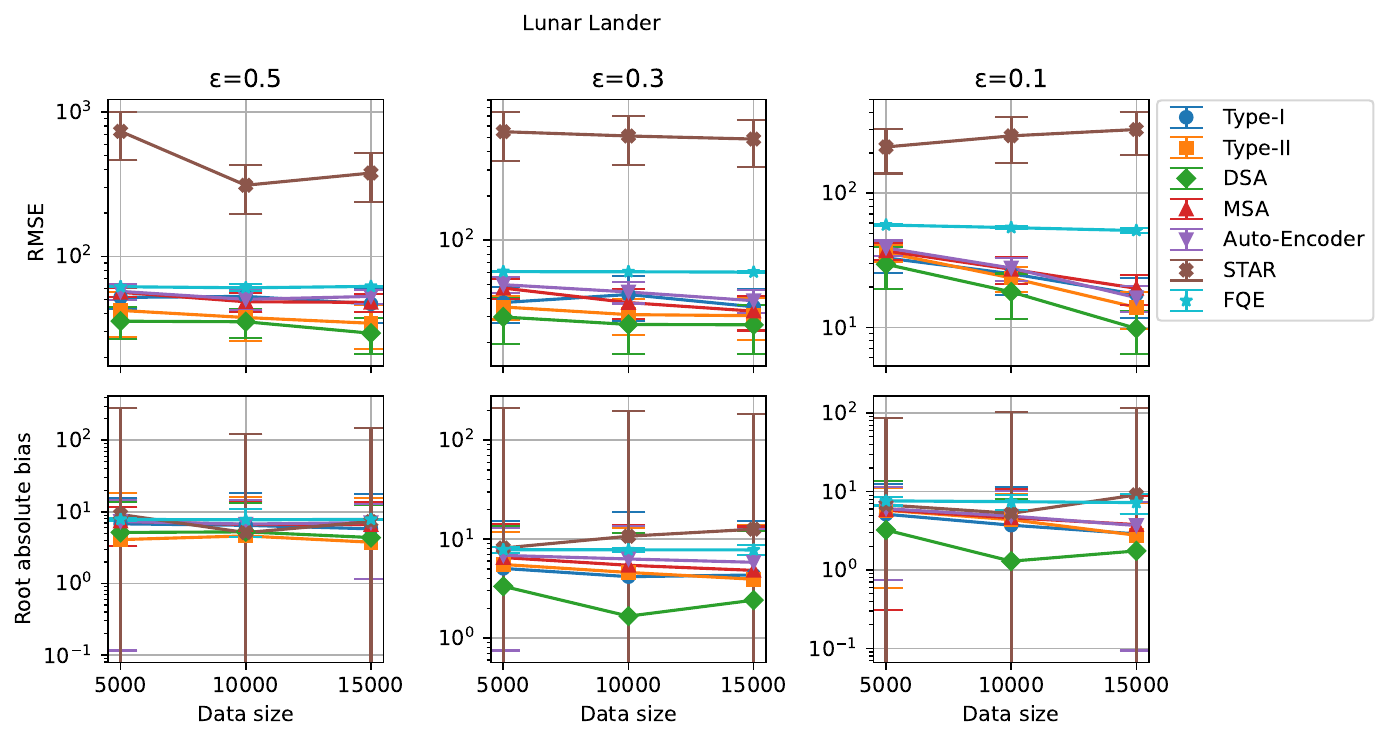}
    \caption{RMSEs and root absolute biases of various estimators for 4 different environments.}\label{all_figs}
\end{figure}

\textbf{Results}. We report the root MSE (RMSE) and root absolute bias of different post-abstraction OPE estimators, as well as the baseline FQE estimator without abstraction, in Figure~\ref{all_figs}. The results demonstrate that the proposed DSA generally outperforms other baseline methods, achieving the smallest RMSE in most cases across "Atlantis-ram-v5", "LunarLander-v2", and "Cartpole-v1". For "SpaceInvaders-ram-v5", DSA achieves the second smallest RMSE, while STAR performs the best in this environment. However, we note that STAR's performance is highly unstable, as it performs the worst in the other three environments. 

To conclude, our analysis answers the following questions: 
\begin{enumerate}[leftmargin=*]
\item \textbf{Is state abstraction useful for OPE?} 
Figure~\ref{all_figs} shows that the baseline FQE applied to the ground state space performs worse than all post-abstraction FQE estimators. This comparison highlights the utility of state abstractions for OPE.
\item \textbf{Is the iterative procedure more effective than single-iteration procedures?} 
Comparisons against single-iteration abstractions (e.g., `Type-I' and `Type-II') demonstrate the advantages of multi-iteration procedures. This formally validates our theoretical finding that finer Markov state abstractions lead to lower MSE in the resulting OPE estimators.
\item \textbf{Is the proposed abstraction better than other abstractions for OPE?} 
In general, yes. The proposed DSA consistently achieves the best or second best performance across environments.
\end{enumerate}

\vspace{-1em}

\section{Conclusion}\label{sec:conclusion}
\vspace{-1em}

 This paper develops an iterative procedure for state abstraction in OPE. Our theory suggests that finer Markov state abstractions generally lead to more accurate OPE estimators. Building on this insight, we propose an iterative procedure to maximally compress the state space while preserving the Markov property. Empirical investigations of our proposal's performance confirm our theoretical findings.

There are certain limitations to our proposal: First, our theoretical analysis does not account for the estimation error incurred during the construction of the abstractions. Second, determining the optimal number of iterations, $K$, remains an open question. This choice involves a trade-off: a larger $K$ yields a smaller abstracted state space, but the estimation error for learning the abstraction may accumulate over iterations. In complex environments such as Atlantis-ram-v5 and SpaceInvaders-ram-v5, we set $K=4$ and observe good empirical performance. However, it remains unclear how to data-adaptively determine this hyperparameter.

\newpage

\bibliography{references.bib}
\bibliographystyle{plainnat}
%\beginSupplementaryMaterials
\appendix
\counterwithin{figure}{section}
\counterwithin{table}{section}
\counterwithin{equation}{section}
\counterwithin{algorithm}{section}
\counterwithin{lemma}{section}
\section*{Appendix}
\setlength{\parskip}{\baselineskip}
\setlength{\parindent}{2pt}

This appendix is structured as follows: Section \ref{sec:confounder} introduces additional related works on confounder selection in causal inference. The implementation details of the proposed  state abstraction are discussed in Section \ref{sec:implement}. Additional information concerning the environment and computing resources utilized is presented in Section \ref{sec:experiments-details}.  All technical proofs can be found in Section \ref{sec:proofs}. 
Toy examples demonstrating the advantages of DSA are presented in Section~\ref{sec:toy}, while additional conditions and properties of the proposed MSAs are discussed in Section~\ref{sec:add-properties}.

\section{Confounder selection  in causal inference}\label{sec:confounder}

Broadly speaking, confounding refers to the problem that even if two variables are not causes of each other, they may exhibit statistical association due to common causes. Controlling for confounding is a central problem in the design of observational studies, and many criteria for confounder selection have been proposed in the literature. A commonly adopted criterion is the ``common cause heuristic'', where the user only controls for covariates that are related to both the treatment and the outcome \citep{glymour2008methodological,austin2011introduction,shortreed2017outcome,koch2020variable}. Another widely used criterion is to simply use all covariates that are observed before the treatment in time \citep{rubin2009should,hernan2010causal,hernan2016using}. However, both of these approaches are not guaranteed to find a set of covariates that are sufficient to control for confounding. From a graphical perspective, confounder selection is essentially about finding a set of covariates that block all ``back-door'' paths \citep{Pearl09}, but this requires full structural knowledge about the causal relationship between the variables which is often not possible. This motivated some methods that only require partial structural knowledge \citep{vander2011new,vanderweele2019principles,guo2023confounder}. All the aforementioned methods need substantive knowledge about the treatment, outcome, and covariates. Other
methods use statistical tests (usually of conditional independence) to trim a set of covariates that are assumed to control for confounding \citep{robins1997causal,greenland1999confounding,hernan2010causal,de2011covariate,belloni2014inference,persson2017data}.  
%More recently, an emerging literature considers the problem of selecting the optimal set of covariates to control for that maximizes the statistical efficiency in estimating the causal effect of interest \citep{kuroki2003covariate,rotnitzky2020efficient,henckel2022graphical,guo2023variable}. 
The reader is referred to \cite{guo2022confounder} for a recent survey of objectives and approaches for confounder selection. 

Confounder selection can be considered as a special example of our problem under certain conditions: (i) The state transition is independent, effectively transforming the MDP into a contextual bandit; (ii) The action space is binary, with the target policy consistently assigning either action 0 or action 1, aimed at assessing the average treatment effect; (iii) State abstractions are confined to variable selections. While our proposed iterative procedure shares similar spirits with the aforementioned algorithms, it addresses a more complex problem involving state transitions. Additionally, our focus is on abstraction that facilitates the engineering of new feature vectors, rather than merely selecting a subset of existing ones.

\section{Implementation details}\label{sec:implement}

In this section, we present implementation details for type-I MSA (Section~\ref{sec:implementforward}) and type-II MSA  (Section~\ref{sec:implementbackward}). %We next provide a pseudocode for summarizing the proposed iterative algorithm in Section~\ref{sec:iterative}.

\subsection{Implementation details for type-I MSA}\label{sec:implementforward} %The objective loss function}
We provide details for implementing the proposed type-I MSA in this subsection. We use deep neural networks to parameterize the type-I MSA and estimate the parameters by minimzing the following loss function:
\begin{eqnarray}\label{forward_loss}
    %\mathcal{L}_1(\theta_r, \theta_\mathcal{T}, \theta_Q) = 
    \alpha_1 \mathcal{L}_r + \beta_1 \mathcal{L}_\mathcal{T}+\delta_1 \mathcal{L}_Q+ \lambda_1\mathcal{L}_{penalty},
\end{eqnarray}
where %$\theta_r, \theta_\mathcal{T}, \theta_Q$ are the parameters in the network, 
$\mathcal{L}_r$, $\mathcal{L}_\mathcal{T}$ and $\mathcal{L}_Q$ are the loss functions detailed below, $\mathcal{L}_{penalty}$ is a penalty term, and $\alpha_1, \beta_1, \delta_1, \lambda_1$ are positive constant hyper-parameters whose values are reported in Table \ref{tab:hyper_rf}. 

By definition, the type-I MSA is required to achieve both model-irrelevance and $\pi$-irrelevance. As discussed in Section \ref{sec:iterative}, our approach is to learn a model-irrelevant abstraction, denoted as $\phi$, and then concatenate it with $\{\pi(a|\bullet): a\in \mathcal{A}\}$. We denote the concatenated abstraction by $\phi_{for}$.

We next detail the loss functions and the penalty term. 
%Given $\mathcal{D}$, the objectives of learning the type-I MSA $\phi_1$  are:
The first two losses $\mathcal{L}_r$ and $\mathcal{L}_\mathcal{T}$ are to ensure reward-irrelevance and transition-irrelevance, respectively,
\begin{align*}
    \mathcal{L}_r =  \frac{1}{|\mathcal{D}|}\sum_{(S,A,R)\in \mathcal{D}} \big[R - \mathcal{R}_{\phi}\big(A,\phi(S)\big)\big]^2,\,\,
\mathcal{L}_{\mathcal{T}} = \frac{1}{|\mathcal{D}|}\sum_{(S,A,S')\in \mathcal{D}} \Vert \mathcal{T}_{\phi}\big(A,\phi(S)\big)-\phi(S') \Vert_2^2,
\end{align*}
where $\mathcal{R}_{\phi}$ and $\mathcal{T}_{\phi}$ are the estimated reward and transition functions applied to the abstract state space parameterized by deep neural networks, and $|\mathcal{D}|$ is the cardinality of the dataset $\mathcal{D}$. 

The inclusion of the third loss function, $\mathcal{L}_Q$, is motivated by the demonstrated benefits of utilizing model-free objectives to guide the training of state abstractions in policy learning \citet{franccois2019combined}.

Given our interest in OPE, we integrate the following FQE loss into the objective function,
\begin{eqnarray*}
    \mathcal{L}_Q = \frac{1}{|\mathcal{D}|}\sum_{(S,A,R,S')\in \mathcal{D}} \Big[R+\gamma \sum_{a\in \mathcal{A}} \pi(a|S') Q^{-} \big(\phi_{for}(S'), a\big) - Q\big(\phi_{for}(S),A\big)\Big]^2,
\end{eqnarray*}
where $Q^-$ and $Q$ represent the estimated $Q^{\pi}_\phi$ function applied to the abstract state space during the previous and current iterations, respectively. 

The above objectives allow us to effectively train type-I MSAs. However, a potential concern is that the resulting abstraction and transition can collapse to some constant $x_0$ such that  $\phi_{for}(S) \rightarrow x_0, ~~\forall S\in \mathcal{S}$. To address this limitation, we include the following penalty function of two randomly drawn states to promote diversity in the abstractions:
\begin{eqnarray*}
\mathcal{L}_{c} = \frac{1}{|\mathcal{D}|(|\mathcal{D}|-1)}\sum_{S,\tilde{S}\in \mathcal{D},S\neq \tilde{S}}\exp (-C_0\Vert \hat\phi(S)-\hat\phi(\tilde{S})\Vert_2)
 %\exp\big[C_0\big(\Vert\phi_{for}(S)- \phi_{for}(S)\Vert_2-d\big)\big],\\
\end{eqnarray*}
for some positive scaling constant $C_0$, and $\hat \phi(s)$ is the estimated abstract state from transition function. $\hat \phi(\tilde{s})$ can be achieved by shuffling  $\hat \phi(s')$ from pairs $(s,s')$ in the batch. Additionally, we add another penalty to penalize consecutive abstract states for being more than some predefined distance $d_0$ away
from each other,
\begin{eqnarray*}
    \mathcal{L}_{s} = \frac{1}{|\mathcal{D}|}\sum_{(S,S')\in \mathcal{D}} C_1[\Vert\phi(S)- \phi(S')\Vert_2-d_0]^2,
\end{eqnarray*}
for some positive constant $C_1$. These components combine into the final penalty function: \[\mathcal{L}_{penalty} = \mathcal{L}_{s}+ \mathcal{L}_{c}.\]

The type-I MSA architecture is as follow:
\begin{verbatim}
    Forward_model(
  (encoder): Encoder_linear(
    (activation): ReLU()
    (encoder_net): Sequential(
      (0): Linear(in_features=300, out_features=64, bias=True)
      (1): ReLU()
      (2): Linear(in_features=64, out_features=64, bias=True)
      (3): ReLU()
      (4): Dropout(p=0.2, inplace=False)
      (5): Linear(in_features=64, out_features=64, bias=True)
      (6): ReLU()
      (7): Dropout(p=0.2, inplace=False)
      (8): Linear(in_features=64, out_features=100, bias=True)
    )
  )
  (transition): Transition(
    (activation): ReLU()
    (T_net): Sequential(
      (0): Linear(in_features=100, out_features=64, bias=True)
      (1): ReLU()
      (2): Linear(in_features=64, out_features=64, bias=True)
      (3): ReLU()
      (4): Dropout(p=0.2, inplace=False)
      (5): Linear(in_features=64, out_features=64, bias=True)
    )
    (lstm): LSTMCell(64, 128)
    (tanh): Tanh()
  )
  (reward): Reward(
    (activation): ReLU()
    (reward_net): Sequential(
      (0): Linear(in_features=100, out_features=64, bias=True)
      (1): ReLU()
      (2): Linear(in_features=64, out_features=64, bias=True)
      (3): ReLU()
      (4): Dropout(p=0.2, inplace=False)
      (5): Linear(in_features=64, out_features=64, bias=True)
      (6): ReLU()
      (7): Dropout(p=0.2, inplace=False)
      (8): Linear(in_features=64, out_features=64, bias=True)
      (9): ReLU()
      (10): Dropout(p=0.2, inplace=False)
      (11): Linear(in_features=64, out_features=64, bias=True)
      (12): ReLU()
      (13): Dropout(p=0.2, inplace=False)
      (14): Linear(in_features=64, out_features=2, bias=True)
    )
  )
  (FQE): FQE(
    (activation): ReLU()
    (action_net): Sequential(
      (0): Linear(in_features=1, out_features=16, bias=True)
      (1): ReLU()
      (2): Linear(in_features=16, out_features=100, bias=True)
    )
    (xa_net): Linear(in_features=200, out_features=100, bias=True)
    (FQE_net): Sequential(
      (0): Linear(in_features=100, out_features=64, bias=True)
      (1): ReLU()
      (2): Linear(in_features=64, out_features=64, bias=True)
      (3): ReLU()
      (4): Dropout(p=0.2, inplace=False)
      (5): Linear(in_features=64, out_features=64, bias=True)
      (6): ReLU()
      (7): Dropout(p=0.2, inplace=False)
      (8): Linear(in_features=64, out_features=2, bias=True)
    )
  )
)
\end{verbatim}

 \begin{table}[t]
  \caption{Hyper-parameters information.  $m$ is the input feature dimension, and $**$ means no value.}\label{tab:hyper_rf}
  \centering
  \begin{tabular}{ccccccc}
    \toprule  
  && Hyper-parameters     & Values &&  Hyper-parameters     & Values \\
    \midrule

 && $\alpha_1$ &1 && $\alpha_2$ & 1    \\
    && $\beta_1$ & 1&&     $\beta_2$  &  1\\
    &&$\gamma_1$ & 1  &&$\gamma_2$&   1\\
     && $\lambda_1$ & $\min(1, \frac{20}{m})$&&   $\lambda_2$ &$\min(1, \frac{20}{m})$\\
       && $C_0$ & 1&&   $C_0$ &$**$\\
         && $C_1$ & 1&&   $C_1$ &1\\
           && $d_0$ & $0.15m$&&   $d_0$ &$0.15m$\\
    \bottomrule
  \end{tabular}
\end{table}

\subsection{Implementation details for type-II MSA}\label{sec:implementbackward} 
%Similar to the aforementioned type-I MSA,
We provide details for implementing the proposed type-II MSA in this subsection. Similar to Section \ref{sec:implementforward}, we use deep neural networks to parameterize the abstraction $\phi_{back}$ and estimate the parameters by solving the following loss function,
\begin{eqnarray*}
\alpha_2 \mathcal{L}_b + \beta_2 \mathcal{L}_{ratio}+\delta_2 \mathcal{L}_{inv}+\lambda_2\mathcal{L}_{s},
\end{eqnarray*}
where $\alpha_2, \beta_2, \delta_2, \lambda_2$ are positive hyper-parameters specified in Table \ref{tab:hyper_rf}.

%We discuss the loss functions one by one. %give the MSE 
Recall that backward-model-irrelevance requires both behavior-irrelevance \eqref {eqn:backirrelevance} and \eqref{eqn:backirrelevance-state}. To enforce behavior-irrelevance, we achieve this by minimizing the following cross-entropy loss for behavior:
\begin{align*}
    \mathcal{L}_b=-\frac{1}{|\mathcal{D}|}\sum_{(S,A)\in \mathcal{D}}\log b\big(A=a|S\big)
\end{align*}
and followed by concatenating with $\{\pi(a|\bullet): a\in \mathcal{A}\}$, which ensures $\pi$-irrelevance. Note that in deeply-abstracted procedure, we replace the behavior loss by
\begin{align*}
    \mathcal{L}_b=-\frac{1}{|\mathcal{D}|}\sum_{(S_t,A_t)\in \mathcal{D}}\log b\big(A_t=a_t|\phi_{1}(S_t),\{A_{t-k},\phi_{1}(S_{t-k})\}_{k=1,2,...}\big)
\end{align*}
which incorporates history information  \eqref{eqn:historydependentbehaviorpolicy} as mentioned in Section~ \ref{sec:iterative}. In practice, we do not use all the history information for behaviour policy, instead we use the history up to past two steps: $b(A_t|\phi_{1}(S_t),A_{t-1},\phi_{1}(S_{t-1}),A_{t-2},\phi_{1}(S_{t-2}))$.

\begin{comment}
\begin{align*}
\mathcal{L}_\rho = \frac{1}{|\mathcal{D}|}\sum_{(S,A)\in \mathcal{D}} \big[ \hat \rho^\pi(A, S) - \rho^\pi_{\phi_{back}}\big(A,\phi_{back}(S)\big)\big]^2,
\end{align*}
where $\hat{\rho}^{\pi}$ denotes some consistent estimator of the IS ratio
    we should replace $\hat{\rho}^{\pi}(A_t,S_t)$ by: 
\[\hat{\rho}^{\pi}_{1}(A_t, \phi_{1}(S_t))=\frac{\pi_{\phi_{1}}(A_t|\phi_{1}(S_t))}{\hat{b}(A_t|\phi_{1}(S_t),A_{t-1},\phi_{1}(S_{t-1}),...)} = \frac{\pi(A|S)}{\hat{b}(A_t|\phi_{1}(S_t),\{A_{t-1},\phi_{1}(S_{t-1})\}},\]
where $\hat b$ is estimated from the abstracted experiences and $\pi(A|S)$ keeps static due to the $\pi$-irrelevance property of type-I MSA. In practice, we did not use all the history information for behaviour policy, we use the history up to past two steps: $\hat{b}(A_t|\phi_{1}(S_t),A_{t-1},\phi_{1}(S_{t-1}),A_{t-2},\phi_{1}(S_{t-2}))$.
\end{comment}

%$\rho^{\pi}$, $\theta_{\rho}$ is the parameter of the according network, and $n$ is the number of samples.  

As commented in Section \ref{sec:iterative}, 
 %to ensure the second condition in Definition \ref{def:backwardmodel} success, we borrow the inverse models and density ratios  in 
\eqref{eqn:backirrelevance-state} holds by satisfying the conditional independence assumption between $(A_t, \phi(S_t))$ and $S_{t+1}$ given $\phi(S_{t+1})$. By Bayesian formula, we can show that it is satisfied by the inverse-model-irrelevance and density-ratio-irrelevance when setting the learning policy $\pi$ to $b$. This motivates us to leverage the %following 
two objectives $\mathcal{L}_{inv}$ and $\mathcal{L}_{ratio}$ used by \citet{allen2021learning} for training MSA. More details regarding these losses can be found in Section 5 of  \citet{allen2021learning}. Note that to obtain non-sequential states $(s, \tilde{s})$ used in $L_{ratio}$, we flip $s'$ in the pairs $(s, s')$ in each batch instead of shuffling.

Finally, $\mathcal{L}_{s}$ corresponds to the smoothness penalty introduced in Section \ref{sec:implementforward}. The type-II MSA architecture is: 
\begin{verbatim}
    Backward_model(
  (encoder): Encoder_linear(
    (activation): ReLU()
    (encoder_net): Sequential(
      (0): Linear(in_features=100, out_features=64, bias=True)
      (1): ReLU()
      (2): Linear(in_features=64, out_features=64, bias=True)
      (3): ReLU()
      (4): Dropout(p=0.2, inplace=False)
      (5): Linear(in_features=64, out_features=64, bias=True)
      (6): ReLU()
      (7): Dropout(p=0.2, inplace=False)
      (8): Linear(in_features=64, out_features=6, bias=True)
    )
  )
  (inverse): Inverse(
    (activation): ReLU()
    (inverse_net): Sequential(
      (0): Linear(in_features=12, out_features=64, bias=True)
      (1): ReLU()
      (2): Linear(in_features=64, out_features=64, bias=True)
      (3): ReLU()
      (4): Dropout(p=0.3, inplace=False)
      (5): Linear(in_features=64, out_features=64, bias=True)
      (6): ReLU()
      (7): Dropout(p=0.3, inplace=False)
      (8): Linear(in_features=64, out_features=64, bias=True)
      (9): ReLU()
      (10): Dropout(p=0.3, inplace=False)
      (11): Linear(in_features=64, out_features=64, bias=True)
      (12): ReLU()
      (13): Dropout(p=0.3, inplace=False)
      (14): Linear(in_features=64, out_features=1, bias=True)
    )
  )
  (density): Density(
    (activation): ReLU()
    (density_net): Sequential(
      (0): Linear(in_features=12, out_features=64, bias=True)
      (1): ReLU()
      (2): Linear(in_features=64, out_features=64, bias=True)
      (3): ReLU()
      (4): Dropout(p=0.3, inplace=False)
      (5): Linear(in_features=64, out_features=64, bias=True)
      (6): ReLU()
      (7): Dropout(p=0.3, inplace=False)
      (8): Linear(in_features=64, out_features=64, bias=True)
      (9): ReLU()
      (10): Dropout(p=0.3, inplace=False)
      (11): Linear(in_features=64, out_features=64, bias=True)
      (12): ReLU()
      (13): Dropout(p=0.3, inplace=False)
      (14): Linear(in_features=64, out_features=1, bias=True)
    )
  )
  (rho): Rho(
    (activation): ReLU()
    (rho_net): Sequential(
      (0): Linear(in_features=6, out_features=64, bias=True)
      (1): ReLU()
      (2): Linear(in_features=64, out_features=64, bias=True)
      (3): ReLU()
      (4): Dropout(p=0.3, inplace=False)
      (5): Linear(in_features=64, out_features=64, bias=True)
      (6): ReLU()
      (7): Dropout(p=0.3, inplace=False)
      (8): Linear(in_features=64, out_features=2, bias=True)
    )
  )
)
\end{verbatim}

\section{Additional Experimental Details}\label{sec:experiments-details}
\subsection{Reproducibility}\label{sec:reproducibility}

%To illustrate the reproducibility, 
We release our code and data on the website at  \\\url{https://github.com/pufffs/state-abstraction}\\
%In addition, we summarize the hyper-parameters information in the applied machine learning methods 
The hyper-parameters to train the proposed type-I MSA and type-II MSA can be found in Table~\ref{tab:hyper_rf}.

\subsection{Experimental settings and additional results} 
For the environment LunarLander, we use Adam \cite{kingma2014adam} optimizer with learning rate $0.003$, and set the learning rate to be $0.001$ for other environments. Model architectures and hyper-parameters are outlined in Section~\ref{sec:implement}. When conducting OPE, the FQE network has $3$ hidden layers with $64$ nodes per hidden layer for abstraction methods, and is equipped with $5$ hidden layers with $128$ nodes per hidden layer for non-abstracted observations (shown as `FQE' in the plot).
%\iffalse
\subsubsection{Atlantis-ram-v5\&SpaceInvaders-ram-v5}
\textbf{Data generating processes}

The observation space is 128-dimensional for both environments, action spaces are 4-dimensional and 6-dimensional for Atlantis and SpaceInvaders, respectively. For both environments, target policies are obtained by training DQNs using the built-in package "stable\_baselines3". The oracle values of target policies of the two environments are $485.37$ and $23.89$ respectively, evaluated by running 5000 episodes using Monte Carlo method. The behavior policies are $\epsilon$-greedy with $\epsilon\in \{0.1,\,0.3,\,0.5\}$. The offline data is collected by the corresponding behavior policies with size $n\in \{1000,2000,4000,10000\}$

\textbf{Model parameters}

For type-I and type-II MSAs, we abstract the original state dimension from $128 \rightarrow 20$. For Atlantis, DSA method reduces dimensions from $128\rightarrow 80 \rightarrow 50 \rightarrow 20 \rightarrow 8$, by [Type-I, Type-II, Type-I, Type-II] order. For SpaceInvaders, DSA method reduces dimensions from $128\rightarrow 80 \rightarrow 50 \rightarrow 10 \rightarrow 8$, by [Type-I, Type-II, Type-I, Type-II] order.

\subsubsection{CartPole-v0}
%We consider 
%both \hao{???? models} for the null variables. 
\textbf{Data generating processes}

We manually insert 296 irrelevant features in the state, each following a first order auto-regressive model (AR(1)) 
\begin{eqnarray*}
\mathbb{P}(S_{t+1,j}|S_t,A_t)=\mathbb{P}(S_{t+1,j}|S_{t,j}), ~~~~~j=5,\dots,300.
\label{AR1-cartpole}
\end{eqnarray*}
We also define a new state-action-dependent reward as \[\mathcal{R}(s_t,a_t)=1-2s_{t,1}^2-5s_{t,3}^2,\] where $s_{t,1}$ and $s_{t,3}$ are the first feature (cart position) and third feature (pole angle) of the state $s_t$, to replace the original constant rewards.
The number of experiences $n$ in the offline dataset is chosen from $\{200, 400, 1000,2000\}$. The target policy is determined by the pole angle: we push the cart to the left if the angle is negative and to the right if it is positive. Namely,
\begin{eqnarray*}
\label{target_behavior}
  \pi(s_t)= \mathbb{1}(s_{t,3}>0).
\end{eqnarray*}
The behavior policy that generates the batch data is set to an $\epsilon$-greedy policy with respect to the target policy, with $\epsilon\in \{0.1,\,0.3,\,0.5\}$.

\textbf{Model parameters}

For type-I and type-II models, we set the abstracted state dimension as $100$. For DSA method, we apply type-II MSA followed by type-I MSA, reducing the dimension from $300 \rightarrow 100 \rightarrow 6$.

\subsubsection{LunarLander-v2}
\textbf{Data generating processes}

We similarly insert 292 irrelevant auto-regressive features in the state:
\begin{eqnarray*}
\mathbb{P}(S_{t+1,j}|S_t,A_t)=\mathbb{P}(S_{t+1,j}|S_{t,j}), ~~~~~j=9,\dots,300.
\label{AR1-lunar}
\end{eqnarray*}
The number of experiences $n$ in the offline dataset is chosen from $\{5000,10000,15000\}$. In this environment, the trajectory length differs significantly, as some lengthy episodes can have length larger than $100000$ while short episodes have fewer than $100$ decision points. We truncate the episode length at 1000 if it exceeds, define it as long episode and those fewer than 1000 as short episodes. For each data size, 2/3 of experiences come from long episodes and 1/3 from short episodes. The target policy is an estimated optimal policy pre-trained by an DQN agent whereas the behavior policy again $\epsilon$-greedy to the target policy with $\epsilon\in \{0.1,\,0.3,\,0.5\}$.
%In addition, 
%for the simpler \hao{???} environment, we report results based on \hao{????} in Figure \hao{??} averaged over \hao{???} runs. 

%The bias and MSE are also provided in Figure \hao{??} and \hao{??} aggregated over \hao{????}
%runs.
\textbf{Model parameters}

For type-I and type-II MSAs, we abstract the original state dimension from $300 \rightarrow 10$, and for DSA method we reduce dimensions from $300\rightarrow 100 \rightarrow 50 \rightarrow 20 \rightarrow 6$, by [Type-II, Type-I, Type-II, Type-I] order.

\textbf{Pre-trained agent}

We pre-train an agent by using DQN as our target policy. The agent is trained until there exists an episode that has accumulative discounted rewards exceeding $200$ with discounted rate $\gamma=0.99$. We evaluated oracle value (61.7) of the  optimized agent by Monte Carlo method with the same discounted rate.
\begin{comment}
  \subsection{Licences for existing assets}\label{sec:licences}
 We consider the environment from OpenAI Gym \citep{brockman2016openai} ``LunarLander-v2'' with the MIT License and Copyright (c) 2016 OpenAI (\url{https://openai.com}).  
\end{comment}

\subsection{Computing resources}\label{sec:computation-details} 
To build Figure \ref{all_figs}, we trained 6 abstraction methods and one non-abstraction method for 4 environments, each on 4 different sizes of data, each with 30 runs, under 3 $\epsilon$ values. In average, each run takes approximately 12 minutes for all models on an 16-core E2-series CPU with 64GB memory on GCP. It takes about 240 compute hours to complete all the experiments in the figure.

\begin{comment}
    \section{Limitations}\label{sec:limitation}
Our proposal presents several limitations. Firstly, although empirical results validate the effectiveness of the proposed state abstraction for OPE, we have not conducted a theoretical analysis to determine if state abstraction leads to a more efficient OPE estimator with reduced MSE compared to estimators without abstraction. Additionally, we have not theoretically examined if the iterative procedure's estimator achieves a smaller MSE than estimators derived from single-iteration forward or type-II MSA. 
Moreover, the reliance on abstracted state spaces could potentially overlook nuanced interactions in the environment. This is particularly concerning for applications in sensitive areas such as healthcare and autonomous driving, where overlooked interactions could impact the safety and efficacy of policy deployment. Addressing these concerns will require further investigation.
\end{comment}
\section{Technical proofs}\label{sec:proofs}
We provide the detailed proofs of our theorems (Theorems \ref{postmse} \& \ref{Tpostmse} and Lemma~\ref{backMSA}) in this section.  

\textbf{Notations}.
\iffalse
Following \cite{li2006towards}, we can define \begin{align*}
&Q_{\phi}^{\pi}(a,x)=\sum_{s\in \phi^{-1}(x)}w_x(s)Q^{\pi}(a,s),\\
 &\rho_{\phi}^{\pi}(a,x)=\sum_{s\in \phi^{-1}(x)}w_x(s)\rho^{\pi}(a,s),\\
 &w_{\phi}^{\pi}(a,x)=\sum_{s\in \phi^{-1}(x)}w_x(s)w^{\pi}(a,s).
\end{align*}
Here, the weight function $w_x(s)$ satisfies $\sum_{s\in \phi^{-1}(x)}w_x(s)=1$. \z{Is the weight function $w$ $x$-dependent? Otherwise, how can guarantee this 'sum to one' contraint for different $x$-s (if we are summing over potentially different sets)?}{\color{red}Yeah, you are right, it also depends on $x$. I have changed it as $w_x(s)$.}
\fi
For events or random variables $A, B, C$, $A \indep B$ means the independence between $A$ and $B$ whereas $A \indep B|C$ means the conditional independence between $A$ and $B$ given $C$. Let $c$ be a positive constant, which may take different values in different contexts.
For two positive sequences $\{a_{n}\}_{n \geq 1}$, $\{b_{n}\}_{n \geq 1}$, we write $a_{n}=O(b_{n})$ if there exists a positive constant $c$ such that $a_{n}\leq c\cdot b_{n}$, and we write $a_{n}=o(b_{n})$ if $a_{n}/b_{n}\rightarrow 0$. 
	Furthermore, if $a_{n}=O(b_{n})$ is satisfied, we write $a_n\lesssim b_n$. If
	$a_n\lesssim b_n$ and $b_n\lesssim a_n$, we write it as $a_n\asymp b_n$ for short. For any function $f$, denote $\Vert f\Vert=\sqrt{\mathbb{E}[f(X)^2]}$ with $X$ being random vector, $\mathbb{P}_n f=(NT)^{-1}\sum_{i=1}^N\sum_{j=1}^T f(A_{i,j},S_{i,j})$, 
$$\hat{\Sigma}=\mathbb{P}_n\left[-\gamma \sum_{a'}\pi(a'|S')\varphi\left(a',S'\right) \varphi(A,S)^{\top}+\varphi(A,S) \varphi(A,S)^{\top}\right],$$ with $\Sigma=\mathbb{E}\left[-\gamma \sum_{a}\pi(a|S')\varphi\left(a,S'\right) \varphi(A,S)^{\top}+\varphi(A,S) \varphi(A,S)^{\top}\right]$ being the population counterpart. 

\textbf{Auxiliary lemmas}. To begin with, we introduce the subsequent lemmas. One lemma exhibits the bias-variance decomposition of diverse OPE estimators for MDP, whose  proof is provided in Section \ref{proofofmse}. Additionally, we present the properties regarding the linear estimation of \(Q\) and \(w\), with the corresponding proof furnished in Section \ref{proofofkappa}.

\begin{lemma}
[Bias-variance decomposition for the MSE]\label{mse}
    Assume Assumptions \ref{asmp:bounded}--\ref{VC} hold. Suppose that $\Vert\hat{Q}^{\pi}-{Q}^{\pi}\Vert=O(R_{\mathrm{max}}(1-\gamma)^{-1}\kappa_q)$, and $\Vert\hat{w}^{\pi}-{w}^{\pi}\Vert=O(\kappa_w)$ are satisfied.
    \begin{itemize}[leftmargin=*]
        \item   The MSE of \textbf{DRL} (Method 4) applied to the MDP is given by:
        \begin{align}\label{MSEDRL}
&\frac{1}{N}\mathbb{V}\left[V^{\pi}\left(S\right)\right]+\frac{1}{NT(1-\gamma)^2}\mathbb{V}\left[w^{\pi}(A,S) \left(R+\gamma V^{\pi}\left(S'\right)-Q^{\pi}(A,S)\right)\right]\nonumber\\
&
+O\left(\frac{R_{\max}^2\max(\kappa_{q},\kappa_w,N^{-1/2})\log (N)}{(1-\gamma)^2N}+\frac{R_{\max}^2}{\sqrt{N}(1-\gamma)^{2}}\max\{\kappa_{q}\kappa_{w},\sqrt{N}\kappa_{q}^2\kappa_{w}^2\}\right).
 \end{align} 
 \item When both the Q-function and MIS ratio are estimated via linear function approximation, the MSEs of \textbf{Q-function-based estimator} (Method 1) and \textbf{MIS} (Method 3) applied to the  MDP  identical to that of DRL.
 \item When $\rho^{\pi}$ is known, the MSE of \textbf{SIS} (Method 2) applied to the  MDP is upper bounded by   
        \begin{align*}
            \frac{4R_{\max}^2}{NT^2(1-\gamma)^2}\sum_{t=1}^T \gamma^{2t-2}\mathbb{E}(\rho_{1:t}^{\pi})^2
+\frac{\gamma^{2T}R_{\max}^2}{(1-\gamma)^2}.
\end{align*}
    \end{itemize}
\end{lemma}

Before we figure out the value of $\kappa_q$ and $\kappa_w$, we need the following assumption.
\begin{asmp}[Positive definite]\label{asmp:positive}
(i) $\lambda_{\min }(\Sigma) \geq {c}$ for some constant ${c}>0$ where $\lambda_{\min }(K)$ denotes the minimum eigen value of a matrix $K$.
\end{asmp}   

\begin{lemma}
[The order of $\hat{Q}^{\pi}$ and $\hat{w}^{\pi}$]\label{kappa}
    Suppose Assumptions \ref{asmp:bounded}--\ref{asmp:positive} hold. When we utilize a linear model to estimate \(Q\) and \(w\), within the tabular setting, it can be demonstrated that both \(\kappa_q\) and \(\kappa_w\) are of the order  $\kappa_q$ and $\kappa_w$ are  $(NT)^{-1/2}|\mathcal{S}||\mathcal{A}|$ and $N^{-1/2}|\mathcal{S}||\mathcal{A}|$, respectively.
\end{lemma}
Based on Lemmas \ref{mse} and \ref{kappa}, we can get the bias-variance decomposition for the MSE under tabular setting.
\begin{Coro}
[Bias-variance decomposition for MSE in tabular setting]\label{Tmse}
    Assume Assumptions \ref{asmp:bounded}--\ref{asmp:positive} hold.  
  The bias-variance decomposition MSE of DRL is given by:
        \begin{align*}
&\frac{1}{N}\mathbb{V}\left[V^{\pi}\left(S\right)\right]+\frac{1}{NT(1-\gamma)^2}\mathbb{V}\left[w^{\pi}(A,S) \left(R+\gamma V^{\pi}\left(S'\right)-Q^{\pi}(A,S)\right)\right]\nonumber\\
&
+O\left(\frac{R_{\max}^2|\mathcal{S}||\mathcal{A}| \log (N)}{(1-\gamma)^2N^{3/2}}\right)+O(\frac{R_{\max}^2}{\sqrt{N}(1-\gamma)^{2}}\frac{|\mathcal{S}|^2|\mathcal{A}|^2}{NT^{1/2}}).
 \end{align*} 
\end{Coro}

\subsection{Proof of Theorem \ref{postmse}}\label{proofofpostMSE}
Since $\phi$ is MSA, then Theorem \ref{postmse} directly follows from Lemma \ref{mse}.

\subsection{Poof of Lemma \ref{mse}}\label{proofofmse}
In the following, we calculate the MSE of DRL, Q-function-based method, MIS and SIS one by one. 
\begin{itemize}[leftmargin=*]
    \item \textbf{MSE of DRL}. 
 Our DRL estimator based on the dataset $\mathcal{D}$ is
\begin{align*}
\hat{J}_{\mathrm{DRL}}\left(\hat{Q}^{\pi},\hat{w}^{\pi}\right)= & \mathbb{P}_n\left[\psi\left(\hat{Q}^{\pi},\hat{w}^{\pi}\right)\right] \\
= & \frac{1}{N} \sum_{i=1}^N \pi(A_{i,1}|S_{i,1})\hat{Q}^{\pi}\left(A_{i,1},S_{i,1}\right) 
 +(1-\gamma)^{-1}\frac{1}{NT} \sum_{i=1}^N\sum_{j=1}^T
\hat{w}^\pi\left(A_{i,j},S_{i,j}\right)\\
&\times\left(R_{i,j}+\gamma\sum_{a}\pi(a|S_{i,j+1})\hat{Q}^\pi\left(a,S_{i,j+1}\right)- \hat{Q}^\pi\left(A_{i,j},S_{i,j}\right) \right),
\end{align*}
where $\hat{Q}^{\pi},\hat{w}^{\pi}$ are some plug-in estimates of $Q^{\pi}$ and $w^\pi$, respectively. 
Now, we will calculate the MSE of $\hat{J}_{\mathrm{DRL}}$. A direct calculation yields
\begin{align*}
&\mathbb{P}_n \psi\left[\left(\hat{Q}^{\pi},\hat{w}^{\pi}\right)\right]-J(\pi) \\
=&\left\{\mathbb{P}_n\left[\psi\left(\hat{Q}^{\pi},\hat{w}^{\pi}\right)-\psi\left(w^{\pi}, Q^{\pi}\right)\right]+ \left\{\mathbb{P}_n\left[\psi\left(w^{\pi}, Q^{\pi}\right)\right]- J(\pi)\right\}\right\}\\
=&\mathbb{P}_n\left[\psi\left(\hat{Q}^{\pi},\hat{w}^{\pi}\right)-\psi\left(w^{\pi}, Q^{\pi}\right)\right]+ \left\{\mathbb{P}_n\left[\psi\left(w^{\pi}, Q^{\pi}\right)\right]- J(\pi)\right\}\\
=&I_1+I_2.
\end{align*}
For $I_1$, we have
\begin{align}
I_1=&\mathbb{P}_n\left[\psi\left(\hat{Q}^{\pi},\hat{w}^{\pi}\right)-\psi\left(w^{\pi}, Q^{\pi}\right)\right]\nonumber\\
=&\mathbb{P}_n\left[\psi\left(\hat{Q}^{\pi},\hat{w}^{\pi}\right)-
\psi\left({w}^{\pi}, \hat{Q}^{\pi}\right)\right]\nonumber\\
&+\mathbb{P}_n\left[\psi\left({w}^{\pi},\hat{Q}^{\pi}\right)-\psi\left(w^{\pi}, Q^{\pi}\right)\right]\nonumber\\
%=&(1-\gamma)^{-1}\frac{1}{NT} \sum_{i=1}^N\sum_{j=1}^T
%\left\{\hat{w}^\pi\left(A_{i,j},S_{i,j}\right)- {w}^\pi\left(A_{i,j},S_{i,j}\right)\right\}\nonumber\\
%&\times\left\{R_{i,j}+\gamma \sum_{a}\pi(a|S_{i,j+1})\hat{Q}^\pi\left(a,S_{i,j+1}\right)-\hat{Q}^\pi\left(A_{i,j},S_{i,j}\right)\right\}\nonumber\\
%&+\frac{1}{N} \sum_{i=1}^N \pi(A_{i,1}|S_{i,1})\left\{\hat{Q}^{\pi}\left(A_{i,1},S_{i,1}\right)-{Q}^{\pi}\left(A_{i,1},S_{i,1}\right)\right\} \nonumber\\
%& +(1-\gamma)^{-1}\frac{1}{NT} \sum_{i=1}^N\sum_{j=1}^T
%{w}^\pi\left(A_{i,j},S_{i,j}\right) \left(\gamma \sum_{a}\pi(a|S_{i,j+1})\left\{\hat{Q}^\pi\left(a,S_{i,j+1}\right)-{Q}^\pi\left(a,S_{i,j+1}\right)\right\}\right.\nonumber\\
%&\left.-\left\{\hat{Q}^\pi\left(A_{i,j},S_{i,j}\right)
%-{Q}^\pi\left(A_{i,j},S_{i,j}\right)\right\} \right)\nonumber\\
=&\frac{1}{N}\sum_{i=1}^N \pi(A_{i,1}|S_{i,1})\left\{\hat{Q}^{\pi}\left(A_{i,1},S_{i,1}\right)-{Q}^{\pi}\left(A_{i,1},S_{i,1}\right)\right\} \nonumber\\
& +(1-\gamma)^{-1}\frac{1}{NT} \sum_{i=1}^N\sum_{j=1}^T
{w}^\pi\left(A_{i,j},S_{i,j}\right)\left(\gamma \sum_{a}\pi(a|S_{i,j+1})\left\{\hat{Q}^\pi\left(a,S_{i,j+1}\right)-{Q}^\pi\left(A_{i,j+1},S_{i,j+1}\right)\right\}\right.\nonumber\\
&\left.-\left\{\hat{Q}^\pi\left(A_{i,j},S_{i,j}\right)
-{Q}^\pi\left(A_{i,j},S_{i,j}\right)\right\} \right) +(1-\gamma)^{-1}\frac{1}{NT} \sum_{i=1}^N\sum_{j=1}^T
\left\{\hat{w}^\pi\left(A_{i,j},S_{i,j}\right)-{w}^\pi\left(A_{i,j},S_{i,j}\right)\right\}\nonumber\\
&\times\left(\gamma\sum_{a} \pi(a|S_{i,j+1})\left\{\hat{Q}^\pi\left(a,S_{i,j+1}\right)-{Q}^\pi\left(a,S_{i,j+1}\right)\right\}-\left\{\hat{Q}^\pi\left(A_{i,j},S_{i,j}\right)
-{Q}^\pi\left(A_{i,j},S_{i,j}\right)\right\} \right)\nonumber\\
&+(1-\gamma)^{-1}\frac{1}{NT} \sum_{i=1}^N\sum_{j=1}^T
\left\{\hat{w}^\pi\left(A_{i,j},S_{i,j}\right)- {w}^\pi\left(A_{i,j},S_{i,j}\right)\right\}\nonumber\\
&\times\left\{R_{i,j}+\gamma \sum_{a}\pi(a|S_{i,j+1}){Q}^\pi\left(a,S_{i,j+1}\right)-{Q}^\pi\left(A_{i,j},S_{i,j}\right)\right\}.
\end{align}
 Denote 
 \begin{align}\label{DRLI11}
 I_{11}=&\frac{1}{N}\sum_{i=1}^N \pi(A_{i,1}|S_{i,1})\left\{\hat{Q}^{\pi}\left(A_{i,1},S_{i,1}\right)-{Q}^{\pi}\left(A_{i,1},S_{i,1}\right)\right\}+(1-\gamma)^{-1}\frac{1}{NT} \sum_{i=1}^N\sum_{j=1}^T
{w}^\pi\left(A_{i,j},S_{i,j}\right) \nonumber\\
&\times \left(\gamma \sum_{a}\pi(a|S_{i,j+1})\left\{\hat{Q}^\pi\left(a,S_{i,j+1}\right)-{Q}^\pi\left(a,S_{i,j+1}\right)\right\}-\left\{\hat{Q}^\pi\left(A_{i,j},S_{i,j}\right)
-{Q}^\pi\left(A_{i,j},S_{i,j}\right)\right\} \right),
 \end{align}
\begin{align}
&I_{12}=(1-\gamma)^{-1}\frac{1}{NT} \sum_{i=1}^N\sum_{j=1}^T
\left\{\hat{w}^\pi\left(A_{i,j},S_{i,j}\right)-{w}^\pi\left(A_{i,j},S_{i,j}\right)\right\} \nonumber\\
&\times\left(\gamma \sum_{a}\pi(a|S_{i,j+1})\left\{\hat{Q}^\pi\left(a,S_{i,j+1}\right)-{Q}^\pi\left(a,S_{i,j+1}\right)\right\}
-\left\{\hat{Q}^\pi\left(A_{i,j},s_{i,j}\right)
-{Q}^\pi\left(A_{i,j},S_{i,j}\right)\right\} \right),
\end{align}
and
\begin{align}\label{DRLI13}
I_{13}=&(1-\gamma)^{-1}\frac{1}{NT} \sum_{i=1}^N\sum_{j=1}^T
\left\{\hat{w}^\pi\left(A_{i,j},S_{i,j}\right)- {w}^\pi\left(A_{i,j},S_{i,j}\right)\right\}\nonumber\\
&\times\left\{R_{i,j}+\gamma\sum_{a} \pi(a|S_{i,j+1}){Q}^\pi\left(a,S_{i,j+1}\right)-{Q}^\pi\left(A_{i,j},S_{i,j}\right)\right\}.
\end{align}

It follows from Assumption \ref{VC}  that we allow the function classes to be infinite hypothesis classes.
To handle this, we can find $\epsilon$-nets of $\mathcal{Q}$ and $\mathcal{W}$, denoted by $\mathcal{Q}_0$ and $\mathcal{W}_0$ with $\epsilon$ proportional to $N^{-1} R_{\max } /(1-\gamma)$ and $N^{-1}$, respectively. This kind of
 restricting to the finite hypothesis class $\mathcal{Q}_0$ and $\mathcal{W}_0$ provides  reasonable approximations for $\mathcal{Q}$ and $\mathcal{W}$ with the approximation error upper bounded by
$$
O\left(\frac{R_{\max }}{(1-\gamma) N }\right)\,\, \mathrm{and} \,\,O\left(\frac{1}{N}\right).
$$
Meanwhile, the number of elements in $\mathcal{Q}_0$  and $\mathcal{W}_0$ is of the order $O\left(N^{V_q} \right)$ and $O\left(N^{V_w} \right)$ for some positive constants $V_q$ and $V_w$.

Now, we analyze the error bound of $I_{11}$. 
Fix $f\in \mathcal{Q}_0$ and  define
\begin{align}
X_i\left(f\right)=&\pi(A_{i,1}|S_{i,1})\left\{f\left(A_{i,1},S_{i,1}\right)-Q^{\pi}\left(A_{i,1},S_{i,1}\right)\right\}+(1-\gamma)^{-1}\frac{1}{T} \sum_{j=1}^T
{w}^\pi\left(A_{i,j},S_{i,j}\right)\nonumber\\
&\times\left(\gamma \sum_{a}\pi(a|S_{i,j+1})\left\{f\left(a,S_{i,j+1}\right)-Q^{\pi}\left(a,S_{i,j+1}\right)\right\}-\left\{f\left(A_{i,j},S_{i,j}\right)
-Q^{\pi}\left(A_{i,j},S_{i,j}\right)\right\} \right),
\end{align}
It is easy to see 
$$
I_{11}=\frac{1}{N} \sum_{i=1}^N X_i\left(\hat{Q}^{\pi}\right)\,\, \mathrm{and}\,\, \mathbb{E}[X(\hat{Q}^{\pi})]=0.
$$

Since $X_i\left(f\right)$ are i.i.d., we apply Bernstein's inequality and the union bound over all $f\in \mathcal{Q}_0$. With probability at least 
$1-\delta$, we have
\begin{align}\label{Bern}
& |\frac{1}{n} \sum_{i=1}^n X_i\left(f\right)-\mathbb{E}\left[X\left(f\right)\right]|\nonumber\\
\leq & \sqrt{\frac{2 \mathbb{V}\left[X\left(f\right)\right] \ln \frac{2|\mathcal{Q}_0|}{\delta}}{N}}+\frac{2c R_{\max } \ln \frac{2|\mathcal{Q}_0|}{\delta}}{3 N(1-\gamma)}.
\end{align}
It follows from Assumption \ref{VC},  and $\Vert\hat{Q}^{\pi}(A,S)-{Q}^{\pi}(A,S)\Vert=O(R_{\mathrm{max}}(1-\gamma)^{-1}\kappa_q)$ and Jensen's inequality that
\begin{align}\label{Vnorm}
\Vert\sum_{a}\pi(a|S)\big[\hat{Q}^{\pi}(a,S)-{Q}^{\pi}(a,S)\big]\Vert=O(R_{\mathrm{max}}(1-\gamma)^{-1}\kappa_q).
\end{align}
Combing Assumption \ref{VC} and equation \eqref{Vnorm}, we can obtain
\begin{align}\label{Qvar}
\mathbb{V}\left[X\left(\hat{Q}^{\pi}\right)\right]\lesssim (1-\gamma)^{-2}R_{\max}^2\kappa_q^2.
\end{align}
More specifically, by substituting \eqref{Qvar} into \eqref{Bern}, it can be shown that with probability at least $1-O\left(N^{-2} \right)$, $I_{11}$ can be upper bounded by
\begin{align}\label{BI11} O\left(\frac{R_{\max } \log (N)}{(1-\gamma) N }\right)+O\left(\frac{R_{\max}\kappa_{q} \sqrt{\log (N )}}{(1-\gamma)\sqrt{N}}\right).
\end{align}
By similar arguments as above, it can be shown that with probability at least $1-O\left(N^{-2} \right)$, $I_{13}$ can be upper bounded by
\begin{align}\label{BI13} O\left(\frac{R_{\max }\log (N )}{(1-\gamma) N }\right)+O\left(\frac{R_{\max }\kappa_{w} \sqrt{\log (N )}}{ (1-\gamma)\sqrt{N}}\right).
\end{align}
For $I_{12}$, direct calculations yield 
\begin{align}\label{BI14}
\mathbb{E}|I_{12}|^2 \lesssim (1-\gamma)^{-2}R_{\max}^2\kappa_{q}^2\kappa_{w}^2.
\end{align}

Thus, we can get
\begin{align}\label{BI1}
\mathbb{E}[I_1^2]\leq& O\left(\frac{R_{\max}^2 \log^2 (N)}{(1-\gamma)^2 N^2 }\right)+O\left(\frac{R_{\max}^2\kappa_{w}^2 \log (N)}{ (1-\gamma)^2N}\right)
+O(\frac{R_{\max}^2}{(1-\gamma)^{2}}\kappa_{q}^2\kappa_{w}^2)\nonumber\\
&+O\left(\frac{R_{\max}^2\kappa_{q}^2 \log (N )}{(1-\gamma)^2N}\right)+O(\frac{R_{\max}^2}{N^2(1-\gamma)^2}).
\end{align}
For $I_2$, direct calculations yield
\begin{align}\label{BI2}
\mathbb{E}[I_2^2]=\frac{1}{N}\mathbb{V}\left[V^{\pi}\left(S\right)\right]+\frac{1}{NT(1-\gamma)^2}\mathbb{V}\left[w^{\pi}(A,S) \left(R+\gamma V^{\pi}\left(S'\right)-Q^{\pi}(A,S)\right)\right].
\end{align}
 By Cauchy-Schwartz inequality, \eqref{BI1} and \eqref{BI2}, we can get
\begin{align}\label{BI1I2}
\mathbb{E}[I_{1}I_{2}]\leq &\sqrt{\mathbb{E}[I_{1}^2]}\sqrt{\mathbb{E}[I_{2}^2]}\nonumber\\
&\leq \frac{R_{\max}}{(1-\gamma)\sqrt{N}}\left\{O\left(\frac{R_{\max}\max(\kappa_{q},\kappa_w,\sqrt{N^{-1}\log(N)}) \sqrt{\log (N )}}{(1-\gamma)\sqrt{N}}\right)+O(\frac{R_{\max}}{(1-\gamma)}\kappa_{q}\kappa_{w})\right\}.
\end{align}
Combing \eqref{BI1}, \eqref{BI2} and \eqref{BI1I2}, we can get
 \begin{align*}
\mathrm{MSE}_{\mathrm{DRL}}=& \mathbb{E}[\hat{J}_{\mathrm{DRL}}\left(\hat{Q}^\pi,\hat{w}\right)-J(\pi)]^2\nonumber\\
=&\frac{1}{N}\mathbb{V}\left[V^{\pi}\left(S\right)\right]+\frac{1}{NT(1-\gamma)^2}\mathbb{V}\left[w^{\pi}(A,S) \left(R+\gamma V^{\pi}\left(S'\right)-Q^{\pi}(A,S)\right)\right]\nonumber\\
&
+O\left(\frac{R_{\max}^2\max(\kappa_{q},\kappa_w,N^{-1/2})\log (N)}{(1-\gamma)^2N}+\frac{R_{\max}^2}{\sqrt{N}(1-\gamma)^{2}}\max\{\kappa_{q}\kappa_{w},\sqrt{N}\kappa_{q}^2\kappa_{w}^2\}\right).
 \end{align*}

 \item \textbf{MSE of Q-function based method}. 
 It follows from Example 7 in  \cite{uehara2020minimax} and $\hat{\Sigma}$ is non-sigular  that, when we use the linear method to estimate $Q^{\pi}$ by MQL, 
 $\hat{Q}^{\pi}(a, s)=\varphi( a,s)^{\top} \hat{\beta}$ where $\phi(a,s) \in$ $\mathbb{R}^d$ is some basis function and $\hat{\beta}$  takes the form
$$
\hat{\Sigma}^{-1} \mathbb{P}_n[R\varphi(A,S)].
$$

A direct calculation yields
\begin{align*}
\hat{\beta}^\top\mathbb{P}_n\left[-\gamma \sum_{a'}\pi(a'|S')\varphi\left(a',S'\right) \varphi(A,S)^{\top}+\varphi(A,S) \varphi(A,S)^{\top}\right]-\mathbb{P}_n[R\varphi( A,S)^\top]=0.
\end{align*}
Consider $w^{\pi}\in \mathcal{F}$ whose parameter is $\alpha$, and denote $\hat{w}^{\pi}(a,s)=\varphi(a,s)^\top\hat{\alpha}$.
It is easy to verify 
\begin{align*}
    0=&\left\{\hat{\beta}^\top\mathbb{P}_n\left[-\gamma \sum_{a'}\pi(a'|S')\varphi\left(a',S'\right) \varphi(A,S)^{\top}+\varphi(A,S) \varphi(A,S)^{\top}\right]-\mathbb{P}_n[R\varphi( A,S)^\top]\right\}\hat{\alpha}\\
    =&\mathbb{P}_n\left[-\gamma \sum_{a'}\pi(a'|S')\hat{\beta}^\top\varphi\left(a',S'\right) \varphi(A,S)^{\top}\hat{\alpha}+\hat{\beta}^\top\varphi(A,S) \varphi(A,S)^{\top}\hat{\alpha}\right]-\mathbb{P}_n[R\varphi( A,S)^\top\hat{\alpha}]\\
    =&-\frac{1}{NT} \sum_{i=1}^N\sum_{j=1}^T
\hat{w}^\pi\left(A_{i,j},S_{i,j}\right)\left(R_{i,j}+\gamma\sum_{a}\pi(a|S_{i,j+1})\hat{Q}^\pi\left(a,S_{i,j+1}\right)- \hat{Q}^\pi\left(A_{i,j},S_{i,j}\right) \right).
\end{align*}
This means
$\hat{J}_{\mathrm{Q}}(\hat{Q}^\pi)=\hat{J}_{\mathrm{DLR}}(\hat{Q}^\pi,\hat{w}^\pi)$.
Thus, we have
\begin{align*}
\mathrm{MSE}_{\mathrm{Q}}=\mathrm{MSE}_{\mathrm{DRL}}.
 \end{align*}

\item \textbf{MSE of MIS}. We assume that both $\mathcal{W}$ and $\mathcal{Q}$ are set to the same linear function space $\mathcal{F}=\{(a,s)$ $\left.\varphi(a,s)^{\top} \beta: \beta \in \mathbb{R}^d\right\}$, and
$\hat{\Sigma}$ is non-singular.
It follows from Example 2 in  \cite{uehara2020minimax} that, the estimator of $w^{\pi}(a,s)$ by MWL takes the following closed form:
$$\hat{w}^\pi(a,s)=\varphi(a,s)^{\top}\hat{\alpha},$$
with
$$
\begin{aligned}
\hat{\alpha}= & \hat{\Sigma}^{-1}  (1-\gamma) \frac{1}{N}\sum_{i=1}^{N}\left[\sum_{a}\pi(a|S_{i,1})\varphi\left(a, S_{i,1}\right)\right] .
\end{aligned}
$$
Direct calculations yield
\begin{align}\label{wequality}
-\hat{\alpha}^{\top} \hat{\Sigma}
+(1-\gamma) \frac{1}{N}\sum_{i=1}^{N}\left[\sum_{a}\pi(a|S_{i,1})\varphi\left(a, S_{i,1}\right)^\top\right] =0.
\end{align}
Consider $Q^{\pi}\in \mathcal{F}$ whose parameter is $\beta$, and denote $\hat{Q}^{\pi}(a,s)=\varphi(a,s)^\top\hat{\beta}$.
It is easy to verify 
$$
\begin{aligned}
0& =\left(-\hat{\alpha}^{\top}\hat{\Sigma} +(1-\gamma) \frac{1}{N}\sum_{i=1}^{N}\left[\sum_{a}\pi(a|S_{i,1})\varphi\left(a, S_{i,1}\right)\right]\right) \hat{\beta}\\
=&\mathbb{P}_n\left[\gamma \hat{\alpha}^{\top} \varphi(A,S) \sum_{a'}\pi(a'|S')\varphi\left(a',S'\right)^{\top} \beta-\hat{\alpha}^{\top} \varphi(A,S) \varphi(A,S)^{\top} \beta\right]\\
&+(1-\gamma) \frac{1}{N}\sum_{i=1}^{N}\left[\sum_{a}\pi(a|S_{i,1})\varphi\left(a, S_{i,1}\right)\right]\hat{\beta} \\
=&(1-\gamma)\frac{1}{N} \sum_{i=1}^N \sum_{a}\pi(a|S_{i,1})\hat{Q}^{\pi}\left(a,S_{i,1}\right) 
 +\frac{1}{NT} \sum_{i=1}^N\sum_{j=1}^T
\hat{w}^\pi\left(A_{i,j},S_{i,j}\right)\\
&\times\left(\gamma\sum_{a}\pi(a|S_{i,j+1})\hat{Q}^\pi\left(a,S_{i,j+1}\right)- \hat{Q}^\pi\left(A_{i,j},S_{i,j}\right) \right).
\end{aligned}
$$
This means
$\hat{J}_{\mathrm{MIS}}(\hat{w}^\pi)=\hat{J}_{\mathrm{DLR}}(\hat{Q}^\pi,\hat{w}^\pi)$.
Thus, we have
\begin{align*}
\mathrm{MSE}_{\mathrm{MIS}}
=\mathrm{MSE}_{\mathrm{DRL}}.
 \end{align*}

 \item \textbf{MSE of SIS}.
Our SIS estimator based on the dataset $\mathcal{D}$ is
\begin{align*}
\hat{J}_{\mathrm{SIS}}\left({\rho^{\pi}}\right)
= & \frac{1}{NT} \sum_{i=1}^N\sum_{j=1} ^{T}\gamma^{j-1}\rho_{1:j}^{\pi}(A_{i,j},S_{i,j})R_{i,j},
\end{align*}
where $T$ is the termination time.
Direct calculations yield 
\begin{align*}
\mathrm{MSE}_{\mathrm{SIS}}=&\mathbb{E}[\hat{J}_{\mathrm{SIS}}\left({\rho^{\pi}}\right)- J(\pi)]^2\\
=&\mathbb{E}\left[\hat{J}_{\mathrm{SIS}}\left({\rho^{\pi}}\right)-\mathbb{E}[{\hat{J}_{\mathrm{SIS}}\left({\rho^{\pi}}\right)}]-\mathbb{E}[{\frac{1}{N} \sum_{i=1}^N\sum_{j=T+1} ^{+\infty}\gamma^{j-1}\rho_{1:j}^{\pi}(A_{i,j},S_{i,j})R_{i,j}}]\right]^2\\
=&\frac{1}{NT^2}\mathbb{V}\left[\sum_{j=1} ^{T}\gamma^{j-1}\rho_{1:j}^{\pi}(A_{i,j},S_{i,j})R_{i,j}\right]+\left(\mathbb{E}[{\frac{1}{N} \sum_{i=1}^N\sum_{j=T+1} ^{+\infty}\gamma^{j-1}\rho_{1:j}^{\pi}(A_{i,j},S_{i,j})R_{i,j}}]\right)^2.
\end{align*}
Denote $\mathcal{H}_{i,t}$ as the $\sigma$-field generated by $\{A_{i,j},S_{i,j},R_{i,j}\}_{j<t}\cup \{A_{i,t},S_{i,t}\}$, and $\mathcal{H}_{i,0}$ as the empty set.
A direct calculation yields
\begin{align}\label{conE}
  \mathbb{E}\left(\sum_{j=t} ^{+\infty}\gamma^{j-1}\rho_{1:j}^{\pi}(A_{i,j},S_{i,j})R_{i,j}|\mathcal{H}_{i,t}\right)=\gamma^{t-1}\rho_{1:t}^{\pi}(A_{i,t},S_{i,t})Q^{\pi}(A_{i,t},S_{i,t}).
\end{align}
Thus, we have 
\begin{align*}
&\mathbb{E}\left[{\frac{1}{N} \sum_{i=1}^N\sum_{j=T+1} ^{+\infty}\gamma^{j-1}\rho_{1:j}^{\pi}(A_{i,j},S_{i,j})R_{i,j}}\right]\\
=&\mathbb{E}\left(\mathbb{E}\left[{\sum_{j=T+1} ^{+\infty}\gamma^{j-1}\rho_{1:j}^{\pi}(A_{i,j},S_{i,j})R_{i,j}}|\mathcal{H}_{i,T+1}\right]\right)\\
=&\gamma^{T}\mathbb{E}\left(\rho_{1:T+1}^{\pi}(A_{i,T+1},S_{i,T+1})Q^{\pi}(A_{i,T+1},S_{i,T+1})\right)\\
=&\gamma^{T}\mathbb{E}\left[\mathbb{E}\left(\rho_{1:T+1}^{\pi}(A_{i,T+1},S_{i,T+1})Q^{\pi}(A_{i,T+1},S_{i,T+1})\right)|S_{i,T+1}\right]\\
\leq &\gamma^{T}\frac{R_{\max}}{1-\gamma}.
\end{align*}
Denote $Q_t^{\pi,T}(a,s)=\mathbb{E}^{\pi}(\sum_{j=t}^{T}\gamma^{t-1}R_{t}|A_{t}=a,S_{t}=s)$.
By the law of total variance in \cite{bowsher2012identifying}, we have 
\begin{align}\label{eqSIS}
&\mathbb{V}\left[\sum_{j=1} ^{T}\gamma^{j-1}\rho_{1:j}^{\pi}(A_{i,j},S_{i,j})R_{i,j}\right]\nonumber\\
=&\sum_{t=1}^{T+1}\mathbb{E}\left[\mathbb{V}\left\{\mathbb{E}\left(\sum_{j=1} ^{T}\gamma^{j-1}\rho_{1:j}^{\pi}(A_{i,j},S_{i,j})R_{i,j}|\mathcal{H}_{i,t}\right)|\mathcal{H}_{i,t-1}\right\}\right]\nonumber\\
=&\sum_{t=1}^{T+1}\mathbb{E}\left[\mathbb{V}\left\{\mathbb{E}\left(\sum_{j=t-1} ^{T}\gamma^{j-1}\rho_{1:j}^{\pi}(A_{i,j},S_{i,j})R_{i,j}|\mathcal{H}_{i,t}\right)|\mathcal{H}_{i,t-1}\right\}\right].
\end{align}
Since 
\begin{align}\label{conEt}
  \mathbb{E}\left(\sum_{j=t} ^{T}\gamma^{j-1}\rho_{1:j}^{\pi}(A_{i,j},S_{i,j})R_{i,j}|\mathcal{H}_{i,t}\right)=\gamma^{t-1}\rho_{1:t}^{\pi}(A_{i,t},S_{i,t})Q_{t}^{\pi,T}(A_{i,t},S_{i,t}).
\end{align}
Combing with \eqref{conEt} and \eqref{eqSIS},
\begin{align*}
&\mathbb{V}\left[\sum_{j=1} ^{T}\gamma^{j-1}\rho_{1:j}^{\pi}(A_{i,j},S_{i,j})R_{i,j}\right]\\=&\sum_{t=1}^{T+1}\mathbb{E}\left[\mathbb{V}\left\{\left(\rho_{1:t-1}^{\pi}(A_{i,t-1},S_{i,t-1})R_{i,t-1}\gamma^{t-1-1}+\gamma^{t-1}\rho_{1:t}^{\pi}(A_{i,t},S_{i,t})Q_t^{\pi,T}(A_{i,t},S_{i,t})\right)|\mathcal{H}_{i,t-1}\right\}\right]\\
=&\sum_{t=1}^{T+1}\mathbb{E}\left[\left\{\gamma^{t-1-1}\rho_{1:t-1}^{\pi}(A_{i,t-1},S_{i,t-1})\right\}^2\mathbb{V}\left\{R_{i,t-1}+\gamma\rho_{t}^{\pi}(A_{i,t},S_{i,t})Q_t^{\pi,T}(A_{i,t},S_{i,t})|\mathcal{H}_{i,t-1}\right\}\right]\\
=&\sum_{t=1}^{T+1}\mathbb{E}\left[\left\{\gamma^{t-1-1}\rho_{1:t-1}^{\pi}(A_{i,t-1},S_{i,t-1})\right\}^2\mathbb{V}\left\{R_{i,t-1}+\gamma\rho_{t}^{\pi}(A_{i,t},S_{i,t})Q_t^{\pi,T}(A_{i,t},S_{i,t})|A_{i,t-1},S_{i,t-1}\right\}\right]\\
=&\sum_{t=1}^{T}\mathbb{E}\left[\left\{\gamma^{t-1}\rho_{1:t}^{\pi}(A_{i,t},S_{i,t})\right\}^2\mathbb{V}\left\{R_{i,t}+\gamma\rho_{t+1}^{\pi}(A_{i,t+1},S_{i,t+1})Q_t^{\pi,T}(A_{i,t+1},S_{i,t+1})|A_{i,t},S_{i,t}\right\}\right].
\end{align*}
It follows from Assumptions \ref{asmp:bounded} and \ref{asmp:coverage} that $Q_t^{\pi,T}(A_{i,t},S_{i,t})$ is upper bounded by $R_{\max}(1-\gamma)^{-1}$. Then,
\begin{align*}
\mathbb{V}\left[R_{i,t}+\gamma\rho_{t+1}^{\pi}(A_{i,t+1},S_{i,t+1})Q_t^{\pi,T}(A_{i,t+1},S_{i,t+1})|A_{i,t},S_{i,t}\right]\lesssim \frac{4R_{\max}^2}{(1-\gamma)^2}.
\end{align*}
Thus, we have 
\begin{align*}
\mathrm{MSE}_{\mathrm{SIS}}
\leq\frac{4R_{\max}^2}{NT^2(1-\gamma)^2}\sum_{t=1}^T \gamma^{2t-2}\mathbb{E}(\rho_{1:t}^{\pi})^2
+\frac{\gamma^{2T}R_{\max}^2}{(1-\gamma)^2}.
\end{align*}
The above proof relies on the assumption that the behavior action is history-independent. In fact, if the behavior action is history-dependent, the result still holds. We simply need to replace \(S_{i,t}\) in the previous formula with the history up to time \(t\).
\end{itemize}
\subsection{Proof of Lemma \ref{kappa}}\label{proofofkappa}

\begin{itemize}[leftmargin=*]
    \item \textbf{The value of $\kappa_q$}.
Denote $\hat{Q}^{\pi}(a, s)=\varphi( a,s)^{\top} \hat{\beta}$ and ${Q}^{\pi}(a, s)=\varphi( a,s)^{\top} \beta^*$.
Then $\beta^*=\Sigma^{-1}\mathbb{E}[R\varphi(A,S)].$
By definition, we have

\begin{comment}

\begin{align}
&\widehat{\beta}-{\beta}^*\nonumber\\
=&\widehat{{\Sigma}}^{-1}\frac{1}{N T} \sum_{i=1}^N\sum_{j=1}^{T} {R_{i,j}\varphi(A_{i,j},S_{i,j})}-\Sigma^{-1}\mathbb{E}[R\varphi(A,S)]\nonumber\\
=&\left\{\widehat{{\Sigma}}^{-1}-\Sigma^{-1}\right\}\frac{1}{N T} \sum_{i=1}^N\sum_{j=1}^{T} {R_{i,j}\varphi(A_{i,j},S_{i,j})}-\Sigma^{-1}\left\{\frac{1}{N T} \sum_{i=1}^N\sum_{j=1}^{T} {R_{i,j}\varphi(A_{i,j},S_{i,j})}-\mathbb{E}[R\varphi(A,S)]\right\}.
\end{align}
Thus,
\begin{align*}
    \hat{Q}^{\pi}(a,s)-Q^{\pi}(a,s)=\varphi(a,s)^\top\left\{\widehat{{\Sigma}}^{-1}\frac{1}{N T} \sum_{i=1}^N\sum_{j=1}^{T} {R_{i,j}\varphi(A_{i,j},S_{i,j})}-\Sigma^{-1}\mathbb{E}[R\varphi(A,S)].\right\}.
\end{align*}
\end{comment}
\begin{align}
&\widehat{\beta}-{\beta}^*\nonumber\\
=&\widehat{{\Sigma}}^{-1}\frac{1}{N T} \sum_{i=1}^N\sum_{j=1}^{T} {R_{i,j}\varphi(A_{i,j},S_{i,j})}-\widehat{\Sigma}^{-1}\widehat{\Sigma}\beta^*\nonumber\\
=&\widehat{\Sigma}^{-1}\frac{1}{N T} \sum_{i=1}^N\sum_{j=1}^{T} \varphi(A_{i,j},S_{i,j})\{R_{i,j}-[\varphi(A_{i,j},S_{i,j})-\gamma\sum_{a}\pi(a|S_{i,j+1})\varphi(a,S_{i,j+1})]\beta^*\}\nonumber\\
=&\widehat{\Sigma}^{-1}\frac{1}{N T} \sum_{i=1}^N\sum_{j=1}^{T} \varphi(A_{i,j},S_{i,j})\{R_{i,j}-[\varphi(A_{i,j},S_{i,j})\beta^*-\gamma\sum_{a}\pi(a|S_{i,j+1})\varphi(a,S_{i,j+1})\beta^*]\}\nonumber\\
=&\widehat{\Sigma}^{-1}\frac{1}{N T} \sum_{i=1}^N\sum_{j=1}^{T} \varphi(A_{i,j},S_{i,j})\{R_{i,j}-Q^{\pi}(A_{i,j},S_{i,j})+\gamma\sum_{a}\pi(a|S_{i,j+1})Q^{\pi}(a,S_{i,j+1})\}\nonumber\\
=&\underbrace{\left\{\widehat{\Sigma}^{-1}-\Sigma^{-1}\right\}\frac{1}{N T} \sum_{i=1}^N\sum_{j=1}^{T} \varphi(A_{i,j},S_{i,j})\{R_{i,j}-Q^{\pi}(A_{i,j},S_{i,j})+\gamma\sum_{a}\pi(a|S_{i,j+1})Q^{\pi}(a,S_{i,j+1})\}}_{I_{1}}\nonumber\\
&+\underbrace{\Sigma^{-1}\frac{1}{N T} \sum_{i=1}^N\sum_{j=1}^{T} \varphi(A_{i,j},S_{i,j})\{R_{i,j}-Q^{\pi}(A_{i,j},S_{i,j})+\gamma\sum_{a}\pi(a|S_{i,j+1})Q^{\pi}(a,S_{i,j+1})\}}_{I_{2}}.
\end{align}
Then 
\begin{align}\label{Bbeta}
\mathbb{E}\Vert\hat{\beta}-\beta\Vert_2^2\leq \mathbb{E}I_1^2+\mathbb{E}I_2^2+2\sqrt{\mathbb{E}I_2^2 \mathbb{E}I_1^2}.
\end{align}
 Let $${\varepsilon}_{i, j}=\{R_{i,j}-Q^{\pi}(A_{i,j},S_{i,j})+\gamma\sum_{a}\pi(a|S_{i,j+1})Q^{\pi}(a,S_{i,j+1})\}.$$
 Let $\mathcal{F}_{i, t}$ denote the sub-$\sigma$-field generated by  $\left\{S_{i, t}, A_{i, t}\right\} \cup\left\{\left(R_{i, j}, A_{i, j}, S_{i, j}\right)\right\}_{1 \leq j<t}$. Since the data is Markov, we have
$$
\mathbb{E}\left(\varepsilon_{i, j} \mid \mathcal{F}_{i, j}\right)=\mathbb{E}\left(\varepsilon_{i, j} \mid S_{i, j}, A_{i, j}\right)=0.
$$
Notice that $\varphi(A_{i,j},S_{i,j})$ is a function of $S_{i, j}$ and $A_{i, j}$ only, we have for any $1 \leq j_1<j_2 \leq T$ that
$$
\mathbb{E}[\varepsilon_{i,j_1}\varepsilon_{i, j_2} \varphi(A_{i,j_1},S_{i,j_2})^{\top} \varphi(A_{i,j_2},S_{i,j_2})]=\mathbb{E}\{\varepsilon_{i, j_1}\varphi(A_{i,j_1},S_{i,j_1})^{\top}\varphi(A_{i,j_2},S_{i,j_2})\mathbb{E}\left(\varepsilon_{i, j_2} \mid \mathcal{F}_{i, j_2}\right)\}=0.
$$
By the independence assumption, we have for any $1 \leq j_1<j_2 \leq T$ and $1 \leq i_1<i_2 \leq n$ that $\mathbb{E} [\varepsilon_{i_1, j_1} \varepsilon_{i_2, j_2} \varphi(A_{i_1},S_{j_1})^{\top} \varphi(A_{i_2},S_{j_2})]=0$. It follows that

$$
\mathbb{E}\left\|\sum_{i=1}^N \sum_{j=1}^{T} \varphi(A_{i, j},S_{i,j}) \varepsilon_{i, j}\right\|_2^2=\sum_{i=1}^N \sum_{j=1}^{T} \mathbb{E} \varepsilon_{i, j}^2 \varphi_{i, t}^{\top} \varphi_{i, t}=N \sum_{j=1}^{T} \mathbb{E} \varepsilon_{1, j}^2 \varphi(A_{1, j},S_{1,j})^{\top} \varphi(A_{1, j},S_{1,j}).
$$
By Assumptions \ref{asmp:bounded}, we obtain
\begin{align}\label{BetaI1}
 \mathbb{E}\left\|\sum_{i=1}^N \sum_{j=1}^{T} \varphi(A_{i, j},S_{i,j}) \varepsilon_{i, j}\right\|_2^2\leq (NT)\frac{R_{\max}^2}{(1-\gamma)^2}\mathbb{E} [\varphi(A_{1, j},S_{1,j})^{\top} \varphi(A_{1, j},S_{1,j})].   
\end{align}
In the following, we try to  bound $\lambda_{\max}(\hat{\Sigma}^{-1}-\Sigma^{-1})$, where $\lambda_{max}$ denotes the singular values of a matrix.
Define ${U}(S)=\sum_{a}\pi(a|S)\varphi(a,S)$, and
$$
{\xi}_i=\frac{1}{T} \sum_{t=1}^{T} \varphi\left(A_{i,t},S_{i,t}\right)\left\{\varphi\left(A_{i,t},S_{i,t}\right)-\gamma {U}\left(S_{i, t+1}\right)\right\}^{\top} .
$$

Denote $L=\max _{a, s}\left\|\varphi\left(a,s\right)\right\|_2^2$, then $\max _{s}\left\|U\left(s\right)\right\|_2^2 \leq \sqrt{L}$. It follows that
\begin{align}\label{lambdaxi}
 \max _{1 \leq i \leq N}\lambda_{\max}\left(\xi_i-\mathbb{E} \xi_i\right) \leq \frac{2}{T} \sum_{t=1}^{T} \sqrt{L}\left(\sqrt{L}+\gamma \sqrt{L}\right) \leq 4 L.   
\end{align}
Let
\begin{align*}
\sigma^2 & =\max \left\{\lambda_{\max}\left(\sum_{i=1}^N \mathbb{E}\left(\xi_i-\mathbb{E} \xi_i\right)\left(\xi_i-\mathbb{E} \xi_i\right)^{\top}\right),\lambda_{\max}\left(\sum_{i=1}^N \mathbb{E}\left(\xi_i-\mathbb{E} \xi_i\right)^{\top}\left(\xi_i-\mathbb{E} \xi_i\right)\right)\right\}.
\end{align*}
By the eigenvalue properties, we can get $\sigma^2\leq 16NL^2$. Combining this together with \eqref{lambdaxi}, an application of the matrix concentration inequality (see Theorem 1.6 in \cite{tropp2012user}) yields that
$$
\mathbb{P}\left(\lambda_{\max}\left(\frac{1}{N}\sum_{i=1}^N(\xi_i-\Sigma)\right) \geq \tau\right) \leq 2 d \exp \left(-\frac{N^2\tau^2}{16NL^2+8 L \tau N/ 3}\right), \quad \forall \tau>0.
$$

Set $\tau=\sqrt{C N \log (N)}L$.  For sufficiently large $N$, we have $8 L\tau / 3 \ll \tau^2$ and hence

$$
\mathbb{P}\left(\lambda_{\max}\left(\sum_{i=1}^N\frac{1}{N}(\xi_i-\Sigma)\right) \geq  \sqrt{C N  \log(N)}\frac{L}{N}\right) \leq \frac{2 d}{N^4}.
$$

Since $d \ll N$, we obtain $2 d / N^4 \ll 1 /\left(N^2 \right)$. Thus, we can show that the following event occurs with probability at least $1-O\left(N^{-2}\right)$,
\begin{align}\label{sigmabias}
\lambda_{\max}(\widehat{\Sigma}-\Sigma)=\lambda_{\max}\frac{1}{N}\sum_{i=1}^N \left(\xi_i-\Sigma\right)\leq O(\sqrt{N^{-1}  \log(N)}L).
\end{align}
Now, we have shown $\lambda_{\max}(\widehat{\Sigma}-\Sigma)=O_p(\sqrt{N^{-1}  \log(N)}L)$. Under the condition that $L=o\{\sqrt{N / \log (N)}\}$, we have $\lambda_{\max}(\widehat{\Sigma}-\Sigma)=o_p(1)$. By definition, this implies that $\lambda_{\max}(\widehat{\Sigma}-\Sigma) \leq c / 2$, with probability  $1-O(N^{-2})$. If follows from Assumption \ref{asmp:positive}, and Cauchy-Schwarz inequality, we have
that $\widehat{\Sigma}$ is invertible and satisfies $\lambda_{\max}\left(\widehat{\Sigma}^{-1}\right) \leq 2 {c}^{-1}$, with probability $1-O(N^{-2})$. Therefore
\begin{align*}
\lambda_{\max}\left(\widehat{\Sigma}^{-1}-\Sigma^{-1}\right)=&\lambda_{\max}\left(\widehat{\Sigma}^{-1}(\widehat{\Sigma}-\Sigma) \Sigma^{-1}\right)\\
\leq&\lambda_{\max}\left(\widehat{\Sigma}^{-1}\right)\lambda_{\max}(\widehat{\Sigma}-\Sigma)\lambda_{\max}\left(\Sigma^{-1}\right)\\
\leq &2 {c}^{-2}\lambda_{\max}(\widehat{\Sigma}-\Sigma)
\end{align*}
with probability  $1-O(N^{-2})$. By \eqref{sigmabias}, we obtain $\lambda_{\max}\left(\widehat{\Sigma}^{-1}-\Sigma^{-1}\right)\leq O\left\{LN^{-1 / 2} \log (N )\right\}$ with probability $1-O(N^{-2})$. 
Combing \eqref{lambdaxi}, we can get 
\begin{align}\label{eqF4}
 \mathbb{E}[I_1^2]\leq \{LN^{-1 / 2} \log (N)+\frac{L}{N^{2}}\}\frac{1}{NT}\frac{R_{\max}^2}{(1-\gamma)^2}\mathbb{E} [\varphi(A_{1, j},S_{1,j})^{\top} \varphi(A_{1, j},S_{1,j})].
\end{align}

It follows from \eqref{BetaI1} and Assumption \eqref{asmp:positive} that
\begin{align}\label{eqF5}
    \mathbb{E}[I_2^2]\leq c^{-2}\frac{1}{NT}\frac{R_{\max}^2}{(1-\gamma)^2}\mathbb{E} [\varphi(A_{1, j},S_{1,j})^{\top} \varphi(A_{1, j},S_{1,j})].
\end{align}
Combing \eqref{Bbeta}, \eqref{eqF4} and \eqref{eqF5} that
\begin{align}\label{Betasq}
\mathbb{E}\Vert\hat{\beta}-\beta\Vert_2^2\leq\frac{1}{NT}\frac{R_{\max}^2}{(1-\gamma)^2}\mathbb{E} [\varphi(A_{1, j},S_{1,j})^{\top} \varphi(A_{1, j},S_{1,j})].
\end{align}
It follows from the definition of $Q^\pi$ and $\hat{Q}^\pi$, we can get
\begin{align}
\mathbb{E}\left\{(\hat{Q}^{\pi}(a,s)-Q^\pi(a,s))^2\right\}&\leq \varphi(a,s)^\top\varphi(a,s)\mathbb{E}\lambda_{\max}[(\hat{\beta}-\beta)(\hat{\beta}-\beta)^\top]\nonumber\\
=&\varphi(a,s)^\top\varphi(a,s)\frac{1}{NT}\frac{R_{\max}^2}{(1-\gamma)^2}\mathbb{E} [\varphi(A_{1, j},S_{1,j})^{\top} \varphi(A_{1, j},S_{1,j})].
\end{align}
Thus, we can get 
\begin{align*}
&\mathbb{E}_{A,S}\mathbb{E}\left\{(\hat{Q}^{\pi}(A,S)-Q^\pi(A,S))^2\right\}\\
\leq& \mathbb{E}_{A,S}[\varphi(A,S)^\top\varphi(A,S)]\frac{1}{NT}\frac{R_{\max}^2}{(1-\gamma)^2}\mathbb{E} [\varphi(A_{1, j},S_{1,j})^{\top} \varphi(A_{1, j},S_{1,j})].
\end{align*}
It follows from the definition of $\kappa_q$, we have
\begin{align}\label{kapaq}
\kappa_q=\sqrt{\frac{1}{NT}\mathbb{E}_{A,S}[\varphi(A,S)^\top\varphi(A,S)]\mathbb{E} [\varphi(A_{1, j},S_{1,j})^{\top} \varphi(A_{1, j},S_{1,j})]}.
\end{align}
In tabular mode, we have
$\mathbb{E}_{A,S}[\varphi(A,S)^\top\varphi(A,S)]\leq d=|\mathcal{S}||\mathcal{A}|$, where $|\cdot|$ denotes the cardinality of a set.
Then, we have 
\begin{align}\label{kapaq2}
\kappa_q=\sqrt{\frac{1}{NT}}|\mathcal{S}||\mathcal{A}|.
\end{align}

\item \textbf{The value of $\kappa_w$}.

Denote ${w}^\pi(a,s)=\varphi(a,s)^{\top}\alpha^*$ with $\alpha^*$ satisfying $\alpha^*=\Sigma^{-1}(1-\gamma)\mathbb{E}[\sum_{a}\pi(a|S_{i,1})\varphi(a,S_{i,1})].$
By definition, we have
\begin{align}\label{EqF9}
&\widehat{\alpha}-{\alpha}^*\nonumber\\
=&\widehat{{\Sigma}}^{-1}(1-\gamma) \frac{1}{N}\sum_{i=1}^{N}\left[\sum_{a}\pi(a|S_{i,1})\varphi\left(a, S_{i,1}\right)\right]-\hat{\Sigma}^{-1}\hat{\Sigma}\alpha^*\nonumber\\
=&\widehat{\Sigma}^{-1}\frac{1}{N T} \sum_{i=1}^N\sum_{j=1}^{T}\left\{(1-\gamma) \sum_{a}\pi(a|S_{i,1})\varphi(a, S_{i,1})-w^{\pi}(A_{i,j},S_{i,j})\right.\nonumber\\
&\left.\times[\varphi(A_{i,j},S_{i,j})-\gamma\sum_{a}\pi(a|S_{i,j+1})\varphi(a,S_{i,j+1})]\right\}\nonumber\\
=&\left\{\widehat{\Sigma}^{-1}-\Sigma^{-1}\right\}\frac{1}{N T} \sum_{i=1}^N\sum_{j=1}^{T}L_{i,j}+\Sigma^{-1}\frac{1}{N T} \sum_{i=1}^N\sum_{j=1}^{T}L_{i,j},
\end{align}
where
\begin{align*}
  L_{i,j}=& \left\{(1-\gamma) \sum_{a}\pi(a|S_{i,1})\varphi(a, S_{i,1})-w^{\pi}(A_{i,j},S_{i,j})\right.\nonumber\\
&\left.\times[\varphi(A_{i,j},S_{i,j})-\gamma\sum_{a}\pi(a|S_{i,j+1})\varphi(a,S_{i,j+1})]\right\}.
\end{align*}
Then, direct calculations yield 
\begin{align*}
\mathbb{V}[\frac{1}{N T} \sum_{i=1}^N\sum_{j=1}^{T}L_{i,j}]=&\frac{1}{N}\mathbb{V}[\frac{1}{ T} \sum_{j=1}^{T}L_{i,j}]\\
=&\frac{1}{N}\mathbb{V}\left[(1-\gamma) \sum_{a}\pi(a|S_{i,1})\varphi(a, S_{i,1})-\frac{1}{T} \sum_{j=1}^{T}w^{\pi}(A_{i,j},S_{i,j})\right.\nonumber\\
&\left.\times[\varphi(A_{i,j},S_{i,j})-\gamma\sum_{a}\pi(a|S_{i,j+1})\varphi(a,S_{i,j+1})]\right].
\end{align*}
Then by similar arguments as \eqref{kapaq}, we can get
\begin{align}\label{kapaw}
\kappa_w=\sqrt{\frac{1}{N}\mathbb{E}_{A,S}\varphi(A,S)^\top\varphi(A,S)\mathbb{E} [\varphi(A_{i,j},S_{i,j})^\top\varphi(A_{i,j},S_{i,j})]}.
\end{align}
In tabular mode, we have 
\begin{align}\label{kapaw2}
\kappa_w=\sqrt{\frac{1}{N}}|\mathcal{S}||\mathcal{A}|.
\end{align}
    
\end{itemize}
\subsection{Proof of Theorem \ref{Tpostmse}}
In this subsection, we will prove each part of Theorem \ref{Tpostmse} individually.
Recall that $\phi$ denote an MSA from $\mathcal{S}$ to $\mathcal{X}$ and $\phi^*$ denote another MSA from $\mathcal{X}$ to certain $\mathcal{X}^*=\{\phi^*(x):x\in \mathcal{X}\}$. Then we have:
 \begin{itemize}
 \item When lookup tables are employed to parameterize the Q-function and MIS ratio, under certain matrix invertibility condition detailed in Appendix~\ref{sec:proofs}, we have
         \begin{eqnarray}\label{eqpostkappa}
\kappa_q(\phi)=O\Big(\frac{{|\mathcal{X}||\mathcal{A}|}}{\sqrt{NT}}\Big)\,\,\hbox{and}\,\,\kappa_w(\phi)=O\Big(\frac{{|\mathcal{X}||\mathcal{A}|}}{\sqrt{N}}\Big).
     \end{eqnarray}
This result directly follows from Lemma~\ref{kappa}.
     \item In this section, we first prove that for any MSA $\phi$, we have $\textrm{Var}^{(1)}\geq\textrm{Var}^{(1)}(\phi)$ with infinite horizon $T$,
where \(\textrm{Var}^{(1)}\) denotes the MSE of DRL applied to the original ground state space.
It follows from \eqref{eqn:DRMSE} that, for any MSA $\phi$, the MSE of DRL is
\begin{align*}
\textrm{Var}^{(1)}(\phi)
=&\frac{1}{N}\mathbb{V}\left[V_{\phi}^{\pi}\left(\phi(S)\right)\right]+\frac{1}{NT(1-\gamma)^2}\mathbb{V}\left[w_{\phi}^{\pi}(A,\phi(S)) \left(R+\gamma V_{\phi}^{\pi}\left(\phi(S')\right)-Q_{\phi}^{\pi}(A,\phi(S))\right)\right]\nonumber\\
&+O\Big(\frac{R_{\max}^2\max(\kappa_{q}(\phi),\kappa_w(\phi),N^{-1/2})\log (N)}{(1-\gamma)^2N}+\frac{R_{\max}^2\max(\kappa_{q}(\phi)\kappa_{w}(\phi), \sqrt{N}\kappa_{q}^2(\phi)\kappa_{w}^2(\phi))}{\sqrt{N}(1-\gamma)^{2}}\Big)
\end{align*}
Under an infinite horizon, we have 
\begin{align*}
   \textrm{Var}^{(1)}(\phi)
=&\frac{1}{N}\mathbb{V}\left[V_{\phi}^{\pi}\left(\phi(S)\right)\right]\\
&+O\Big(\frac{R_{\max}^2\max(\kappa_{q}(\phi),\kappa_w(\phi),N^{-1/2})\log (N)}{(1-\gamma)^2N}+\frac{R_{\max}^2\max(\kappa_{q}(\phi)\kappa_{w}(\phi), \sqrt{N}\kappa_{q}^2(\phi)\kappa_{w}^2(\phi))}{\sqrt{N}(1-\gamma)^{2}}\Big). 
\end{align*}
 A direct calculation yields$V_{\phi}^{\pi}\left(\phi(S)\right)=\mathbb{E}\left[V^{\pi}\left(S\right)|\phi(S)\right]$. Combing with $$\mathbb{V}\left[\mathbb{E}\left\{V^{\pi}\left(S\right)|\phi(S)\right\}\right]+\mathbb{E}\left[\mathbb{V}\left\{V^{\pi}(S)|\Phi(S)\right\}\right]=\mathbb{V}\left[V^{\pi}\left(S\right)\right],$$ we have
 \begin{align*}
\mathbb{V}\left[V_{\phi}^{\pi}\left(\phi(S)\right)\right]+ \mathbb{E}\left[\mathbb{V}\left\{V^{\pi}(S)|\Phi(S)\right\}\right]=\mathbb{V}\left[V^{\pi}\left(S\right)\right].
 \end{align*}

 This means that after one Markov state abstraction, the MSE of DRL will decrease, that is, $\displaystyle\lim_{T\to \infty} \textrm{Var}^{(1)}\ge \lim_{T\to \infty} \textrm{Var}^{(1)}(\phi)$.

 Since both  $\phi$ and $\phi^*$ are MSA, then $\phi^*\circ\phi_1$ is also an MSA. So we have  $\displaystyle\lim_{T\to \infty} \textrm{Var}^{(1)}(\phi)\ge \lim_{T\to \infty} \textrm{Var}^{(1)}(\phi^*\circ \phi)$. 
     
  Additionally, from Lemma~\ref{kappa} and equation \eqref{eqpostkappa}, we can conclude that \(\kappa_q\) and \(\kappa_w\) are proportional to \(|\mathcal{S}|\), while \(\kappa_q(\phi)\) and \(\kappa_w(\phi)\) depend on \(|\mathcal{\phi(S)}|\), meaning the bias term will also decrease. 
      
     \item 
We know that the conditional expectation \(\mathbb{E}[\rho_{1:t}^{\pi}(A_{i,j},S_{i,j})|A_{i,j},\phi(S_{i,j})]=\rho_{\phi,1:t}^{\pi}(A_{i,j},\phi(S_{i,j}))\) holds.

By applying Jensen's inequality,  we can derive the following result:
\[
\mathbb{E}[(\rho_{1:j}^{\pi})^2(A_{i,j},S_{i,j})] \geq \mathbb{E}[(\rho_{\phi,1:j}^{\pi})^2(A_{i,j},\phi(S_{i,j}))] 
\]
This inequality implies that after performing one Markov state abstraction operation, the upper bound of the MSE for the SIS method will decrease.
Hence, by deduction, we have
 $\textrm{Var}^{(2)}(\phi)\ge\textrm{Var}^{(2)}(\phi^*\circ \phi)$.  
     \item The bias term in \eqref{eqn:ISMSE} directly follows Assumption \ref{asmp:bounded} and \ref{asmp:coverage}, which is independent of $\phi$. 
 \end{itemize}

\subsection{Proof of Lemma \ref{backMSA}}\label{sec:MSA}
We provide the more complex version, where the type-II MSA \(\phi\) is obtained by applying the refined backward-model-irrelevance condition to the original MDP \((S_t, A_t, R_t,S_{t+1})_{t \geq 1}\) with a history-dependent behavior policy. In this case, the reduced process \((\phi(S_t), A_t, R_t,\phi(S_{t+1}))_{t \geq 1}\) remains an MDP. The type-II MSA \(\phi\), obtained by applying the original condition, directly follows from this refinement.

We start by presenting the following lemma and its proof. 
 \begin{lemma}\label{lemmaE2}
 For any $a,x,t$ and $s_{t+1}$, $x_{t+1}$ such that  $\phi(s_{t+1})=x_{t+1}$, we have 
 \begin{align}
 \sum_{s\in \phi^{-1}(x)}\mathbb{P}(A_t=a,S_t=s|S_{t+1}=s_{t+1})=\mathbb{P}(A_t=a,\phi(S_t)=x|\phi(S_{t+1})=x_{t+1})\label{back_state}.
  \end{align}
 Additionally, for any $a_t$, $s_{t}$, $x_{t}$ such that  $\phi(s_{t+1})=x_{t+1}$, we have 
 \begin{align}\label{action_history}
 \frac{\mathbb{P}(A_t=a_t|S_t=s_t)}{\mathbb{P}(A_t=a_t|\phi(S_t)=x_t)}=\frac{\mathbb{P}(A_t=a_t|S_t=s_t,\{A_{t-k}=a_{t-k},\phi(S_{t-k})=x_{t-k}\}_{k \in G})}{\mathbb{P}(A_t=a_t|\phi(S_t)=x_t,\{A_{t-k}=a_{t-k},\phi(S_{t-k})=x_{t-k}\}_{k\in G})},
 \end{align}
 for any $G=\{1,2,\ldots,\ell\}$ with any $\ell\in \{1,2,\ldots,t-1\}$.
\end{lemma}
\textbf{Proof of Lemma \ref{lemmaE2}}. Using similar arguments to \eqref{eqn:rhoirrelevance2}, we have
\begin{align}\label{eqn:proofback_state}
\begin{split}
 &\mathbb{P}(A_t=a,\phi(S_t)=x|\phi(S_{t+1})=x_{t+1})\\
 =& \frac{\mathbb{P}(A_t=a,\phi(S_t)=x,\phi(S_{t+1})=x_{t+1})}{\mathbb{P}(\phi(S_{t+1})=x_{t+1})}\\
 =&\sum_{s_{t+1}\in \phi^{-1}(x_{t+1})}\frac{\mathbb{P}(A_t=a,\phi(S_t)=x,S_{t+1}=s_{t+1})}{\mathbb{P}(\phi(S_{t+1})=x_{t+1})}\\
 =&\sum_{s'_{t+1}\in \phi^{-1}(x_{t+1})}\mathbb{P}(A_t=a,\phi(S_t)=x|S_{t+1}=s_{t+1})\mathbb{P}(S_{t+1}=s'_{t+1}|\phi(S_{t+1})=x_{t+1})\\
  =&\mathbb{P}(A_t=a,\phi(S_t)=x|S_{t+1}=s_{t+1})\\
  =&\sum_{s\in \phi^{-1}(x)}\mathbb{P}(A_t=a,S_t=s|S_{t+1}=s_{t+1}),
\end{split}
\end{align}
where the third equation is due to the backward-transition-irrelevance \eqref{eqn:backirrelevance-state} condition, under which $\mathbb{P}(A_t=a,\phi(S_t)=x|S_{t+1}=s_{t+1})$ equals $\mathbb{P}(A_t=a,\phi(S_t)=x|S_{t+1}=s_{t+1}')$. 
This proves \eqref{back_state}. 

Next, under the stationarity assumption in Assumption \ref{asmp:stationary} and the history-dependent-behavior-policy-irrelevance condition, we have for any $\{s_\ell^{(1)}\}_\ell$, $\{s_\ell^{(2)}\}_t$ such that $\phi(s_{\ell}^{(1)})=\phi(s_{\ell}^{(2)})=x_{\ell}$ for all $\ell\ge 1$ that
\begin{align}\label{eqn:proofbackwardMSA}
\begin{split}
   &\mathbb{P}(A_{t}=a_{t}|S_{t}=s_{t}^{(1)},\{A_{t-k}=a_{t-k},S_{t-k}=s_{t-k}^{(1)}\}_{k\in G}))\\ =&\mathbb{P}(A_{t}=a_{t}|S_{t}=s^{(2)}_{t},\{A_{t-k}=a_{t-k},S_{t-k}=s_{t-k}^{(2)}\}_{k\in G}),
\end{split}
\end{align}
for any $t$, $\{a_{\ell}\}_{\ell}$ and $G$. This in turn yields,
\begin{align}\label{state_history_G}
\begin{split}
   &\frac{\mathbb{P}(A_{t}=a_{t}|S_{t}=s_{t}^{(1)})}{\mathbb{P}(A_{t}=a_{t}|S_{t}=s_{t}^{(2)})}\\ =&\frac{\mathbb{P}(A_{t}=a_{t}|S_{t}=s_{t}^{(1)},\{A_{t-k}=a_{t-k},S_{t-k}=s_{t-k}^{(1)}\}_{G}))}{\mathbb{P}(A_{t}=a_{t}|S_{t}=s_{t}^{(2)},\{A_{t-k}=a_{t-k},S_{t-k}=s_{t-k}^{(2)}\}_{G})}.
\end{split}
\end{align}
%Next, it follows from \eqref{eqn:proofbackwardMSA} that
With some calculations, we have
\begin{align}\label{Bayesian_one}
\begin{split}
&\mathbb{P}(A_t=a_t|S_t=s_t,\{A_{t-k}=a_{t-k},\phi(S_{t-k})=x_{t-k}\}_{k \in G})\\
=&\frac{\mathbb{P}(A_t=a_t,\{A_{t-k}=a_{t-k},\phi(S_{t-k})=x_{t-k}\}_{k \in G}|S_t=s_t)}{\mathbb{P}(\{A_{t-k}=a_{t-k},\phi(S_{t-k})=x_{t-k}\}_{k \in G}|S_t=s_t)}\\
=&\sum_{s_{t-k}\in \phi^{-1}(x_{t-k}),k\in G}\frac{\mathbb{P}(A_t=a_t,\{A_{t-k}=a_{t-k},S_{t-k}=s_{t-k}\}_{k \in G}|S_t=s_t)}{\mathbb{P}(\{A_{t-k}=a_{t-k},\phi(S_{t-k})=x_{t-k}\}_{k \in G}|S_t=s_t)}\\
=&\sum_{s_{t-k}\in \phi^{-1}(x_{t-k}),k\in G}\frac{\mathbb{P}(\{A_{t-k}=a_{t-k},S_{t-k}=s_{t-k}\}_{k \in G}|S_t=s_t)}{\mathbb{P}(\{A_{t-k}=a_{t-k},\phi(S_{t-k})=x_{t-k}\}_{k \in G}|S_t=s_t)}\\
&\times  \mathbb{P}(A_t=a_t|S_t=s_t,\{A_{t-k}=a_{t-k},S_{t-k}=s_{t-k}^{(1)}\}_{k \in G})\\
=& \mathbb{P}(A_t=a_t|S_t=s_t,\{A_{t-k}=a_{t-k},S_{t-k}=s_{t-k}^{(1)}\}_{k \in G}),
\end{split}
\end{align}
where the last equation follows from \eqref{eqn:proofbackwardMSA}. 

Combing \eqref{state_history_G} with \eqref{Bayesian_one}, we obtain that
\begin{align*}
&\frac{\mathbb{P}(A_t=a_t|S_t=s_t^{(1)},\{A_{t-k}=a_{t-k},\phi(S_{t-k})=x_{t-k}\}_{k \in G})}{\mathbb{P}(A_t=a_t|S_t=s_t^{(2)},\{A_{t-k}=a_{t-k},\phi(S_{t-k})=x_{t-k}\}_{k \in G})}\nonumber\\
=&\frac{\mathbb{P}(A_t=a_t|S_t=s_t^{(1)},\{A_{t-k}=a_{t-k},S_{t-k}=s_{t-k}^{(1)}\}_{k \in G})}{\mathbb{P}(A_t=a_t|S_t=s_t^{(2)},\{A_{t-k}=a_{t-k},S_{t-k}=s_{t-k}^{(2)}\}_{k \in G})}\nonumber\\
=&\frac{\mathbb{P}(A_{t}=a_t|S_{t}=s_t^{(1)})}{\mathbb{P}(A_{t}=a_t|S_{t}=s_t^{(2)})},
\end{align*}
or equivalently, 
\begin{align}\label{action_statemap1}
\begin{split}
&\frac{\mathbb{P}(A_t=a_t|S_t=s_t^{(1)})}{\mathbb{P}(A_t=a_t|S_t=s_t^{(1)},\{A_{t-k}=a_{t-k},\phi(S_{t-k})=x_{t-k}\}_{k \in G})}\\
=&\frac{\mathbb{P}(A_t=a_t|S_t=s_t^{(2)})}{\mathbb{P}(A_t=a_t|S_t=s_t^{(2)},\{A_{t-k}=a_{t-k},\phi(S_{t-k})=x_{t-k}\}_{k \in G})}.
\end{split}
\end{align}
Using similar arguments to \eqref{eqn:rhoirrelevance2} and \eqref{eqn:proofback_state}, the LHS can be represented by 
\begin{eqnarray*}
    \frac{\mathbb{P}(A_t=a_t|\phi(S_t)=x_t)}{\mathbb{P}(A_t=a_t|\phi(S_t)=x_t,\{A_{t-k}=a_{t-k},\phi(S_{t-k})=x_{t-k}\}_{k \in G})}
\end{eqnarray*}
\eqref{action_history} follows directly from \eqref{action_statemap1}.

\textbf{Proof of the Markov property}. We next prove that the refined type-II   abstraction is indeed an MSA, despite that the behavior policy is no longer Markovian. Toward that end, we first show that the evolution of $\phi(S_t)$ remains Markovian. Specifically, we aim to show
\begin{align}\label{action_independent}
   (A_{t-k},\phi(S_{t-k}))_{1\leq k\leq t-1}
   \indep S_{t+1}|(\phi(S_{t}),A_{t}).
\end{align}
Indeed, by setting the time index $t$ in \eqref{back_state} to $t+1$, we obtain that 
\begin{align}\label{belief}
\frac{\mathbb{P}(S_t=s_t|A_{t-1}=a_{t-1},\phi(S_{t-1})=x_{t-1})}{\mathbb{P}(\phi(S_t)=x_t|A_{t-1}=a_{t-1},\phi(S_{t-1})=x_{t-1})}=\frac{\mathbb{P}(S_t=s_t)}{\mathbb{P}(\phi(S_t)=x_t)}.
\end{align}
Combing \eqref{belief} with \eqref{action_history}, we have
\begin{align}\label{markov_one}
&\frac{\mathbb{P}(S_t=s_t|A_{t-1}=a_{t-1},\phi(S_{t-1})=x_{t-1})\mathbb{P}(A_t=a_t|S_t=s_t,A_{t-1}=a_{t-1},\phi(S_{t-1})=x_{t-1})}{\mathbb{P}(\phi(S_t)=x_t|A_{t-1}=a_{t-1},\phi(S_{t-1})=x_{t-1})\mathbb{P}(A_t=a_t|\phi(S_t)=x_t,A_{t-1}=a_{t-1},\phi(S_{t-1})=x_{t-1})}\nonumber\\
=&\frac{\mathbb{P}(S_t=s_t)\mathbb{P}(A_t=a_t|S_t=s_t)}{\mathbb{P}(\phi(S_t)=x_t)\mathbb{P}(A_t=a_t|\phi(S_t)=x_t)},
\end{align}
or equivalently,
\begin{align}\label{eqn:someusefulequation}
\begin{split}
&\frac{\mathbb{P}(A_t=a_t,S_t=s_t|A_{t-1}=a_{t-1},\phi(S_{t-1})=x_{t-1})}{\mathbb{P}(A_t=a_t,\phi(S_t)=x_t|A_{t-1}=a_{t-1},\phi(S_{t-1})=x_{t-1})}\\
=&\frac{\mathbb{P}(A_t=a_t,S_t=s_t)}{\mathbb{P}(A_t=a_t,\phi(S_t)=x_t)}.
\end{split}
\end{align}
Since the original process $(S_t,A_t,R_t)_{t\ge 1}$ is an MDP, we have
\begin{align*}
&\frac{\mathbb{P}(A_t=a_t,S_t=s_t|A_{t-1}=a_{t-1},\phi(S_{t-1})=x_{t-1})}{\mathbb{P}(A_t=a_t,\phi(S_t)=x_t|A_{t-1}=a_{t-1},\phi(S_{t-1})=x_{t-1})}\\
&\times \mathbb{P}(S_{t+1}=s_{t+1}|A_t=a_t,S_t=s_t,A_{t-1}=a_{t-1},\phi(S_{t-1})=x_{t-1})\\
=&\frac{\mathbb{P}(S_{t+1}=s_{t+1}|A_t=a_t,S_t=s_t)\mathbb{P}(A_t=a_t,S_t=s_t)}{\mathbb{P}(A_t=a_t,\phi(S_t)=x_t)},
\end{align*}
leading to
\begin{align}\label{state_one}
\begin{split}
& \mathbb{P}(S_{t+1}=s_{t+1}|A_t=a_t,\phi(S_t)=x_t,A_{t-1}=a_{t-1},\phi(S_{t-1})=x_{t-1})\\
=&\mathbb{P}(S_{t+1}=s_{t+1}|A_t=a_t,\phi(S_t)=x_t).
\end{split}
\end{align}
 Equation \eqref{state_one} implies that when $k=1$, \eqref{action_independent} holds. 
 
Furthermore, by summing over $s_{t+1}\in \phi^{-1}(x_{t+1})$ on both sides of \eqref{state_one}, we obtain that
\begin{align*}
&\mathbb{P}(\phi(S_{t+1})=x_{t+1}|A_t=a_t,\phi(S_t)=x_t,A_{t-1}=a_{t-1},\phi(S_{t-1})=x_{t-1})\\
=&\mathbb{P}(\phi(S_{t+1})=x_{t+1}|A_t=a_t,\phi(S_t)=x_t).
\end{align*}
This together with \eqref{state_one} yields
\begin{align*}
& \frac{\mathbb{P}(S_{t+1}=s_{t+1}|A_t=a_t,\phi(S_t)=x_t,A_{t-1}=a_{t-1},\phi(S_{t-1})=x_{t-1})}{\mathbb{P}(\phi(S_{t+1})=x_{t+1}|A_t=a_t,\phi(S_t)=x_t,A_{t-1}=a_{t-1},\phi(S_{t-1})=x_{t-1})}\nonumber\\
=&\frac{\mathbb{P}(S_{t+1}=s_{t+1}|A_t=a_t,\phi(S_t)=x_t)}{\mathbb{P}(\phi(S_{t+1})=x_{t+1}|A_t=a_t,\phi(S_t)=x_t)}=\frac{\mathbb{P}(S_{t+1}=s_{t+1})}{\mathbb{P}(\phi(S_{t+1})=x_{t+1})},
\end{align*}
where the last equation again, follows from the backward-transition-irrelevance \eqref{eqn:backirrelevance-state}. 

Under the stationarity assumption, it leads to 
\begin{align*}
& \frac{\mathbb{P}(S_{t}=s_{t}|A_{t-1}=a_{t-1},\phi(S_{t-1})=x_{t-1},A_{t-2}=a_{t-2},\phi(S_{t-2})=x_{t-2})}{\mathbb{P}(\phi_2(S_{t})=x_{t}|A_{t-1}=a_{t-1},\phi(S_{t-1})=x_{t-1},A_{t-2}=a_{t-2},\phi(S_{t-2})=x_{t-2})}\nonumber\\
=&\frac{\mathbb{P}(S_{t}=s_{t})}{\mathbb{P}(\phi(S_{t})=x_{t})}.
\end{align*}
Applying the same arguments can be repeatedly for $t-2$ times, we obtain that
\begin{align}\label{state_k}
\begin{split}
& \frac{\mathbb{P}(S_{t}=s_{t}|\{A_{t-k}=a_{t-k},\phi(S_{t-k})=x_{t-k}\}_{1\leq k\leq t-1})}{\mathbb{P}(\phi_2(S_{t})=x_{t}|\{A_{t-k}=a_{t-k},\phi(S_{t-k})=x_{t-k}\}_{1\leq k\leq t-1})}\\
=&\frac{\mathbb{P}(S_{t}=s_{t})}{\mathbb{P}(\phi(S_{t})=x_{t})}.
\end{split}
\end{align}
Now, using the same arguments to \eqref{markov_one} and \eqref{state_one}, we obtain 
\begin{align}\label{state_last}
\begin{split}
& \mathbb{P}(S_{t+1}=s_{t+1}|\{A_{t-k}=a_{t-k},\phi(S_{t-k})=x_{t-k}\}_{0\leq k\leq t-1})\\
=&\mathbb{P}(S_{t+1}=s_{t+1}|A_t=a_t,\phi(S_t)=x_t).
\end{split}
\end{align}
This proves \eqref{action_independent}. It is immediate to see that \eqref{action_independent} yields
\begin{align*}
   (A_{t-k},\phi(S_{t-k}))_{1\leq k\leq t-1}
   \indep \phi(S_{t+1})|(\phi(S_{t}),A_{t}),
\end{align*}
which implies that the evolution of $\{\phi(S_t)\}_t$ is Markovian. 

Next, we demonstrate that the reward function when confined to the abstract state space also satisfies the  Markov property. Similar to \eqref{eqn:someusefulequation}, by combining \eqref{state_k} and \eqref{action_history}, we obtain that 
\begin{align}\label{sta_action_last}
\begin{split}
& \frac{\mathbb{P}(A_{t}=a_{t},S_{t}=s_{t}|\{A_{t-k}=a_{t-k},\phi(S_{t-k})=x_{t-k}\}_{1\leq k\leq t-1})}{\mathbb{P}(A_{t}=a_{t},\phi(S_{t})=x_{t}|\{A_{t-k}=a_{t-k},\phi(S_{t-k})=x_{t-k}\}_{1\leq k\leq t-1})}\\
=&\frac{\mathbb{P}(A_{t}=a_{t},S_{t}=s_{t})}{\mathbb{P}(A_{t}=a_{t},\phi(S_{t})=x_{t})}.
\end{split}
\end{align}
Notice that in the original MDP, the reward function satisfies the Markov property, i.e., the conditional mean of the reward is independent of $\{A_{t-k},\phi(S_{t-k})\}_{1\leq k\leq t-1})$, given $A_t$ and $S_t$. Consequently, we can multiply the $\sum_r r \mathbb{P}(R_t=r|A_t=a_t,S_t=s_t)$ on both sides of \eqref{sta_action_last} and obtain that
\begin{align*}
&\frac{\mathbb{E}[R_{t}\mathbb{I}(A_{t}=a_{t},S_{t}=s_{t})|\{A_{t-k}=a_{t-k},\phi(S_{t-k})=x_{t-k}]\}_{1\leq k\leq t-1}}{\mathbb{P}(A_{t}=a_{t},\phi(S_{t})=x_{t}|\{A_{t-k}=a_{t-k},\phi(S_{t-k})=x_{t-k}\}_{1\leq k\leq t-1})}\\
=&\frac{\mathbb{E}[R_{t}\mathbb{I}(A_{t}=a_{t},S_{t}=s_{t})]}{\mathbb{P}(A_{t}=a_{t},\phi(S_{t})=x_{t})}.
\end{align*}
By summing $s_t$ over $\phi^{-1}(x_t)$ on both sides of the equation, we obtain
\begin{align*}
& \mathbb{E}(R_{t}|A_{t}=a_{t},\phi(S_{t})=x_{t},\{A_{t-k}=a_{t-k},\phi(S_{t-k})=x_{t-k}\}_{0\leq k\leq t-1})\nonumber\\
=&\mathbb{E}(R_{t}|A_{t}=a_{t},\phi(S_{t})=x_{t}).
\end{align*}
This proves the Markov property of the reward function when restricted to the abstract state space. The proof is hence completed.

%where the first equation follows from Part III of Theorem \ref{thm2} and the third equation follows from Part I of Theorem \ref{thm4}.
%}
%\end{itemize}

%According to the DRL's double robustness property, its Fisher consistency is also proven given MIS and Q-function-based estimator's Fisher consistency.
%Then we completes the proof of Theorem \ref{thm4}.

\section{Toy examples}\label{sec:toy}

To elaborate the usefulness of DSA in reducing state cardinality, we analyze two examples: a bandit example and an MDP example. In both examples, we focus on a specific type of state abstraction known as variable selection, which selects a sub-vector from the original state.
Additionally, we focus on the class of state-agnostic target policies where $\pi$ is independent of the states. This type of policy is prevalent in causal inference and A/B testing, where the objective is to learn the global treatment effect of applying either a new or old policy consistently over time \citep[see e.g.,][]{hu2022switchback,leung2022rate,bojinov2023design,shi2023dynamic,xiong2024data}.
\begin{figure}[t]
    \centering    
    \includegraphics[width=8cm,height=3.2cm]{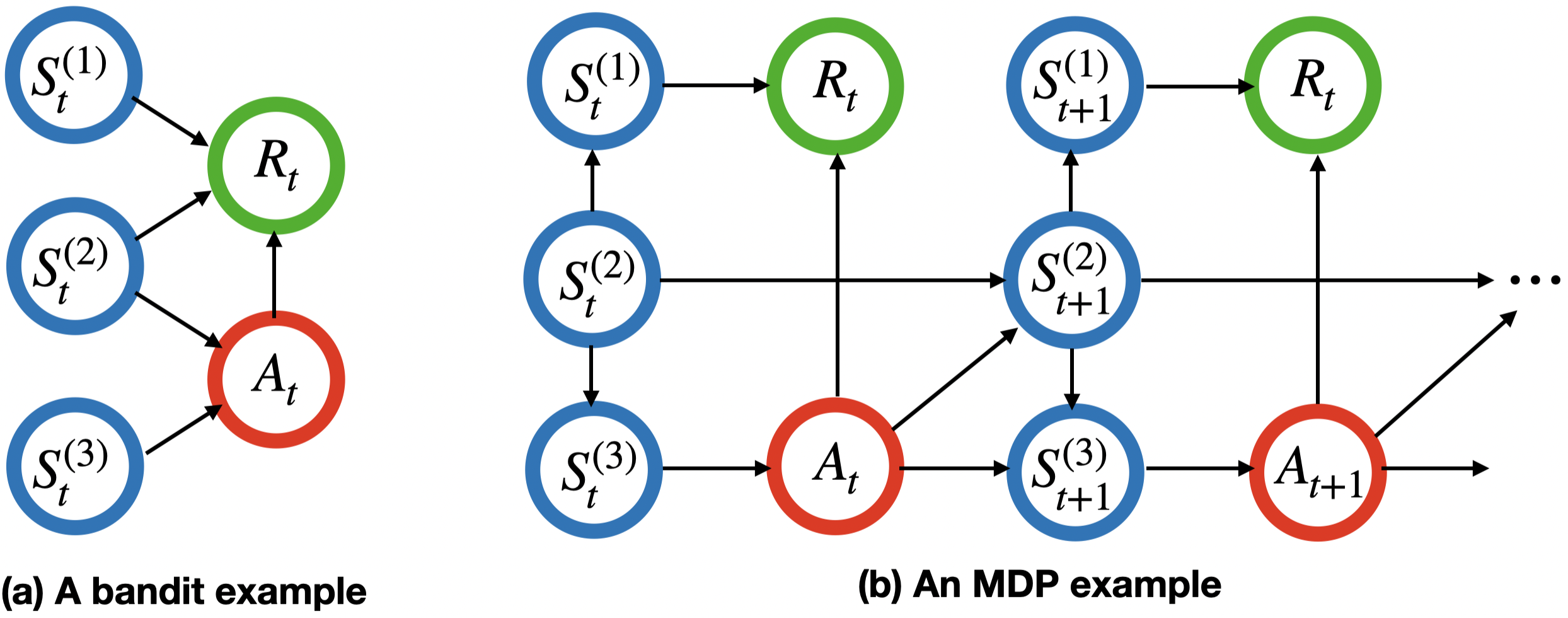}
    \caption{Two illustrative examples.}\label{fig:mdpexample}
    \vspace{-1.5em}
\end{figure}

\textbf{Example I (A bandit example)}: To illustrate the main idea, we start by considering the contextual bandit setting \citep[CB,][]{dudik2014doubly}, which can be regarded as a special MDP with independent state transitions. Under this setting, the states are i.i.d. generated, and model-irrelevance is reduced to reward-irrelevance whereas the proposed backward-model-irrelevance simplifies to behavior-policy-irrelevance. When specialized to variable selection in CB, our proposal is reduced to the iterative confounder selection algorithm in causal inference \citep[see e.g.,][]{guo2022confounder}; see also the review of confounder selection in Appendix \ref{sec:confounder}. We assume the states can be divided into three independent groups, denoted by $S_t^{(1)}$, $S_t^{(2)}$ and $S_t^{(3)}$, respectively. Each group influences the system differently: 
$S_t^{(1)}$ affects only the reward, $S_t^{(2)}$ impacts both the action and the reward, and $S_t^{(3)}$ solely influences the action; see Figure \ref{fig:mdpexample}(a) for an illustration. As formally proven in Lemma \ref{lemmaexample} (see Appendix \ref{subsec:example}):
\begin{itemize}[leftmargin=*]
    \item The type-I MSA selects the first two groups $S_t^{(1)}$ and $S_t^{(2)}$;
    \item The proposed type-II MSA selects the last two groups $S_t^{(2)}$ and $S_t^{(3)}$;
    \item The proposed DSA selects their intersection $S_t^{(2)}$ and converges in two steps, resulting in a smaller subset of variables compared to the two non-iterative procedures.
\end{itemize}

\textbf{Example II (An MDP example)}: We next consider an MDP with three groups of states, %denoted by $\{S_{t,1}\}_t$, $\{S_{t,2}\}_t$ and $\{S_{t,3}\}_t$, 
depicted in Figure \ref{fig:mdpexample}(b). Key observations from this example are as follows: (i) The reward depends on the state only through the first group of variables; (ii) The evolution of the state depends only on the second group. Specifically, the second group evolves first at each time step and subsequently influences the rest two groups; (iii) The behavior policy depends only on the last group; (iv) The second group is directly influenced by the previous action. Based on these observations, we show have that:
\begin{itemize}[leftmargin=*]
    \item According to (i), selecting the first group of variables achieves reward-irrelevance.
    \item According to (iii), selecting the last group of variables achieves behavior-policy-irrelevance.
    \item According to (ii) and (iv), selecting the second group achieves both transition-irrelevance (see the second equation in \ref{eqn:model-irrelevant}) and backward-transition-irrelevance (see \eqref{eqn:backirrelevance-state}). 
\end{itemize}
Consequently, the type-I  and type-II MSAs select the first two and last two groups of variables, respectively. The iterative procedure again selects their intersection $S_t^{(2)}$ and converges in two iterations, leading to in a smaller state space. The reader is referred to Appendix \ref{subsec:example} for formal justifications.

%According to (i), selecting the first group of variables achieves reward-irrelevance. Combined with (ii) and (iii), choosing the first two groups achieves model-irrelevance. Assuming the target policy is agnostic to the state, the proposed type-I MSA will select the first two groups of variables. 

%According to (iv) and that the target policy is state-agnostic, selecting the last two groups attains $\rho^{\pi}$-irrelevance. Meanwhile, according to (ii) and (v), selecting these variables also achieves backward-model-irrelevance. Thus, the proposed type-II MSA will select the last two groups.

%In the iterative procedure, the type-I MSA first eliminates the third group of variables. Given conditions (ii)-(v), selecting just the second group suffices to achieve backward-model-irrelevance, leading to the elimination of the first group in the subsequent type-II MSA. After two iterations, the procedure produces only one group of variables, demonstrating its efficiency in reducing dimensions compared to using either forward or type-II MSA alone.

In both examples, we have demonstrated the advantage of DSA in reducing state space cardinality over non-iterative procedures.

\subsection{Lemma \ref{lemmaexample} and its proof}\label{subsec:example}
We first state Lemma \ref{lemmaexample}. 
\begin{lemma}\label{lemmaexample}
Suppose the reward is a deterministic function of the state-action pair. Then, the followings hold for both the bandit and MDP examples:
\begin{itemize}[leftmargin=*]
    \item The type-I MSA selects the first two groups $S_t^{(1)}$ and $S_t^{(2)}$;
    \item The proposed type-II MSA selects the last two groups $S_t^{(2)}$ and $S_t^{(3)}$;
    \item The proposed DSA selects their intersection $S_t^{(2)}$ and converges in two steps, resulting in a smaller subset of variables compared to the two non-iterative procedures.
\end{itemize}
\end{lemma}
We next prove this lemma. Notice that reward-irrelevance requires the reward function (i.e., the conditional mean of the immediate reward given the state-action pair) to depend on the state only through its abstraction. Under the deterministic reward assumption in Lemma \ref{lemmaexample}, such a conditional mean independence is equivalent to conditional independence. In other words, reward-irrelevance is achieved if the reward is conditionally independent of the state given the action and the abstract state.
\subsubsection{Proof for the bandit example}
We first consider the bandit example. As commented in the main text, in the contextual bandit setting, model-irrelevance is reduced to reward-irrelevance whereas backward-model-irrelevance is reduced to behavior-policy-irrelevance. Consequently, it is immediate to see that the assertions in the first two bullet points hold. 

To prove the last bullet point, notice that according to the first bullet point, DSA would select $S_t^{(1)}$ and $S_t^{(2)}$ in the first iteration. In the second iteration, DSA would select $S_t^{(2)}$, due to the conditional independence between $A_t$ and $S_{t}^{(1)}$ given $S_t^{(2)}$. To verify such conditional independence, notice that there are two paths from $S_t^{(1)}\to A_t$: (i) $S_{t}^{(1)}\to R_t\leftarrow A_t$; (ii) $S_{t}^{(1)}\to R_t\leftarrow S_t^{(2)} \to A_2$. The second path is blocked by $S_t^{(2)}$ whereas the first path contains a collider $R_t$ which is a child of $S_t^{(2)}$. Consequently, both paths fail to $d$-connect $S_t^{(1)}$ and $A_t$ given $S_t^{(2)}$, leading to the desired conditional independence property. Since $R_t$ is a child of $S_t^{(2)}$, in the third iteration, $S_t^{(2)}$ will be selected as well. Similarly, in the subsequent iteration, $S_t^{(2)}$ will again be selected since $A_t$ is a child of $S_t^{(2)}$. Consequently, DSA converges after two iterations. 
\subsection{Proof for the MDP example}
As discussed in the main text:
\begin{itemize}[leftmargin=*]
    \item Selecting the first group of variables achieves reward-irrelevance.
    \item Selecting the last group of variables achieves behavior-policy-irrelevance.
    \item Selecting the second group of variables achieves both transition-irrelevance and backward-transition-irrelevance. 
\end{itemize}
It is immediate to see that the the assertions in the first two bullet points hold. To prove the last bullet point, again, notice that DSA would select $S_t^{(1)}$ and $S_t^{(2)}$ in the first iteration. In the second iteration, DSA would select $S_t^{(2)}$, due to (i) the conditional independence between $S_t^{(1)}$ and $A_t$ given $S_t^{(2)}$ and (ii) that between $S_{t+1}^{(1)}$ and ($A_t$, $S_t^{(2)}$) given $S_{t+2}^{(1)}$. This is because (i) implies behavior-policy-irrelevance and (ii) implies backward-transition-irrelevance (see the discussion below \eqref{eqn:backirrelevance-state}) when restricted to the space of the first two groups of variables.  

It remains to verify (i) and (ii). To prove (i), notice that all paths from $S_t^{(1)}$ to $A_t$ is either blocked by $S_t^{(2)}$, or include the collider $S_t^{(1)}\to R_t\leftarrow A_t$. To prove (ii), similarly, notice that all paths from ($A_t$, $S_t^{(2)}$) to $S_{t+1}^{(1)}$ is either blocked by $S_t^{(2)}$, or include the collider $S_{t+1}^{(1)}\to R_{t+1}\leftarrow A_{t+1}$. 

Thus, we have shown that DSA would select $S_t^{(2)}$ in the second iteration. In the third iteration, notice that there is a path $S_t^{(2)}\to S_t^{(1)}\to R_t$ which is not blocked by $A_t$. Consequently, DSA would select $S_t^{(2)}$ in the third iteration as well. Similarly, in the subsequently iteration, DSA would select $S_t^{(3)}$, due to the path $S_t^{(2)}\to S_t^{(3)}\to A_t$. 
As such, it converges after two iterations.

\section{Additional Properties of the Iterative Procedure}\label{sec:add-properties}
In this section, we discuss additional properties of the iterative procedure, including: (i) defining a set of irrelevance conditions for learning state abstractions in OPE; and (ii) proving the Fisher consistency of various OPE estimators when applied to our proposed abstract state spaces.

Let $\mathcal{M}=\langle \mathcal{S},\mathcal{A},\mathcal{T},\mathcal{R},\rho_0,\gamma\rangle$ be the ground MDP. A state abstraction $\phi$ is a mapping from state space $\mathcal{S}$ to a certain abstract state space $\mathcal{X}=\{\phi(s): s\in \mathcal{S}\}$. 
In practical terms, the transformation of raw MDP data into a new sequence of state-action-reward tuples $(\phi(S), A, R,\phi(S'))$. 
Denote $Q_{\phi}^{\pi}$ as the $Q$-function when restricted to the abstract state space, i.e., 
$$Q_{\phi}^{\pi}(a,\phi(s))=\sum_{t\ge 1}\gamma^{t-1}\mathbb{E}^{\pi}[R_t|A_1=a,\phi(S_1)=\phi(s)],$$  $\rho^{\pi}_{\phi}$ as the IS ratio defined on the abstract state space, i.e, $$\rho_{\phi}^{\pi}(a,x)=\sum_{s'\in \phi^{-1}(x)} f(s'|x) \rho^{\pi}(a,s')$$ with  $f(\bullet|x)$ being a conditional probability mass function (pmf) such that $\sum_{s'\in \phi^{-1}(x)} f(s'|x)=1$. We similarly define
\begin{eqnarray*}
    w_{\phi}^{\pi}(a,x)=(1-\gamma)\frac{\sum_{t\geq 1}\gamma^{t-1}\mathbb{P}^{\pi}(A_t=a,\phi(S_t)=x)}{\lim_{t\rightarrow \infty} \mathbb{P}(A_t=a,\phi(S_t)=x)}. 
\end{eqnarray*}
The identification formulas presented in Section \ref{sec:OPEmethod} remain valid when replacing the oracle Q-function or (M)IS ratio with those projected into the proposed abstract state space. For the identifibality of them, 
we first introduce several model-free irrelevance conditions 
tailored for OPE.
\begin{comment}

\begin{Def}[$\pi$-irrelevance]\label{def:pi}
    $\phi$ is $\pi$-irrelevant if for any $s^{(1)}, s^{(2)} \in \mathcal{S}$ whenever $\phi(s^{(1)})=\phi(s^{(2)})$, we have 
$\pi(a|s^{(1)})=\pi(a|s^{(2)})$ for any $a\in \mathcal{A}$.
\end{Def}
\end{comment}
\begin{Def}[$Q^{\pi}$-irrelevance]\label{def:Qpi}
    $\phi$ is $Q^{\pi}$-irrelevant if for any $s^{(1)}, s^{(2)} \in \mathcal{S}$  whenever $\phi(s^{(1)})=\phi(s^{(2)})$, we have $Q^{\pi}(a,s^{(1)})=Q^{\pi}(a,s^{(2)})$ for any $a\in \mathcal{A}$. 
\end{Def}
\begin{Def}[$\rho^{\pi}$-irrelevance]\label{def:rhopi}
    $\phi$ is $\rho^{\pi}$-irrelevant if for any $s^{(1)}, s^{(2)} \in \mathcal{S}$ whenever $\phi(s^{(1)})=\phi(s^{(2)})$, we have $\rho^{\pi}(a,s^{(1)})=\rho^{\pi}(a,s^{(2)})$ for any $a\in \mathcal{A}$.
\end{Def}
\begin{Def}[$w^{\pi}$-irrelevance]\label{def:wpi}
    $\phi$ is $w^{\pi}$-irrelevant if for any $s^{(1)}, s^{(2)} \in \mathcal{S}$  whenever $\phi(s^{(1)})=\phi(s^{(2)})$, we have $w^{\pi}(a,s^{(1)})=w^{\pi}(a,s^{(2)})$ for any $a\in \mathcal{A}$.
\end{Def}
 These definitions are specifically tailored for the Q-function-based method and IS estimators (see Section~\ref{sec:OPEmethod}). These irrelevance conditions encourage us to conduct OPE on the abstract state space to reduce sample complexity. 
 Nevertheless, methods for deriving abstractions that satisfy these conditions (particularly $Q^{\pi}$- and $w^{\pi}$-irrelevance) remain unclear. Furthermore, the state-action-reward triplets transformed via these abstractions $(\phi(S), A, R,\phi(S'))$ might not maintain the MDP structure. This  complicates the process of learning $Q_{\phi}^{\pi}$ and $w_{\phi}^{\pi}$. These challenges %underscore the need for developing model-based abstractions, which we explore 
 motivate us to consider model-based irrelevance conditions introduced in the  Section~\ref{sec:iterative}. 
\subsection{Lemma \ref{thm1} and its proof}\label{prooflemmathm1}

\textbf{An auxiliary lemma}. To begin with, we introduce the following lemma which validates the unbiasedness of various OPE estimators under the model-free irrelevance conditions in Definitions \ref{def:Qpi} -- \ref{def:wpi}, whose proof is given in Section \ref{prooflemmathm1}. 

\begin{lemma}[Unbiasedness under model-free irrelevance conditions]\label{thm1}
    Under $Q^{\pi}$-, $\rho^{\pi}$- or $w^{\pi}$-irrelevance, the corresponding methods are unbiased when applied to the abstract state space, assuming the oracle Q-function or (M)IS ratio is identifiable from the data:  
    \begin{itemize}[leftmargin=*]
        \item Under $Q^{\pi}$-irrelevance,  
        Q-function-based method (Method 1) remains unbiased, i.e., the Q-function $Q_{\phi}^{\pi}$ defined on the abstract space satisfies $\mathbb{E} [f_1(Q^{\pi})]=\mathbb{E} [f_1(Q^{\pi}_{\phi})]$; 
        \item Under $\rho^{\pi}$-irrelevance, SIS (Method 2) remains unbiased, i.e., the IS ratio $\rho_{\phi}^{\pi}$ defined on the abstract state space satisfies $\mathbb{E} [f_2(\rho^{\pi})]=\mathbb{E} [f_2(\rho^{\pi}_{\phi})]$; 
        \item Under $w^{\pi}$-irrelevance, MIS (Method 3) remains unbiased, i.e., the MIS ratio $w_{\phi}^{\pi}$ defined on the abstract state space  satisfies $\mathbb{E} [f_3(w^{\pi})]=\mathbb{E} [f_3(w^{\pi}_{\phi})]$. 
    \end{itemize}
    Moreover, when $\phi$ satisfies either $Q^{\pi}$-irrelevance or $w^{\pi}$-irrelevance, DRL (Method 4) remains unbiased, i.e., $Q_{\phi}^{\pi}$ and $w_{\phi}^{\pi}$ defined on the abstract state space satisfy $\mathbb{E}[f_4(Q^{\pi},w^{\pi})]=\mathbb{E} [f_4(Q^{\pi}_{\phi},w^{\pi}_{\phi})]$. 
\end{lemma}
Lemma \ref{thm1} proves the unbiasedness of the four OPE methods presented in Section \ref{sec:OPEmethod} when applied to the abstract state space, under the corresponding irrelevance conditions. Notably, DRL requires weaker irrelevance conditions compared to the Q-function-based method and MIS, owing to its inherent double robustness property.

We prove Lemma \ref{thm1} in this subsection.
We first prove that under $Q^{\pi}$-, $\rho^{\pi}$- or $w^{\pi}$-irrelevance, the corresponding methods remain unbiased when applied to the abstract state space:  
\begin{itemize}[leftmargin=*]
\item \textbf{Unbiasedness under $Q^{\pi}$-irrelevance}.  By definition, $Q^{\pi}$ is the expected return given an initial state $S_1$ and $A_1$. Under $Q^{\pi}$-irrelevance, the Q-function depends on $S_1$ only through $\phi(S_1)$. It follows that $Q^{\pi}$ equals the expected return given $\phi(S_1)$ and $A_1$, the latter being $Q_{\phi}^{\pi}$ -- the $Q$-function when restricted to the abstract state space, i.e., $Q_{\phi}^{\pi}(a,\phi(s))=\sum_{t\ge 1}\gamma^{t-1}\mathbb{E}^{\pi}[R_t|A_1=a,\phi(S_1)=\phi(s)]$. It follows that
%It follows from the definition of $Q^{\pi}$-irrelevance that 
\begin{align*}
\mathbb{E} [f_1(Q^{\pi})]=&\sum_{a,s}\pi(a|s)Q^{\pi}(a,s)\mathbb{P}(S_1=s)\\
=&\sum_{a,s}\pi(a|s)Q_{\phi}^{\pi}(a,\phi(s))\mathbb{P}(S_1=s)\\
%=&\sum_{a,x}Q_{\phi}^{\pi}(a,x)\sum_{s\in \phi^{-1}(x)}\pi(a|s)\mathbb{P}(S_1=s)\\
%=&\sum_{a,x}Q_{\phi}^{\pi}(a,x)\sum_{s\in \phi^{-1}(x)}\pi(a|s)\frac{\mathbb{P}(S_1=s)}{\mathbb{P}(\phi(S_1)=x)}\mathbb{P}(\phi(S_1)=x)\\
%=&\sum_{a,x}Q_{\phi}^{\pi}(a,x)\pi(a|x)\mathbb{P}(\phi(S_1)=x)\\
=&
\mathbb{E} [f_1(Q^{\pi}_{\phi})].
\end{align*}
 \item \textbf{Unbiasedness under $\rho^{\pi}$-irrelevance}. We first establish the equivalence between $\rho^{\pi}$ and $\rho^{\pi}_{\phi}$ -- the IS ratio defined on the abstract state space. Under $\rho^{\pi}$-irrelevance, $\rho^{\pi}(a,s)$ becomes a constant function of $x=\phi(s)$. Consequently, for any conditional probability mass function (pmf) $f(\bullet|x)$ such that $\sum_{s'\in \phi^{-1}(x)} f(s'|x)=1$, we have $\rho^{\pi}(a,s)=\sum_{s'\in \phi^{-1}(x)} f(s'|x) \rho^{\pi}(a,s')$. %{\color{red} Your original one is: for any conditional probability mass function (pmf) \hao{$f(s|x)$} such that \hao{$\sum_{s\in \phi^{-1}(x)} f(s|x)=1$}, we have \hao{$\rho^{\pi}(a,s)=\sum_{s\in \phi^{-1}(x)} f(s|x) \rho^{\pi}(a,s)$.} }
 By setting $f(\bullet|x)$ to the pmf of $S_t$ given $A_t=a$ and $\phi(S)=x$, it follows that
 \begin{eqnarray}\label{eqn:rhoirrelevance1}
     \rho^{\pi}(a,s)=\sum_{s'\in \phi^{-1}(x)} \mathbb{P}(S_t=s'|A_t=a,\phi(S_t)=x) \rho^{\pi}(a,s').
 \end{eqnarray}
 Notice that
 \begin{eqnarray*}
     \mathbb{P}(S_t=s'|A_t=a,\phi(S_t)=x)=\frac{\mathbb{P}(A_t=a,S_t=s'|\phi(S_t)=x)}{\mathbb{P}(A_t=a|\phi(S_t)=x)}.
 \end{eqnarray*}
 The denominator equals $b_{\phi,t}(a|x)$, the behavior policy when restricted to the abstract state space at time $t$. Notice that this behavior policy can be non-stationary over time, despite that $b$ being time-invariant. 
 As for the numerator, it is straightforward to show that it equals $b(a|s')\mathbb{P}(S_t=s'|\phi(S_t)=x)$. This together with \eqref{eqn:rhoirrelevance1} yields
 \begin{eqnarray}\label{eqn:rhoirrelevance2}
     \rho^{\pi}(a,s)=\sum_{s' \in \phi^{-1}(x)} \frac{\pi(a|s')}{b_{\phi,t}(a|x)} \mathbb{P}(S_t=s'|\phi(S_t)=x)=\frac{\pi_{\phi,t}(a|x)}{b_{\phi,t}(a|x)},%=\rho_{\phi,t}^{\pi}(a|x),
 \end{eqnarray}
 where $\pi_{\phi,t}$ %and $\rho_{\phi,t}^{\pi}$ 
 denotes the target policy %and the IS ratio 
 confined on the abstract state space at time $t$, namely,  $\pi_{\phi,t}(a|x)=\sum_{s' \in \phi^{-1}(x)}\pi(a|s')\mathbb{P}(S_t=s'|\phi(S_t)=x)$.
 The last term in \eqref{eqn:rhoirrelevance2} is given by $\rho_{\phi,t}^{\pi}$. 
 Consequently, the sequential IS ratio $\rho_{1:t}^{\pi}$ is equal to $\prod_{k=1}^t \rho_{\phi,k}^{\pi}(A_k,\phi(S_k))$. This in turn yields  $\mathbb{E} [f_2(\rho^{\pi})]=\mathbb{E} [f_2(\rho^{\pi}_{\phi})]$. 

\item \textbf{Unbiasedness under $w^{\pi}$-irrelevance}. Similar to the proof under $\rho^{\pi}$-irrelevance, the key lies in establishing the equivalence between $w^{\pi}(a,s)$ and $w^{\pi}_{\phi}(a,\phi(s))$, the latter being the MIS ratio defined on the abstract state space. Once this has been proven, it is immediate to see that $\mathbb{E}[f_3(w^{\pi})]=\mathbb{E} [f_3(w^{\pi}_{\phi})]$, so that the MIS remains Fisher consistent when applied to the abstract state space. 

As discussed in Section \ref{sec:OPEmethod}, to guarantee the unbiasedness of the MIS estimator, we additionally require a stationarity assumption. Under this requirement, for a given state-action pair $(S,A)$ in the offline data, its joint pmf function can be represented as $p_{\infty} \times b$ where $p_{\infty}$ denotes the marginal state distribution under the behavior policy. Additionally, let $p^{\pi}_t$ denote the pmf of $S_t$ generated under the target policy $\pi$. The MIS ratio can be represented by 
\begin{eqnarray*}
    w^{\pi}(a,s)=\frac{(1-\gamma)\sum_{t\ge 1}\gamma^{t-1}p_t^{\pi}(s)\pi(a|s)}{p_{\infty}(s)b(a|s)}.
\end{eqnarray*}
Similar to \eqref{eqn:rhoirrelevance2}, under $w^{\pi}$-irreleavance, it follows that
\begin{eqnarray*}
    w^{\pi}(a,s)&=&(1-\gamma)\sum_{s'\in \phi^{-1}(x)} \frac{\sum_{t\ge 1}\gamma^{t-1}p_t^{\pi}(s')\pi(a|s')}{p_{\infty}(s')b_{\phi}(a|x)}\mathbb{P}(S=s'|\phi(S)=x)\\
    &=& \frac{(1-\gamma)\sum_{s'\in \phi^{-1}(x)} \sum_{t\ge 1}\gamma^{t-1}p_t^{\pi}(s')\pi(a|s')}{p_{\infty}(x)b_{\phi}(a|x)}.
\end{eqnarray*}
Here, the subscript $t$ in $b_{\phi}$ and $S$ is dropped due to stationarity. Additionally, $p_{\infty}(x)$ is used to denote the probability mass function (pmf) of $\phi(S)$, albeit with a slight abuse of notation. Moreover, the numerator represents the discounted visitation probability of $(A,\phi(S))$ under $\pi$. This proves that $w^{\pi}(a,s)=w^{\pi}_{\phi}(a,\phi(s))$. 
\end{itemize}
Finally, we establish the unbiasedness of DRL. According to the doubly robustness property, DRL is Fisher consistent when either $Q^{\pi}$ or $w^{\pi}$ is correctly specified. Under $Q^{\pi}$-irrelevance, we have $Q^{\pi}(a,s)=Q^{\pi}_{\phi}(a,\phi(s))$ and thus DRL remains Fisher consistent when applied to the abstract state space. Similarly, we have $w^{\pi}(a,s)=w^{\pi}_{\phi}(a,\phi(s))$ under $w^{\pi}$-irrelevance, which in turn implies DRL's unbiasedness. This completes the proof.

\subsection{Lemma \ref{thm2} and its proof}\label{proofoflemmathm2}

\textbf{An auxiliary lemma}. The following lemma summarizes the the relationships beteewn defnition \ref{def:model} and \ref{def:Qpi} -- \ref{def:wpi}. In particular, we first discuss the connections between type-I MSA   given in Definition \ref{def:model} and the notions of $Q^{\pi}$-, $\rho^{\pi}$- and $w^{\pi}$-irrelevance, 
and establish Fisher consistency. 
Results in the first two bullet points are based on those in the existing literature \citep[see e.g.,][]{li2006towards,pavse2023scaling}.The proof is given in Section \ref{proofoflemmathm2}. 

\begin{lemma}[OPE under Type-I MSA]\label{thm2}
    Assume Assumptions \ref{asmp:bounded}--\ref{asmp:stationary} hold. Let $\phi$ denote a $\pi$-irrelevant type-I MSA.  Then:%that satisfies \eqref{eqn:model-irrelevant}.   
    \begin{itemize}[leftmargin=*]
        \item \textbf{$Q^{\pi}$-irrelevance \& Fisher consistency of Q-function-based method}: $\phi$ is also $Q^{\pi}$-irrelevant, and the corresponding Q-function-based method (Method 1) is thus Fisher consistent, i.e., the Q-function $Q_{\phi}^{\pi}$ defined on the abstract space is identifiable and satisfies $\mathbb{E} [f_1(Q^{\pi})]=\mathbb{E} [f_1(Q^{\pi}_{\phi})]$; 
        
        \item \textbf{Fisher consistency of MIS}: While $\phi$ is not necessarily $w^{\pi}$-irrelevant, MIS (Method 3) is  Fisher consistent when applied to the abstract state space, i.e., the MIS ratio $w_{\phi}^{\pi}$ defined on the abstract state space is identifiable and satisfies $\mathbb{E} [f_3(w^{\pi})]=\mathbb{E} [f_3(w^{\pi}_{\phi})]$;
        
        \item \textbf{Fisher consistency of SIS}: While $\phi$ is not necessarily $\rho^{\pi}$-irrelevant, SIS (Method 2) with a history-dependent IS ratio (detailed in the proof of Lemma \ref{thm2} in Appendix \ref{prooflemmathm1}) is  Fisher consistent when applied to the abstract space; %Indeed, the Fisher consistency only requires reward-irrelevance (see the first part of \eqref{eqn:model-irrelevant}).  
        %is Fisher consistent when applied to the abstract state space if $\phi$ is additionally $\pi$-irrelevant. %Indeed, the validity only requires reward-irrelevance. 

        \item \textbf{Fisher consistency of DRL}: DRL (Method 4) is  Fisher consistent when applied to the abstract state space. %Indeed, the validity only requires reward-irrelevance. 
        %the sequential importance sampling estimator \eqref{eqn:sisest} constructed on the abstract state space remains unbiased.
%        {\item  if $\phi$ is inverse model-irrelevant, it is not necessarily a $Q^{\pi}$-irrelevant. However, the sequential importance sampling {\color{blue}(value-based?)} estimator \eqref{eq3} constructed on the abstract state space remains unbiased.}
    \end{itemize}
\end{lemma}
%The first bullet point is grounded in Theorem 2 of \cite{li2006towards}. 
The first bullet point establishes the link between model-irrelevance and $Q^{\pi}$-irrelevance and proves the Fisher consistency of the Q-function-based method when applied to the abstract state space. %This finding has previously been established in Theorem 2 of \citet{li2006towards}. 
To satisfy $Q^{\pi}$-irrelevance, we need both model-irrelevance and $\pi$-irrelevance. 
%The additional requirement of $\pi$-irrelevance 
%The latter is readily attainable: %in the first bullet point 
%is mild. 
%For instance, it is automatically satisfied when $\pi$ is a non-dynamic policy, consistently assigning the same action at every time point. Such a policy is frequently employed in A/B testing to evaluate the global treatment effect of a particular action \citep{tang2022reinforcement,bojinov2023design,shi2023dynamic}. Additionally, when the action space is limited, 
%given a model-irrelevant abstraction $\phi(s)$, we propose to %define $\phi'(s)$ as the union of 
 %\z{Do we mean '$Q^{\pi}$-irrelevant', or has the sentence been incidentally chopped (and continues)?}
In our implementation, %we use deep neural networks to parameterize the abstraction. 
we first adapt existing algorithms %\citep[see e.g.,][]{gelada2019deepmdp} 
to train a model-irrelevant abstraction $\phi$, parameterized via deep neural networks. We next combine $\phi(s)$ with $\{\pi(a|s): a\in \mathcal{A}\}$ to obtain a new abstraction $\phi_{for}(s)$. This augmentation ensures $\phi_{for}(s)$ is $\pi$-irrelevant, and hence $Q^{\pi}$-irrelevant. Refer to Appendix \ref{sec:implementforward} for the detailed procedures. 
%In fact, the incorporating  of the information from  target policy $\pi$ distinguishes our proposal from the abstractions in policy learning. 

The last three bullet points validate the SIS, MIS and DRL estimators, despite $\phi$ being neither $w^{\pi}$-irrelevant nor $\rho^{\pi}$-irrelevant. By definition, $\rho^{\pi}$-irrelevance can be achieved by selecting state features that adequately predict the IS ratio. However, methods for constructing $w^{\pi}$-irrelevant abstractions remain less clear. In the following, we prove that the backward MDP model-based irrelevance condition that ensures $w^{\pi}$-irrelevance.  
We prove Lemma \ref{thm2} in this subsection. We first show $Q^{\pi}$-irrelevance under model-irrelevance $\&$ $\pi$-irrelevance, and prove the Fisher consistency of the Q-function-based method. Then, we establish Fisher consistency of MIS. Next, we derive Fisher consistency of SIS. Finally, we prove the Fisher consistency of DRL.
\begin{itemize}[leftmargin=*]
\item \textbf{Fisher consistency of Q-function-based method}. 
%It suffices to show that, for any $s^{(1)}$ and $s^{(2)}$ satisfying \eqref{eqn:model-irrelevant} and , 
We first use the induction method to prove that 
\begin{align}\label{eqn:Qirrelevance}
Q^{\pi}(a,s^{(1)})=Q^{\pi}(a,s^{(2)}),
\end{align}
whenever $s^{(1)}$ and $s^{(2)}$ satisfy $\phi(s^{(1)})=\phi(s^{(2)})$. This demonstrates $Q^{\pi}$-irrelevance, which further implies $\mathbb{E} [f_1(Q^{\pi})]=\mathbb{E} [f_1(Q^{\pi}_{\phi})]$ according to Lemma \ref{thm1}. We next establish the identifiability of $Q^{\pi}_{\phi}$. 

%To prove this result, we can 
%Toward that end, we use the induction method.
Define 
\begin{align*}
&Q_j^{\pi}(a,s)=\mathbb{E}^{\pi}\left[\sum_{t=1}^j\gamma^{t-1}R_t|S_1=s,A_1=a\right],\,\,\hbox{and}\,\,\\
&V_j^{\pi}(s)=\mathbb{E}^{\pi}\left[\sum_{t=1}^j\gamma^{t-1}R_t|S_1=s\right].
\end{align*}
%It follows from the definition of $Q_j^{\pi}(s,a)$, $s^{(1)}$ and $s^{(2)}$, 
Under reward-irrelevance, we have
\begin{align*}
Q_1^{\pi}(a,s^{(1)})
=&\mathbb{E}^{\pi}\left[R_1|S_1=s^{(1)},A_1=a\right]\\
=&\mathcal{R}(a,s^{(1)})\\
=&\mathcal{R}(a,s^{(2)})\\
=&Q_1^{\pi}(a,s^{(2)}).
\end{align*}
Together with $\pi$-irrelevance, we obtain that
\begin{align*}
V_1^{\pi}(s^{(1)})
=&\sum_{a\in \mathcal{A}}Q_1^{\pi}(a,s^{(1)})\pi(a|s^{(1)})\\
=&\sum_{a\in \mathcal{A}}Q_1^{\pi}(a,s^{(2)})\pi(a|s^{(2)})\\
%=&\sum_{a\in \mathcal{A}}\mathcal{R}(a,s^{(2)})\pi(a|s^{(2)})\\
=&V_1^{\pi}(s^{(2)}).
\end{align*}
%Then, if we assume that for $j\leq T-1$, the following equations hold,
Suppose we have shown that the following holds for any $j<T$,
\begin{align}\label{eq2}
Q_j^{\pi}(a,s^{(1)})=Q_j^{\pi}(a,s^{(2)})\,\,\mathrm{and}\,\,
V_j^{\pi}(s^{(1)})=V_j^{\pi}(s^{(2)}),
\end{align}
whenever $\phi(s^{(1)})=\phi(s^{(2)})$. 
Our goal is to show \eqref{eq2} holds with $j=T$. 

We similarly define $Q_{j,\phi}^{\pi}$ and $V_{j,\phi}^{\pi}$ as the Q- and value functions on the abstract state space. Similar to the proof of Theorem \ref{thm1}, we can show that $Q_j^{\pi}=Q_{j,\phi}^{\pi}$ and $V_j^{\pi}=V_{j,\phi}^{\pi}$ for any $j<T$.

Direct calculations yield 
\begin{align*}
Q^{\pi}_T(a,s^{(1)})=&\mathbb{E}^{\pi}\left[\sum_{t=1}^{T}\gamma^{t-1}R_t|S_1=s^{(1)},A_1=a\right]\\
=&\mathbb{E}^{\pi}\left[\sum_{t=2}^{T}\gamma^{t-1}R_t|S_1=s^{(1)},A_1=a\right]+\mathcal{R}(a,s^{(1)})\\
=&\mathbb{E}^{\pi}\sum_{s'\in\mathcal{S}}\left[\sum_{t=2}^{T}\gamma^{t-1}R_t|S_2=s'\right]\mathcal{T}(s'|s^{(1)},a)+\mathcal{R}(a,s^{(1)})\\
=&\gamma\mathbb{E}^{\pi}\sum_{x'\in\mathcal{X}}\sum_{s'\in \phi^{-1}(x')}\left[\sum_{t=2}^{T}\gamma^{t-2}R_t|S_2=s'\right]\mathcal{T}(s'|s^{(1)},a)+\mathcal{R}(a,s^{(1)})\\
%=&\gamma\sum_{x'\in\mathcal{X}}\sum_{s'\in \phi^{-1}(x')}\underbrace{\mathbb{E}^{\pi}\left[\sum_{t=1}^{T-1}\gamma^{t-1}R_t|S_1=s'\right]\mathcal{T}(s'|s^{(1)},a)}_{\mathrm{Markov  property}}+\mathcal{R}_1(a,s^{(1)})\\
=&\gamma\sum_{x'\in\mathcal{X}}
\sum_{s'\in \phi^{-1}(x')}V_{T-1}^{\pi}(s')\mathcal{T}(s'|s^{(1)},a)+\mathcal{R}(a,s^{(1)})\\
%=&\gamma\sum_{x'\in\mathcal{X}}\underbrace{V_{T-1,\phi}^{\pi}(x')}_{\mathrm{by}\,\, \eqref{eq2}}\sum_{s'\in \phi^{-1}(x')}\mathcal{T}(s'|s^{(1)},a)+\mathcal{R}(a,s^{(1)})\\
=&\gamma\sum_{x'\in\mathcal{X}}\sum_{s'\in \phi^{-1}(x')}V_{T-1}^{\pi}(s')\mathcal{T}(s'|s^{(2)},a)+\mathcal{R}(a,s^{(2)})\\
=&Q^{\pi}_T(a,s^{(2)}),
\end{align*}   
where the second last equation holds due to transition-irrelevance in \eqref{eqn:model-irrelevant} and \eqref{eq2}, which states that $V_{T-1}^{\pi}(s')$ is constant as a function of $s'\in \phi^{-1}(x')$. 

This together with $\pi$-irrelevance proves $V_{T}^{\pi}$-irrelevance. By induction, we have shown that \eqref{eq2} holds for any $j\ge 1$. Under the boundness assumption in Assumption \ref{asmp:bounded}, 
$Q_j^{\pi}\to Q^{\pi}$ as $j\to \infty$. We thus obtain \eqref{eqn:Qirrelevance}, which yields $Q^{\pi}$-irrelevance. 

Next, we prove the identifiability of $Q^{\pi}_{\phi}$. Similarly, we define
\begin{align}\label{eqn:Qjphi}
Q_{j,\phi}^{\pi}(a,x)=\sum_{t=1}^j \gamma^{t-1} \mathbb{E}^{\pi}\left[R_t|\phi(S_1)=x,A_1=a\right]. 
\end{align}
By setting $j=1$, it reduces to $\mathbb{E}^{\pi}[R_1|\phi(S_1)=x,A_1=a]$. Under the MDP model assumption, the conditional mean of the immediate reward depends solely on the current state-action pair, independent of past history. This together with the reward-irrelevance condition further implies that the conditional mean of the reward depends solely on the abstract-state-action pair. Consequently, 
\begin{eqnarray*}
    \mathbb{E}^{\pi}[R_1|\phi(S_1)=x,A_1=a]=\underbrace{\mathbb{E}[R_1|\phi(S_1)=x,A_1=a]}_{\mathcal{R}_{\phi}(a,x)}.
\end{eqnarray*}
Notice that the expectation on the right-hand-side (RHS) is defined with respect to the behavior policy. It can thus be consistently estimated using the offline data under the coverage assumption in Assumption \ref{asmp:coverage}. This yields the identifiability of $Q_{1,\phi}^{\pi}$. 

Similarly, we can show that $$\mathbb{P}^{\pi}[\phi(S_2)=x'|A_1=a,\phi(S_1)=x]=\underbrace{\mathbb{P}[\phi(S_2)=x'|A_1=a,\phi(S_1)=x]}_{\mathcal{T}_{\phi}(x'|a,x)},$$ under transition-irrelevance, which establishes the identifiability of the left-hand-side (LHS). 

Notice that for any $j\ge 1$, under the MDP model assumption,  $Q_{j,\phi}^{\pi}$ can be represented using $\mathbb{E}^{\pi}[R_1|\phi(S_1)=x,A_1=a]$ and $\mathbb{P}^{\pi}[\phi(S_2)=x'|A_1=a,\phi(S_1)=x]$. Both have been proven identifiable. This the establishes identifiability of $Q_{j,\phi}^{\pi}$. Again, by letting $j\to \infty$, we obtain the identifiability of $Q_{\phi}^{\pi}$ under the boundedness assumption in Assumption \ref{asmp:bounded}. The proof is hence completed.

%The proof is hence completed. 
  %Notice that we only require reward-irrelevance in the above proof. 

\item \textbf{Fisher consistency of MIS}. We use $p_{t,\phi}^{\pi}(a,x)$ to denote the probability $\mathbb{P}^{\pi}(A_t=a,\phi(S_t)=x)$ and $p_t^{\pi}(s)$ to denote $\mathbb{P}^{\pi}(S_t=s)$. 
Under the stationary assumption, direct calculations yield 
\begin{align*}
\mathbb{E} [f_3(w^{\pi}_{\phi})]=
& \mathbb{E}\left[ (1-\gamma)^{-1} w_{\phi}^{\pi}(A,\phi(S))R\right]\\
         =& \mathbb{E}\left[(1-\gamma)^{-1} w_{\phi}^{\pi}(A,\phi(S))\mathcal{R}\big(A,S\big)\right]\\
            =& \mathbb{E}\left[ (1-\gamma)^{-1}w_{\phi}^{\pi}(A,\phi(S))\underbrace{\mathcal{R}_{\phi}\big(A,\phi(S)\big)}_{\mathrm{reward-irrelevant}}\right]\\
    =&\sum_{a\in \mathcal{A},x\in \mathcal{X}}\sum_{t=1}^{+\infty}\gamma^{t-1}p_t^{\pi}(a,x)\mathcal{R}_{\phi}(a,x)\\
      =&\sum_{a\in \mathcal{A},x\in \mathcal{X}}\sum_{s\in \phi^{-1}(x)}\sum_{t=1}^{+\infty}\gamma^{t-1}\pi(a|s)p_t^{\pi}(s)\mathcal{R}(a,s)\\
    =&\sum_{t=1}^{+\infty} \gamma^{t-1}\mathbb{E}^{\pi}({R}_t)\\
    =&\mathbb{E} [f_3(w^{\pi})].
\end{align*}
To complete the proof, it remains to establish the identifiability of $w_{\phi}^{\pi}$. 

Under the stationarity assumption in Assumption \ref{asmp:stationary}, $\omega_{\phi}^{\pi}$ can be represented by
\begin{eqnarray*}
    \frac{\sum_{t\ge 1}\gamma^{t-1} p_{t,\phi}^{\pi}(a,x)}{\mathbb{P}(A_1=a,\phi(S_1)=x)}.
\end{eqnarray*}
It is immediate to see that the denominator is identifiable, as the probability is calculated with respect to the behavior policy. It suffices to show that for any $t\ge 1$, $p_{t}^{\pi}$ is identifiable as well. Under transition-irrelevance, the data triplets $(\phi(S),A,R)$ forms an MDP, satisfying the Markov assumption. As such, we can rewrite $p_{t}^{\pi}(a_t,x_t)$ as
\begin{eqnarray*}
    \sum_{\substack{a_1,\cdots,a_{t-1}\in \mathcal{A}\\ x_1,\cdots,x_{t-1}\in \mathcal{X}}}\rho_{0,\phi}(x_1)\prod_{k=1}^{t-1} \Big[\pi_{\phi}(a_k|x_k) \mathcal{T}_{\phi}(x_{k+1}|a_k,x_k)\Big]\pi_{\phi}(a_t|x_t),
\end{eqnarray*}
where $\rho_{0,\phi}$ denotes the pmf of $\phi(S_1)$ which is independent of $\pi$, and both $\pi_{\phi}$ and $\mathcal{T}_{\phi}$ are identifiable under $\pi$- and transition-irrelevance, respectively. This proves the identifiability of $p_t^{\pi}$, and hence, the identifiability of $w_{\phi}^{\pi}$. 
%Notice that $\mathcal{T}_{\phi}$ is independent of the target policy $\pi$.

\item \textbf{Fisher consistency of SIS}. Recall that we require the behavior policy to be Markovian, meaning that at any time $t$, $A_t$ is independent of historical observations given $S_t$. A key challenge in state abstraction for the SIS estimator is that, after abstraction, the behavior policy  can be history-dependent, leading to the inconsistency of SIS. Toward that end, we employ a history-dependent IS ratio to address this challenge. Specifically, let $\rho_{j,\phi}^{\pi}$ denote
%To account for the abstract space obtained after the forward state abstraction, we employ a history-dependent IS ratio  rather than the standard IS ratio in the one-step procedure described in Method 2. That is, given an arbitrary abstraction $\phi$, each $\rho_{1:t,\phi}^{\pi}$ equals to
\begin{eqnarray}\label{eqn:denominator}
    \rho_{j,\phi}^{\pi}=\frac{\pi_{\phi}(A_j|\phi(S_j))}{b_{j,\phi}(A_j|\phi(S_j),A_{j-1},\phi(S_{j-1}),\ldots,A_1,\phi(S_1))}.
\end{eqnarray}
Under $\pi$-irrelevance, the numerator is well-defined and identifiable. However, unlike the standard IS ratio where the denominator depends solely on the current state, the denominator in \eqref{eqn:denominator} depends on the entire history. Notice that the denominator is essentially the data distribution of $A_j$ given $\phi(S_j),A_{j-1},\phi(S_{j-1}),\ldots,A_1,\phi(S_1)$, it is thus identifiable from the offline data as well. 

Under the coverage assumption in Assumption \ref{asmp:coverage}, the behavior policy is bounded away from zero. Since the behavior policy is stationary, this conditional pmf can be represented by
\begin{eqnarray*}
    \mathbb{E} [b(\bullet|S_j)|\phi(S_j),A_{j-1},\phi(S_{j-1}),\ldots,A_1,\phi(S_1)],
\end{eqnarray*}
which is bounded away from zero as well. Consequently, $\rho_{t,\phi}^{\pi}$ is bounded and identifiable. 
  
%Then, to prove this result, 
Let $\rho_{1:t,\phi}^{\pi}$ denote the SIS ratio $\prod_{j=1}^t \rho_{j,\phi}^{\pi}$. It suffices to show
 \begin{eqnarray}\label{eqn:rhopivalid}
     \mathbb{E} (\rho_{1:t}^{\pi} R_t)=\mathbb{E}(\rho_{1:t,\phi}^{\pi}R_t),
 \end{eqnarray}
 for any $t$. Under the Markov assumption, $R_t$ is independent of past state-action pairs given $A_t$ and $S_t$. Consequently, the left-hand-side can be represented as
 \begin{eqnarray*}
     \mathbb{E}[\mathbb{E} (\rho_{1:t-1}^{\pi}|A_t,S_t) \rho^{\pi}(A_t,S_t) R_t].
 \end{eqnarray*}
 Additionally, since the generation $A_t$ depends only on $S_t$, the inner expectation equals $\mathbb{E} (\rho_{1:t-1}^{\pi}|S_t)$ which can be further shown to equal to $p_t^{\pi}(S_t)/p_{\infty}(S_t)$. This allows us to represent the left-hand-side of \eqref{eqn:rhopivalid} by 
 \begin{eqnarray}\label{eqn:rhopivalideq2}
     \mathbb{E} \Big[ \frac{p_t^{\pi}(S_t)}{p_{\infty}(S_t)} \rho^{\pi}(A_t,S_t) R_t\Big].
 \end{eqnarray}
Using similar arguments to the proof of Fisher consistency of MIS estimator, under reward-irrelevance, \eqref{eqn:rhopivalideq2} can be shown to equal to 
\begin{eqnarray*}
    \sum_{\substack{a_1,\cdots,a_{t}\in \mathcal{A}\\ x_1,\cdots,x_{t}\in \mathcal{X}}}\rho_0(x_1)\prod_{k=1}^{t-1} \Big[\pi_{\phi}(a_k|x_k) \mathcal{T}_{\phi}(x_{k+1}|a_k,x_k)\Big]\pi_{\phi}(a_t|x_t)\mathcal{R}_{\phi}(a_t,x_t).
\end{eqnarray*}
Notice that both $\mathcal{T}_{\phi}$ and $\mathcal{R}_{\phi}$ independent of the target policy $\pi$. Using the change of measure theorem, we can  represent above expression by $\mathbb{E} (\rho_{1:t,\phi}^{\pi} R_t)$. This completes the proof.
    
\item \textbf{Fisher consistency of DRL under model-irrelevance}. Since model-irrelevance and $\pi$-irrelevance imply $Q^{\pi}$-irrelevance and the identifiability of $Q^{\pi}$, the conclusion  directly follows from the double robustness of DRL and that in the first bullet point. %, and the first two conclusions of Theorem \ref{thm2}.
   \end{itemize}
\subsection{Theorem \ref{thm3} and its Proof}\label{proofofthm3}
\begin{theorem}[OPE under Type-II MSA]\label{thm3}
    Assume Assumptions \ref{asmp:bounded}--\ref{asmp:stationary} hold, and $\phi$ is a $\pi$-irrelevant type-II MSA.  %abstraction. 
    \begin{itemize}[leftmargin=*]
        \item Then $\phi$ is both $\rho^{\pi}$-irrelevant and $w^{\pi}$-irrelevant. 
        \item Additionally, %both SIS (Method 2) and MIS (Method 3) 
        all the four OPE methods (i.e., Q-function-based, SIS, MIS, DRL) are Fisher consistent when applied to the abstract state space.
        %\item While $\phi$ is not necessarily $Q^{\pi}$-irrelevant, the Q-function-based method (Method 1) is Fisher consistent when applied to the abstract state space.
        %\item DRL (Method 4) is Fisher consistent when applied to the abstract state space.
    \end{itemize}
\end{theorem}
%The first bullet point in 
Theorem \ref{thm3} validates the four OPE methods when applied to the abstract state space. %under the proposed backward-model-irrelevance, whereas the last two bullet points validate the Q-function-based method and DRL.  
%Notice that $\rho^{\pi}$-irrelevance directly follows from the definition of backward-model-irrelevance and $\pi$-irrelevance. 
%We first establish $w^{\pi}$-irrelevance. Under $\rho^{\pi}$- and $w^{\pi}$-irrelevance, the Fisher consistencies of SIS and MIS follow immediately from Theorem \ref{thm1}. We next establish the Fisher consistencies of the Q-function-based method and DRL, respectively. 
We establish the Fisher consistencies of SIS, MIS, Q-function-based method and DRL one by one. 
\begin{itemize}[leftmargin=*]
\item \textbf{Fisher consistency of SIS}. Notice that $\rho^{\pi}$-irrelevance directly follows from the definition of backward-model-irrelevance and $\pi$-irrelevance. It follows from Lemma \ref{thm1} that $\mathbb{E} [f_2(\rho^{\pi})]=\mathbb{E} [f_2(\rho^{\pi}_{\phi})]$. 

Additionally, under $\pi$-irrelevance and behavior-policy-irrelevance, both the numerator and the denominator of the IS ratio $\rho_{\phi}^{\pi}$ are identifiable. Consequently, $\rho_{\phi}^{\pi}$ is identifiable as well. This establishes the Fisher consistency of SIS. 

\item \textbf{Fisher consistency of MIS}. %$\rho^{\pi}$-irrelevance directly follows from the definition of backward-model-irrelevance. 
We first establish the $w^{\pi}$-irrelevance. We next establish the identifiability of $w_{\phi}^{\pi}$. 

To prove the $w^{\pi}$-irrelevance, we begin by defining the marginal density ratio at a given time $t$ as 
\begin{align*}
w^{\pi}_t(a,s)=\frac{\mathbb{P}^{\pi}(A_t=a,S_t=s)}{\mathbb{P}(A_t=a,S_t=s)}.
\end{align*}
Under the stationarity assumption, the denominator is independent of $t$. Notice that $w^{\pi}(a,s)=(1-\gamma)\sum_{t=1}^{\infty} \gamma^{t-1}w_t^{\pi}(a,s)$. Hence, it is sufficient to prove that $\phi$ is $w_t^{\pi}$-irrelevance, for any $t$. 

We prove this by induction. First, when $t=1$, $w_t^{\pi}$ is reduced to $\rho^{\pi}$. Consequently, $w_1^{\pi}$-irrelevance is readily obtained by backward-model-irrelevance and $\pi$-irrelevance. 

Second, suppose we have established $w_t^{\pi}$-irrelevance. We aim to show $w_{t+1}^{\pi}$-irrelevance. 
With some calculations, we obtain that
\begin{align}\label{eqn:wtpirhopi}
\begin{split}
    &\mathbb{E}[w^{\pi}_t(A_{t},S_{t})\rho^{\pi}(A_{t+1},S_{t+1})|A_{t+1}=a,S_{t+1}=s]\\
    =&\sum_{a',s'} w^{\pi}_t(a',s')\frac{\pi(a|s)}{b(a|s)} \mathbb{P}(A_t=a',S_t=s'|A_{t+1}=a,S_{t+1}=s)\\
         =&\sum_{a',s'}\frac{\mathbb{P}^{\pi}(A_t=a',S_t=s')}{\mathbb{P}(A_t=a',S_t=s')} \frac{\pi(a|s)}{b(a|s)}\frac{\mathbb{P}(A_t=a',S_t=s',A_{t+1}=a,S_{t+1}=s)}{\mathbb{P}(A_{t+1}=a,S_{t+1}=s)}\\
    =&\sum_{a',s'}\mathbb{P}^{\pi}(A_t=a',S_t=s') \frac{\pi(a|s)}{b(a|s)} 
    \times\frac{\mathbb{P}(A_{t+1}=a|S_{t+1}=s,A_t=a',S_t=s')}{\mathbb{P}(A_{t+1}=a,S_{t+1}=s)}\mathbb{P}(S_{t+1}=s|A_t=a',S_t=s)\\
   =&\sum_{a',s'}\mathbb{P}^{\pi}(A_t=a',S_t=s') \frac{\pi(a|s)}{\mathbb{P}(A_{t+1}=a,S_{t+1}=s)} 
   \mathbb{P}(S_{t+1}=s|A_t=a',S_t=s)\\
     =&\frac{\mathbb{P}^{\pi}(A_{t+1}=a,S_{t+1}=s)}{\mathbb{P}(A_{t+1}=a,S_{t+1}=s)}\\
    =&w^{\pi}_{t+1}(a,s),
\end{split}
\end{align}
where the third last equality is due to that the behavior policy is stationary. This establishes the link between $w_t^{\pi}$, $\rho^{\pi}$ and $w_{t+1}^{\pi}$.     
    
To prove $w_{t+1}^{\pi}$-irrelevance, we first prove the following equation holds:
\begin{align}\label{back-action}
\begin{split}
  &\sum_{s'\in \phi^{-1}(x')} \mathbb{P}(A_t=a',S_t=s'|A_{t+1}=a,S_{t+1}=s^{(1)})\\
   =  &\sum_{s'\in \phi^{-1}(x')} \mathbb{P}(A_t=a',S_t=s'|A_{t+1}=a,S_{t+1}=s^{(2)}),
\end{split}
\end{align}
whenever $\phi(s^{(1)})=\phi(s^{(2)})$.

Indeed, by  \eqref{eqn:backirrelevance-state}, we obtain that
\begin{align*}
   &\sum_{s'\in \phi^{-1}(x')} \mathbb{P}(A_t=a',S_t=s'|A_{t+1}=a,S_{t+1}=s^{(1)})\\
     =&\sum_{s'\in \phi^{-1}(x')} \frac{\mathbb{P}(A_t=a',S_t=s',A_{t+1}=a,S_{t+1}=s^{(1)})}{\mathbb{P}(A_{t+1}=a,S_{t+1}=s^{(1)})}\\
     =&\sum_{s'\in \phi^{-1}(x')} \frac{\mathbb{P}(A_{t+1}=a|S_{t+1}=s^{(1)},A_t=a',S_t=s')}{\mathbb{P}(A_{t+1}=a,S_{t+1}=s^{(1)})}
     \mathbb{P}(S_{t+1}=s^{(1)},A_t=a',S_t=s')\\
     =&\sum_{s'\in \phi^{-1}(x')} 
     \mathbb{P}(A_t=a',S_t=s'|S_{t+1}=s^{(1)})\\
    =&\sum_{s'\in \phi^{-1}(x')} 
     \mathbb{P}(A_t=a',S_t=s'|S_{t+1}=s^{(2)})\\
   = &\sum_{s'\in \phi^{-1}(x')} \mathbb{P}(A_t=a',S_t=s'|A_{t+1}=a,S_{t+1}=s^{(2)}).
\end{align*}
Consequently, 
%To prove \eqref{mdr-irrelevance}, we employ the induction method. 
%In fact, when $t=1$, $ w^{\pi}_1(a,s^{(1)})=\rho^{\pi}(a,s^{(1)})=\rho^{\pi}(a,s^{(2)})=w^{\pi}_1(a,s^{(2)})$.  
%Assume that for any $k\leq t$, \eqref{mdr-irrelevance}
%  holds. Then for $k=t+1$, we have
\begin{align*}
    &w^{\pi}_{t+1}(a,s^{(1)})\\
 %   &\mathbb{E}[w^{\pi}_t(A_{t},S_{t})\rho^{\pi}(A_{t+1},S_{t+1})|A_{t+1}=a,S_{t+1}=s^{(1)}]\\
    =&\sum_{a',s'} w^{\pi}_t(a',s')\rho^{\pi}(a,s^{(1)}) \mathbb{P}(A_t=a',S_t=s'|A_{t+1}=a,
    S_{t+1}=s^{(1)})\\  
     =&\sum_{a',s'} w^{\pi}_t(a',s')\rho^{\pi}(a,s^{(2)}) \mathbb{P}(A_t=a',S_t=s'|A_{t+1}=a,
    S_{t+1}=s^{(2)})\\ 
    =&w^{\pi}_{t+1}(a,s^{(2)}),
    \end{align*}
where the second last equation follows from $w_t^{\pi}$-irrelevance, $\rho^{\pi}$-irrelevance and \eqref{back-action}. This yields $w_{t+1}^{\pi}$-irrelevance, and subsequently $w^{\pi}$-irrelevance, by induction. By Lemma \ref{thm1}, $w^{\pi}$-irrelevance further yields $\mathbb{E} [f_3(w^{\pi})]=\mathbb{E} [f_3(w^{\pi}_{\phi})]$. 

It remains to prove the identifiability of $w_{\phi}^{\pi}$. We similarly define
\begin{eqnarray*}
    w_{t,\phi}^{\pi}(a,x)=\frac{\mathbb{P}^{\pi}(A_t=a,\phi(S_t)=x)}{\mathbb{P}(A_t=a,\phi(S_t)=x)}. 
\end{eqnarray*}
It follows that $w_{\phi}^{\pi}=\sum_{t\ge 1} \gamma^{t-1} w_{t,\phi}^{\pi}$. Again, $w_{1,\phi}^{\pi}$ corresponds to $\rho_{\phi}^{\pi}$, which is identifiable under backward-model-irrelevance and $\pi$-irrelevance. 

Based on the aforementioned arguments, we can show that
\begin{eqnarray*}
    w_{t+1,\phi}^{\pi}(a,x)=\sum_{a',x'} w_{t,\phi}^{\pi}(a',x')\rho_{\phi}(a|x)\mathbb{P}(A_t=a',\phi(S_t)=x'|A_{t+1}=a,\phi(S_{t+1})=x),
\end{eqnarray*}
where the last term on the RHS is well-defined according to \eqref{back-action}. Suppose we have shown the identifiability of $w_{t,\phi}^{\pi}$. Then each term on the RHS is identifiable. This proves the identifiability of $w_{t+1,\phi}^{\pi}$. By induction, $w_{t,\phi}^{\pi}$ is identifiable for each $t\ge 1$.  

According to the coverage and stationarity assumptions in Assumptions \ref{asmp:coverage} and \ref{asmp:stationary}, the denominators in $\{w_{t,\phi}^{\pi}\}$ are bounded away from zero. Consequently, $\{w_{t,\phi}^{\pi}\}$ are uniformly bounded. By letting $t\to \infty$, we obtain the identifiability of $w_{\phi}^{\pi}$. The proof is thus completed.

\item \textbf{Fisher consistency of Q-function-based method}. 
We first show that $\mathbb{E} [f_1(Q^{\pi}_{\phi})]=\mathbb{E} [f_1(Q^{\pi})]$ under $\pi$-irrelevance. This is immediate by noting that 
\begin{eqnarray*}
    \mathbb{E} [f_1(Q^{\pi})]=J(\pi)=\sum_{t\ge 1} \gamma^{t-1}\mathbb{E}^{\pi}(R_t)=\sum_{a,x}\sum_{t\ge 1} \gamma^{t-1}\mathbb{E}^{\pi}(R_t|A_1=a,\phi(S_1)=x)\\
    \times \mathbb{P}^{\pi}(A_1=a|\phi(S_1)=x)\mathbb{P}(\phi(S_1)=x)=\mathbb{E} [f_1(Q^{\pi}_{\phi})],
\end{eqnarray*}
where the first term on the second line equals $\pi(a|s)$ for any $s$ such that $\phi(s)=x$, under $\pi$-irrelevance. 

It remains to prove the identifiability of $Q_{\phi}^{\pi}$ under $\pi$- and backward-model-irrelevance. First, we establish the identifiability of $\mathbb{E}^{\pi}(R_1|A_1=a,\phi(S_1)=x)$. By definition
\begin{eqnarray*}
    \mathbb{E}^{\pi}(R_1|A_1=a,\phi(S_1)=x)=\frac{\mathbb{E}^{\pi}[R_1\mathbb{I}(A_1=a,\phi(S_1)=x)]}{\mathbb{P}^{\pi}(A_1=a,\phi(S_1)=x)}.
\end{eqnarray*}
Using the change of measure theorem, the numerator equals
\begin{eqnarray*}
    \mathbb{E}\Big[\rho^{\pi}(a|S_1)R_1\mathbb{I}(A_1=a,\phi(S_1)=x)\Big]=\mathbb{E}\Big[\rho_{\phi}^{\pi}(a|x)R_1\mathbb{I}(A_1=a,\phi(S_1)=x)\Big]\\
    =\rho_{\phi}^\pi(a|x)\mathbb{E}\Big[R_1\mathbb{I}(A_1=a,\phi(S_1)=x)\Big],
\end{eqnarray*}
where the first equation holds due to $\pi$- and behavior-policy-irrelevance. Notice that the denominator equals $\mathbb{P}(\phi(S_1)=x)\pi_{\phi}(a|x)$, it follows that
\begin{eqnarray*}
    \mathbb{E}^{\pi}(R_1|A_1=a,\phi(S_1)=x)=\mathbb{E}(R_1|A_1=a,\phi(S_1)=x),
\end{eqnarray*}
which is identiable from the data. 

Similarly, one can show that $\mathbb{P}^{\pi}(\phi(S_2)=x'|A_1=a,\phi(S_1)=x)=\mathbb{P}(\phi(S_2)=x'|A_1=a,\phi(S_1)=x)$ is identifiable as well. 

Now, the identifiability can be readily obtained if we show $(\phi(S_t),A_t,R_t)_{t\ge 1}$ remains an MDP. In that case, standard Q-learning algorithms can be applied to such a reduced MDP to consistently identify $Q_{\phi}^{\pi}$. Such an MDP property will be proven in Section \ref{sec:MSA} under a more challenging setting that allows the behavior policy to be history-dependent.

\item \textbf{Fisher consistency of DRL}. Due to the double robustness property of DRL, the conclusion directly follows from the last conclusion of Theorem \ref{thm1} and the first two conclusions of Theorem \ref{thm3}.

\end{itemize}

\subsection{Theorem \ref{thm4} and its proof}\label{proofof thm4}

\begin{theorem}[OPE under the iterative procedure]\label{thm4} 
Assume Assumptions \ref{asmp:bounded}--\ref{asmp:stationary} hold. With the refined backward-model-irrelevance, 
the four OPE methods are Fisher consistent when applied to the abstract state space produced by the proposed DSA, regardless of the number of iterations conducted.
 %   \end{itemize}
\end{theorem}
First, we notice that according to the DRL's double robustness property, its Fisher consistency is achieved when either the MIS or the Q-function-based estimator is Fisher consistent. Consequently, it suffices to prove the Fisher consistencies of the rest three estimators. 

Additionally, at the first iteration, these Fisher consistencies directly follows from Lemma \ref{thm2}. Consequently, it suffices to prove the Fisher consistencies at later iterations. Below, we first prove the Fisher consistencies of SIS, MIS and Q-function-based estimator at the second iteration. Next, we prove the resulting abstraction is a Markov state abstraction \citep{allen2021learning} in that the data generating process when confined to the abstract state space remains an MDP. This together with Lemma \ref{thm2} proves the Fisher consistencies at the third iteration. Using similar arguments, we can establish the Fisher consistencies at any $K>3$ iterations. The proof can thus be completed. 
\subsubsection{Fisher consistencies at the 2nd iteration}
It is worthwhile mentioning that at the second iteration, the refined type-II MSA is defined with respect to the abstract state space induced by type-I MSA $\phi_1$ at the first iteration instead of the ground state space. In particular, we require 
\begin{eqnarray}\label{eqn:bphi1t}
    b_{\phi_1,t}(a_t|x_t^{(1)},a_{t-1},x_{t-1}^{(1)},\cdots,x_1^{(1)})=b_{\phi_1,t}(a_t|x_t^{(2)},a_{t-1},x_{t-1}^{(2)},\cdots,x_1^{(2)}),
\end{eqnarray}
for any $t$ and $\{a_t\}_t$, whenever $\phi_2(x_t^{(1)})=\phi_2(x_t^{(1)})$ for any $\{x_t^{(1)}\}_t$ and $\{x_t^{(2)}\}_t$, where $b_{\phi_1,t}$ denote the history-dependent behavior policy (see also the denominator of Equation \ref{eqn:denominator}), and
\begin{eqnarray}\label{eqn:backwardcond}
\begin{split}
    &\sum_{x\in \phi_2^{-1}(x_2)}\mathbb{P}(A_t=a,\phi_1(S_t)=x|\phi_1(S_{t+1})=x^{(1)})\\
    =&\sum_{x\in \phi_2^{-1}(x_2)}\mathbb{P}(A_t=a,\phi_1(S_t)=x|\phi_1(S_{t+1})=x^{(2)}),
\end{split}
\end{eqnarray}
whenever $\phi_2(x^{(1)})=\phi_2(x^{(2)})$. 

In the following, we prove the Fisher consistencies of SIS, MIS and Q-function-based method one by one:
\begin{itemize}[leftmargin=*]
    \item \textbf{Fisher consistency of SIS}. When restricting to the abstract state space induced by $\phi_1$, the resulting behavior policy is not guaranteed to be Markovian. To address this challenge, SIS employs the history-dependent IS ratio defined in Equation \ref{eqn:denominator} to maintain consistency. Let $\rho_{1:t,\phi_1}^{\pi}$ and $\rho_{1:t,\phi_2\circ \phi_1}^{\pi}$ denote the history-dependent SIS ratios at the first and second iterations, respectively. Under $\pi$-irrelevance and the refined history-dependent-behavior-policy-irrelevance (see \eqref{eqn:bphi1t}), it is immediate to see that $\rho_{1:t,\phi_1}^{\pi}=\rho_{1:t,\phi_2\circ \phi_1}^{\pi}$ so that $\rho^{\pi}$-irrelevance is achieved at the second iteration. This in turn validates the unbiasedness of the SIS estimator based on $\{\rho_{1:t,\phi_2\circ \phi_1}^{\pi}\}_t$.

    Finally, notice that the denominators in $\rho_{1:t,\phi_2\circ \phi_1}^{\pi}$ are identifiable since these probabilities are defined with respect to the offline data distribution. Meanwhile, under $\pi$-irrelevance, the numerator is identifiable as well. This proves the identifiability of these history-dependent IS ratios. The Fisher consistency of SIS thus follows. 
    
    \item \textbf{Fisher consistency of MIS}. We first show that the abstraction produced by DSA at the second iteration achieves $w^{\pi}$-irrelevance, i.e., $w^{\pi}_{\phi_1}=w^{\pi}_{\phi_2\circ \phi_1}$. Next, we establish the identifiability of the MIS ratio $w^{\pi}_{\phi_2\circ \phi_1}$. 

    The proof is very similar to that of Theorem \ref{thm2}. Specifically, define
    \begin{eqnarray*}
        w_{t,\phi_1}^{\pi}(a,x)=\frac{\mathbb{P}^{\pi}(A_t=a,\phi_1(S_t)=x)}{\mathbb{P}(A_t=a,\phi_1(S_t)=x)},
    \end{eqnarray*}
    we have $w_{\phi_1}^{\pi}=(1-\gamma)\sum_{t\ge 1}\gamma^{t-1} w_{t,\phi_1}^{\pi}$. It suffices to establish the irrelevance in $w_{t,\phi_1}^{\pi}$ for any $t$. 
 
    When $t=1$, $w_{1,\phi_1}^{\pi}$ is reduced to the IS ratio $\pi_{\phi_1}(a|x)/b_{1,\phi_1}(a,x)$. 
    Under $\pi$-irrelevance and behavior-policy-irrelevance (by setting $j$ in Equation \ref{eqn:bphi1t} to $1$), the numerator $\pi_{\phi_1}$ and denominator $b_{1,\phi_1}$ equal $\pi_{\phi_2\circ \phi_1}$ and $b_{1,\phi_2\circ \phi_1}$ (the behavior policy when restricting to the abstract state space produced by DSA at the 2nd iteration), respectively. This establishes the irrelevance in $w_{1,\phi_1}^{\pi}$.

    Suppose we have proven the irrelevance in $w_{t,\phi_1}^{\pi}$, we aim to show the irrelevance in $w_{t+1,\phi_1}^{\pi}$. Under the stationarity assumption in Assumption \ref{asmp:stationary}, by setting $j$ in Equation \ref{eqn:historydependentbehaviorpolicy} to $2$, we obtain that
    \begin{eqnarray}\label{eqn:bphi2t}
    \begin{split}
        &\mathbb{P}(A_t=a_2|\phi_1(S_t)=x_2^{(1)},A_{t-1}=a_{1},\phi_1(S_{t-1})=x_{1}^{(1)})\\
        =&\mathbb{P}(A_t=a_2|\phi_1(S_t)=x_2^{(2)},A_{t-1}=a_{1},\phi_1(S_{t-1})=x_{1}^{(2)}),
    \end{split}
    \end{eqnarray}
    for any $t$, $a_1$ and $a_2$, whenever $\phi_2(x_1^{(1)})=\phi_2(x_1^{(2)})$ and $\phi_2(x_2^{(1)})=\phi_2(x_2^{(2)})$. 

    Let $X_t$ denote $\phi_1(S_t)$ for any $t$. We next claim that
    \begin{eqnarray}\label{eqn:wt+1}
        w_{t+1,\phi_1}(a,x)=\mathbb{E}\Big[w_{t,\phi_1}(A_t,X_t) \frac{\pi_{\phi_1}(A_t|X_t)}{b_{2,\phi_1}(A_{t+1}|X_{t+1},A_t,X_t)} |A_{t+1}=a,X_{t+1}=x\Big].
    \end{eqnarray}
    Notice that this formula is very similar to \eqref{eqn:wtpirhopi}. The only difference lies in that the denominator of the IS ratio on the RHS is no longer Markovian. Rather, it depends on the current state as well as the previous state-action pair. Meanwhile, \eqref{eqn:wt+1} can be proven using similar arguments to \eqref{eqn:wtpirhopi}. 

    Based on \eqref{eqn:wt+1}, we are ready to establish the irrelevance in $w_{t+1,\phi}^{\pi}$. In particular, looking at the RHS of \eqref{eqn:wt+1}, both $w_{t,\phi_1}$ and the IS ratio $\pi_{\phi_1}/b_{2,\phi_1}$ depend on $X_t$ and $X_{t+1}$ only through their abstractions $\phi_2(X_t)$ and $\phi_2(X_{t+1})$. Meanwhile, the conditional distribution of $A_t,X_t$ given $A_{t+1},X_{t+1}$ depends on $X_t$ and $X_{t+1}$ through their abstractions, as well, given \eqref{eqn:bphi2t} and \eqref{eqn:backwardcond}. This establishes the irrelevance in $w_{t+1,\phi_1}$. By induction, we have proven the irrelevance in $w_{t,\phi_1}$ for any $t$. Under the coverage assumption in Assumption \ref{asmp:coverage}, these ratios are uniformly bounded. It follows that the limit $\lim_T \sum_{t=1}^T \gamma^{t-1}w_{t,\phi_1}^{\pi}$ is well-defined. By setting $T\to \infty$, we obtain the irrelevance in $w_{\phi_1}^{\pi}$.

    So far, we have established the $w^{\pi}$-irrelevance. This in turn yields $\mathbb{E}[f_3(w_{\phi_1}^{\pi})]=\mathbb{E}[f_3(w_{\phi_2\circ \phi_1}^{\pi})]$, according to Lemma \ref{thm1}. It remains to prove the identifiability of $w_{\phi_2\circ \phi_1}^{\pi}$. However, this can be proven using similar arguments to the proof of Theorem \ref{thm2}. Specifically, we first observe that $w_{\phi_2\circ \phi_1}^{\pi}=\lim_T \sum_{t=1}^T \gamma^{t-1}w_{t,\phi_2\circ \phi_1}^{\pi}$. Next, when setting $t=1$, the identifiability of $w_{1,\phi_2\circ \phi_1}^{\pi}$ is readily available, given that of $\rho_{\phi_2\circ \phi_1}^{\pi}$. Finally, since $ w_{t+1,\phi_2\circ \phi_1}(a,x_2)$ equals
    \begin{eqnarray*}
       \mathbb{E}\Big[w_{t,\phi_2\circ\phi_1}(A_t,\phi_2(X_t)) \frac{\pi_{\phi_2\circ \phi_1}(A_t|\phi_2(X_t))}{b_{2,\phi_2\circ \phi_1}(A_{t+1}|\phi_2(X_{t+1}),A_t,\phi_2(X_t))} |A_{t+1}=a,\phi_2(X_{t+1})=x_2\Big],
    \end{eqnarray*}
    we can employ similar arguments to the proof of Theorem \ref{thm2} to prove the identifiability of the above expression, assuming $w_{t,\phi_2\circ\phi_1}$ is identifiable. By induction, this establishes the identifiability of $w_{\phi_2\circ\phi_1}$. 
    \item \textbf{Fisher consistency of Q-function-based method}. The Fisher consistency of Q-function-based method can be established in a similar manner to that in Theorem \ref{thm2}. Specifically, under $\pi$-irrelevance, it is trivial to show $\mathbb{E} [f_1(Q^{\pi}_{\phi_1})]=J(\pi)=\mathbb{E} [f_1(Q^{\pi}_{\phi_2\circ \phi_1})]$. Meanwhile, its identifiability is readily obtained based on the results in the following section, which proves that the process $(\phi_2(\phi_1(S_t)),A_t,R_t)_{t\ge 1}$ remains an MDP. 
\end{itemize}
\end{document}